\documentclass{article} 
\usepackage{conf/iclr2026_conference,times}

\usepackage{amsmath,amsfonts,bm}

\def\eqref#1{equation~\ref{#1}}

\def\1{\bm{1}}

\DeclareMathAlphabet{\mathsfit}{\encodingdefault}{\sfdefault}{m}{sl}
\SetMathAlphabet{\mathsfit}{bold}{\encodingdefault}{\sfdefault}{bx}{n}

\usepackage[utf8]{inputenc} 
\usepackage[T1]{fontenc}    
\usepackage{hyperref}       
\usepackage{url}            
\usepackage{booktabs}       
\usepackage{amsfonts}       
\usepackage{nicefrac}       
\usepackage{microtype}      
\usepackage[capitalize,noabbrev]{cleveref}
\usepackage{graphicx}
\usepackage{placeins}

\usepackage{enumitem}
\usepackage[dvipsnames]{xcolor}
\usepackage{wrapfig}

\definecolor{cornflowerblue}{rgb}{0.39, 0.58, 0.93}
\hypersetup{
    colorlinks=true,
    linkcolor=cornflowerblue,
    filecolor=magenta,      
    urlcolor=teal,
    citecolor=cornflowerblue,
    pdftitle={Teaching Pretrained Language Models to Think Deeper with Retrofitted Recurrence},
    pdfpagemode=FullScreen,
}

\usepackage{xspace}

\newcommand{\olmo}{OLMo\xspace}
\newcommand{\tinyllama}{TinyLlama\xspace}
\newcommand{\llama}{Llama\xspace}
\newcommand{\sqrtsched}{\(1\text{-sqrt}\)\xspace}

\title{Teaching Pretrained Language Models to Think Deeper with Retrofitted Recurrence}

\author{Sean McLeish$^{1}$\thanks{Correspondence to: \texttt{smcleish@umd.edu}.} , Ang Li$^{2}$, John Kirchenbauer$^{1}$, Dayal Singh Kalra$^{1}$, Brian R. Bartoldson$^{3}$, 
\\ \textbf{Bhavya Kailkhura$^{3}$, Avi Schwarzschild$^{4}$, Jonas Geiping$^{5}$, Tom Goldstein$^{1}$,} \\ \textbf{Micah Goldblum$^{6}$}\\
$^{1}$ University of Maryland,
$^{2}$ New York University, \\
$^{3}$ Lawrence Livermore National Laboratory,
$^{4}$ University of North Carolina, \\
$^{5}$ ELLIS Institute T\"ubingen, Max Planck Institute for Intelligent Systems,  T\"ubingen AI Center, \\ 
$^{6}$ Columbia University
}

\iclrfinalcopy 
\begin{document}

\maketitle

\begin{abstract}

Recent advances in depth-recurrent language models show that recurrence can decouple train-time compute and parameter count from test-time compute.
In this work, we study how to convert existing pretrained non-recurrent language models into depth-recurrent models.
We find that using a curriculum of recurrences to increase the effective depth of the model over the course of training preserves performance while reducing total computational cost.
In our experiments, on mathematics, we observe that converting pretrained models to recurrent ones results in better performance at a given compute budget than simply post-training the original non-recurrent language model.
\\\textbf{Code}: \href{https://github.com/mcleish7/retrofitting-recurrence}{github.com/mcleish7/retrofitting-recurrence}
\\\textbf{Models}: \href{https://huggingface.co/collections/tomg-group-umd/retrofitting-recurrence}{huggingface.co/collections/tomg-group-umd/retrofitting-recurrence}

\end{abstract}

\section{Introduction}~\label{sec:intro}
\begin{wrapfigure}{r}{0.5\textwidth}
    \centering
    \vspace{-0.5cm}
    \includegraphics[width=\linewidth]{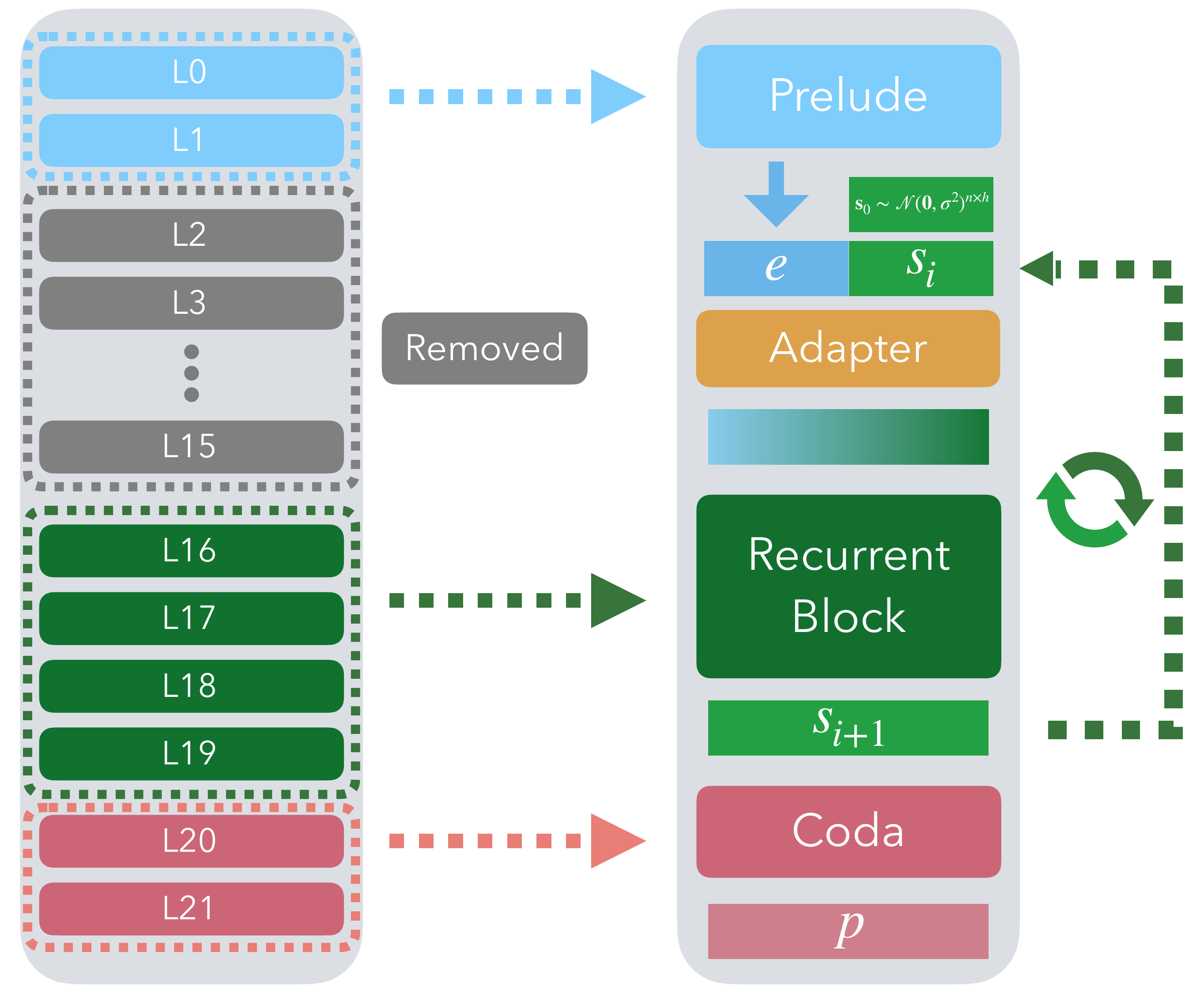}
    \caption{\textbf{We take layers from pretrained language models and recur a core block.} We take early layers to form the \textit{prelude} and later layers to form the \textit{recurrent block} and \textit{coda}, removing the layers in between. After each recurrence, we concatenate the output of the prelude with the output of the recurrent block (or random noise at time zero) and apply a linear adapter.}
    \label{fig:explainer}
    \vspace{-0.5\baselineskip}
\end{wrapfigure}
\looseness -1
{\em Test-time compute scaling} refers to the use of additional computation during inference to improve model outputs.  By decoupling computation intensity from model size, test-time compute scaling achieves superior benchmark scores without requiring more model parameters or additional pretraining.  The mainstream paradigm for test-time scaling involves generating many tokens, either in chain-of-thought traces or by generating many candidate solutions and choosing the best \citep{snell2024scaling,guo2025deepseek}. 
An emerging alternative paradigm for test-time scaling leverages depth-recurrence, by which a language model can simply recur layers for more iterations to expend more compute.  Depth-recurrence has the advantage that increasing compute does not increase memory consumption or context size during inference.  Moreover, not requiring the model to verbalize thoughts as tokens may allow for more complex reasoning to happen within the latent space where there is higher information bandwidth. Finally, recurrent networks can be trained on standard data sources and do not require training with bespoke reasoning traces in the domain of interest. 

\hspace{0pt}\citet{geiping2025scaling} pretrain a depth-recurrent transformer from scratch on \(800\) billion tokens at substantial cost.  
Although their model can reuse parameters at test time to scale up compute and improve performance, their work also uses a large number of recurrent iterations during training, which significantly slows down training compared to a fixed depth model with the same parameter count.  
This inspires us to focus on the training efficiency of depth-recurrent models. 

In this work, we study fast procedures for converting fixed depth models into depth-recurrent models through continued pretraining, visualized in \Cref{fig:explainer}.
Because transformer models include residual connections \citep{he2015deep} that write updates back into the same residual stream, transformer layers operate in a shared representation space \citep{elhage2021mathematical}. This makes it possible to ``loop'' a block of layers from a pretrained language models by feeding the output of the block back into itself as input.  By training a model while it operates in this looped mode, the model learns to exploit recurrence to improve performance.
Our main experiments demonstrate that \textit{\tinyllama-1.1B-intermediate-step-1431k-3T} \citep{zhang2024tinyllama}, \textit{\olmo-2-0425-1B} \citep{olmo20242} and \textit{\llama-3.2-1B} \citep{grattafiori2024llama} can be converted into depth-recurrent transformers. 
We view post training fixed depth models with recurrence as a simple addition to the training pipeline, similar to how one would extend the context length during the later stages of pretraining \citep{grattafiori2024llama}. We observe that doing so improves performance on reasoning tasks that are known to differentially benefit from additional test-time compute~\citep{geiping2025scaling}.

We focus on two efficiency goals. First, we want the model initialized from pretrained weights to outperform a model trained from scratch on a per-training-FLOP basis.  
Since parameters are both added to and removed from the original model when converting it into a depth-recurrent one, 
this knowledge transfer goal is non-trivial.  We show in \Cref{fig:long-runs} that initializing a depth-recurrent model from \llama-3.2-1B weights strongly outperforms the randomly initialized model in terms of loss and benchmark accuracy per training FLOP spent.  Second, we want the performance of the pre-trained model to increase after conversion to recurrent form.  We find that with a well-formed data curriculum, recurrence results in an increase in accuracy on math tasks while maintaining high accuracy on a broad suite of language modeling benchmarks (see \Cref{fig:tinyllama-data-mix} and \Cref{tab:data-mix}).

Overall, we show that retrofitting recurrence into pretrained language models is an efficient way to train performant depth-recurrent models. In summary, our contributions are as follows:
\begin{enumerate}[topsep=0.0cm,itemsep=0.0cm,leftmargin=0.6cm]
    \item We show that initializing parameters of recurrent models from those of a pretrained fixed depth model is significantly more efficient than using a random initialization (\Cref{fig:long-runs}).
    \item We propose a curriculum over recurrent depths, slowly increasing the average number of recurrent iterations during training to maintain performance while improving training speed (\Cref{fig:schedule-mean-n}).
    \item \looseness -1 We show that, using Common Crawl math data, we can convert \tinyllama, \olmo, and \llama models into recurrent models that achieve better GSM8K and MATH performance than base models (Figures \ref{fig:tinyllama-GSM8K} and~\ref{fig:olmo-MATH}).
    \item Since we remove layers when converting fixed depth models to recurrent ones, we find that introducing a ``healing'' period with minimal distribution shift helps recover basic language modeling performance before switching to task-specific data to further refine the depth-recurrent model's reasoning performance (\Cref{fig:tinyllama-data-mix} and \Cref{tab:data-mix}).
\end{enumerate}

\section{Related Work}~\label{sec:rel-work}
\textbf{Recurrent models.\ } 
It has been shown that ``universal transformers'' based on recurrence are Turing-complete \citep{dehghani2018universal}.
Recurrent transformers with weight shared layers but a fixed layer repetition count have been studied in detail~\citep{lan2019albert,takase2021lessons,fan2024looped,bae2024relaxed,gao2024algoformer,ng2024loop,csordas2024moeut,mcleish2024transformers,saunshi2025reasoning,zeng2025pretraining}.
Adaptive-depth mechanisms have been studied with the specific goal of increasing computation efficiency~\citep{graves2016adaptive,elbayad2019depth, schwarzschild2021can, bansal2022end}.
A more advanced class of recurrent transformer can utilize an internal mechanism to exit after a data-dependent number of recurrences~\citep{geiping2025scaling,aleksandrov2025abbie,chen2025inner,bae2025mixture}.
\citet{raposo2024mixture} propose mixture of depths models which adaptively route tokens through or around each transformer block. 
\citet{mohtashami2023cotformer} augment mixture of depths with weight sharing, extended by \citet{bae2025mixture} with adaptive exiting to further increase efficiency.

\textbf{Model surgery.\ } There is a rich literature on methods for making post-hoc changes to model architecture and size~\citep{chen2015net2net,wei2016network}.
\citet{li2025zero} finetune looped models initialized from the GPT-2 \citep{brown2020language} and OPT \citep{zhang2022opt} checkpoints finding small gains from finetuning and looping under-trained models on multiple choice benchmarks over the base checkpoints.
In particularly relevant prior work, \citet{bae2024relaxed} study converting pretrained transformer language models into recurrent models using just \(2\) or \(3\) recursions. 
Notably, the authors maintain the same shape as the base model and require low rank adapters \citep{hu2022lora} to recover performance of the base model.
\citet{bae2024relaxed} also find that recurring more leads to performance decreases in the post-trained model which means that their approach cannot effectively leverage additional compute at test time.
Unlike \citet{bae2024relaxed}, our approach does not require distillation or auxiliary adapters and does benefit from additional test time computation.
Finally, in concurrent work \citet{koishekenov2025encode} convert \olmo \citep{olmo20242} models into depth recurrent models.
They also use prelude, recurrent block, and coda structures but do not use input injection, keep all parameters when converting the model, and train with a fixed number of recurrences. While they do demonstrate modest performance improvements they do not present their results in terms of training or inference compute making the degree of cost-benefit afforded by their method difficult to discern.

\textbf{Latent reasoning.} \citet{wang2025hierarchical} introduce the Hierarchical Reasoning Model (HRM); an architecture designed to better align with certain anthropomorphic biases for compositional intelligence.
However, \citet{arc_hrm_2025} performs further ablations on the HRM architecture and finds only the main recurrence is needed for reasoning performance, reducing the HRM to a model similar to that of \citet{geiping2025scaling} without the ability to extrapolate in recurrence. We begin our own research by re-purposing aspects of the pretraining recipe developed by
\citet{geiping2025scaling} to train a large recurrent language model from scratch; the first work to establish that latent reasoning as a scalable, alternate approach for pretraining transformer language models.
We detail how our architecture and training recipe is derived from theirs more formally in \Cref{sec:details}.

We provide an extended discussion of other related work in \Cref{app-sec:app-rel-work}.

\section{Experimental Setup}~\label{sec:details}
\textbf{Model Definition.\ }
Using the same notation as \citet{geiping2025scaling}, here we define the structure of the class of recurrent models we study.
We define \(P\) as the prelude, \(R\) as the recurrent block and \(C\) as the coda; each of which is a set of unique transformer blocks with the embeddings included in \(P\) and unembeddings in \(C\).
We visualize the architecture in \Cref{fig:explainer} (right).
\(R\) begins with a linear adapter which takes the concatenation of \(s_i\) and \(e\), hence if the width of the model is \(h\) the linear layer goes from \(2h\) to \(h\).
Given vocabulary set \(V\), for an input sequence $\mathbf{x} \in V^n$ and a number of recurrences $r$, the model output distribution $\mathbf{p}$ is defined as follows.

\begin{align*}
    & Prelude & Recurrent\ Block &  &Coda\quad\ &\\
    &\mathbf{e} = P(\mathbf{x})
    &\mathbf{s}_0 \sim \mathcal{N}(\mathbf{0}, \sigma^2)^{n \times h},
    \ \ 
    \mathbf{s}_i = R(\mathbf{e}, \mathbf{s}_{i-1})\ \ &\textnormal{for}\ \  i \in \lbrace 1, \dots, r \rbrace 
    &\mathbf{p} = C(\mathbf{s}_r)&
\end{align*}

\citet{geiping2025scaling} use a ``scalable initialization'' \citep{takase2023spike} for their \texttt{Huginn-0125} model. Such schemes allow model shape to be altered whilst maintaining training stability. 
We also use this random initialization when training from scratch.
To allow for adaptive recurrence at test time, \citet{geiping2025scaling} sample \(r\) from a Poisson-Lognormal distribution with a mean of \(32\) at each training step.
They also employ a truncated backpropagation procedure, only propagating gradients through at most the last \(8\) passes through \(R\).
This reduces training time and allows for very large values of \(r\) without exhausting GPU memory.
When we say a model is trained with train recurrence \(=k\), this means that the mean of the Poisson-Lognormal distribution is equal to \(k\) during training.
We note that the prelude parameters are still updated as the skip connection to the recurrent block allows for gradient propagation from the output.

\textbf{Model Surgery.\ }
Similar to \citet{geiping2025scaling}, we use tuple notation to define the number of transformer layers in each of the prelude, recurrent block, and coda. For example, \((2,4,2)\) means there are \(2\) transformer layers in the prelude, \(4\) in the recurrent block, and \(2\) in the coda.
To improve efficiency at large numbers of test recurrences, we do not use every layer from the pretrained model when adapting it into a depth-recurrent model.
We find that selecting the early layers for the prelude and later layers for the recurrent block and coda performs best (see Appendix Figures \ref{fig:which-layers} and \ref{fig:which-layers-shortgpt}).
For example, if the model we are using has \(22\) layers and we take a \((4,8,4)\) configuration. This corresponds to selecting layers \([0,1,2,3],[10,11,12,13, 14,15,16,17],[18,19,20,21]\); we use this selection for our \((4,8,4)\) \tinyllama based models.
We visualize our methodology in \Cref{fig:explainer}.
We detail the exact parameter counts and layers taken from pretrained models for our recurrent models in \Cref{app-sec:param-counts}.
In Appendix \Cref{fig:which-layers-shortgpt}, we compare to the ShortGPT pruning method \citep{men2024shortgpt} to select layers to drop from the parent model when forming the recurrent model.
We find our selection to be better for depth-recurrent model post-training.
We also compare to taking all layers from \tinyllama to form a \((6,10,6)\) model and to a \((2,4,2)\) \tinyllama model in Appendix \Cref{fig:which-layers-all-layers}.

We inherit the conventions of the models we are converting.
Specifically, \citet{geiping2025scaling} use normalizations four times in each decoder block and additionally use the final norm before the coda; we reduce to two norms in each decoder block and remove the dual use of the final layer norm.
We also use grouped-query attention \citep{ainslie2023gqa}, train all models with a context length of 1024, and do not weight-tie the embedding and unembedding layers.
We present additional technical training details in \Cref{app-sec:app-details}.

We emphasize that although two of models we analyze in this paper share the ``llama'' name they are different models, trained independently.
The two models are different shapes, with \tinyllama being \(6\) layers deeper than \llama, but narrower (smaller residual stream) as they both contain approximately \(1\) billion parameters.
\tinyllama uses the \llama-2 vocabulary, whereas \llama-3 uses a vocabulary over \(4\times\) larger.
Finally, \tinyllama is trained with a next token prediction cross entropy loss from random initialization for \(3\) trillion tokens, whereas \llama is initialized by pruning \llama-3.1-8B and then using logit level distillation from \llama-3.1-8B and \llama-3.1-70B for \(9\) trillion tokens \citep{meta_llama_3_2_announcement}. 
Furthermore, the \olmo models use QK-norm and a post-normalization scheme unlike the llama models which use a pre-normalization scheme and do not use a QK-norm.

\textbf{Calculating Training Cost.\ }
For a recurrent model, the number of {\em unique parameters} refers to the number of distinct, trainable parameters in the model without double counting parameters that are shared across recurrences; we simply use the term \textit{parameters} in this paper\footnote{We also exclude embedding and unembedding parameters in this count.}.
One can also consider the {\em effective parameters} of a recurrent model by including repetitions across recurrences.
However, for clarity, throughout the rest of the work we quantify the size of a recurrent model evaluated at different depths in terms of Floating Point Operations (\textit{FLOPs}) rather than describing parameter re-use.
In other words, increasing the number of iterations performed by the recurrent block increases the amount of computation invested while number of actual parameters in the model remains fixed.

When calculating training FLOPs for standard fixed depth transformers, we use the approximation \(\text{FLOPs} = 6ND\) \citep{kaplan2020scaling}, where \(N\) is non-embedding parameters and \(D\) is number of training tokens.
However, recurrent models require a different rule.
As we only backpropagate through at most the last \(8\) iterations of the recurrent block, we split the \textit{effective parameter} count (\(N\)) into two parts: \(N_1\) which includes all parameters with gradients recorded and \(N_2\) which includes all parameters that are used in the forward pass without gradients.
We calculate \(N_1\) and \(N_2\) using the mean number of recurrences during training.
This gives \(FLOPs=(6 N_1 + 2N_2)D\) for our recurrent models.

\section{Training Recurrent Language Models}~\label{sec:results}
Our main experimental results are presented in four subsections.
In \Cref{subsec:inits}, we find that a pretrained initialization outperforms a random initialization in terms of loss and benchmark performance.
Then, in \Cref{subsec:scheduling}, we use a curriculum to schedule the mean of the Poisson-Lognormal distribution, showing that this can reduce training costs without negatively impacting loss.
In \Cref{subsec:tinyllama-llama}, we show that depth-recurrent post-training is more efficient than training non-recurrent models for math problems.
Finally, in \Cref{subsec:data-mix}, we demonstrate that with a good data curriculum, depth-recurrent models can be good general language models in addition to achieving higher accuracy on grade school math problems despite having fewer parameters.

\subsection{Efficiently Initializing Recurrent Transformers}~\label{subsec:inits}
\begin{figure}[t!]
    \centering
    \includegraphics[width=\linewidth]{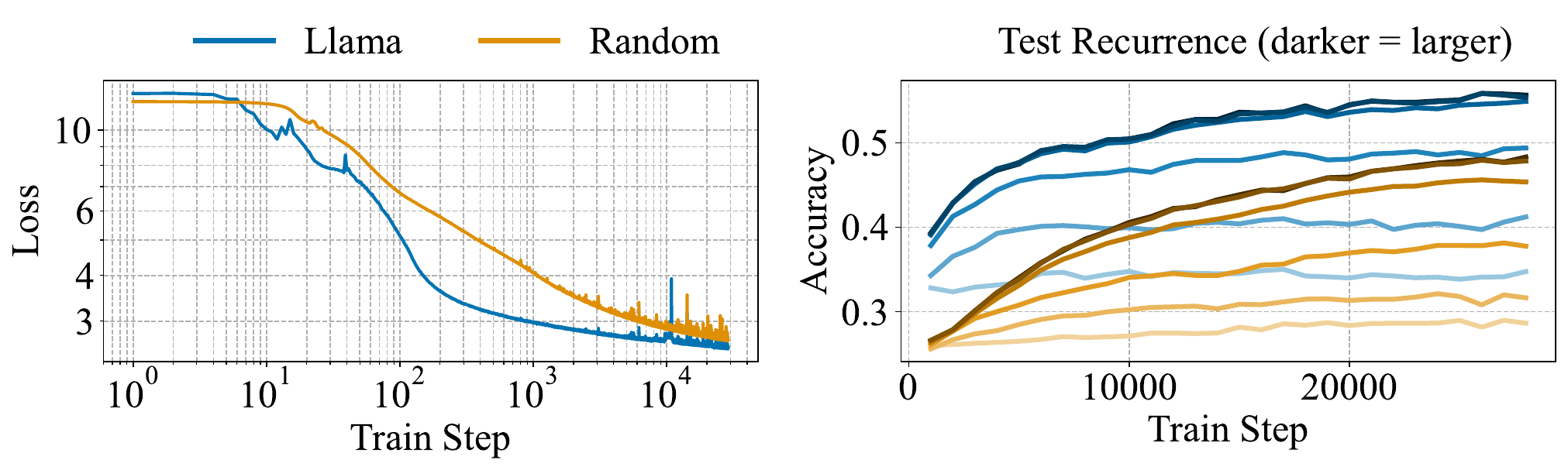}
    \caption{\textbf{Initializing from pretrained \llama layers gives a significant advantage in loss and benchmark accuracy.}
    \textbf{Left: } Loss over training step for \(120\) billion tokens for models initialized from \llama-3.2-1B layers and randomly \citep{takase2023spike}. Although starting higher, the model initialized from \llama weights achieves lower losses consistently than the model initialized randomly. \textbf{Right: } Zero shot accuracy on Hellaswag \citep{zellers2019hellaswag} over training step for recurrences \([1,2,4,8,16,32]\). We see the \llama based model (blue) achieves higher accuracy quicker and leverages recurrence effectively from early training steps. We record accuracy over recurrence for a suite of language modeling benchmarks in Appendix \Cref{app-tab:random_init_benchmarks_full}.}
    \label{fig:long-runs}
\end{figure}
We begin by demonstrating that \textbf{using a pretrained initialization outperforms a random initialization for depth-recurrent models}.
We train two models for approximately \(120\) billion tokens on FineWeb-Edu \citep{penedo2024fineweb} data with a mean number of recurrences of \(32\).
\Cref{fig:long-runs} visualizes the training loss and Hellaswag accuracy curves over training for a \((2,4,2)\) model initialized from \llama-3.2-1B and from random initialization, following \citet{takase2023spike}.
On the left, we see the initialization from pretrained \llama layers yields a large efficiency gain in terms of loss.
On the right, we show that the model initialized from pretrained \llama layers achieves higher benchmark accuracy earlier on Hellaswag \citep{zellers2019hellaswag}.
By training step \(1000\), the \llama initialized model is already leveraging recurrence to increase accuracy, unlike the random initialization for which all recurrences are achieving just over random accuracy.

In Appendix \Cref{app-tab:random_init_benchmarks_full}, we show the accuracy at \(28,000\) steps for both models over multiple recurrence levels on a suite of language modeling benchmarks, finding that \textbf{initializing from pretrained \llama weights causes a significant increase in accuracy in all cases}.
In \Cref{app-subsec:model-surgery}, we also present additional experiments including a cooldown for \(12\) billion additional tokens.
Extrapolating the loss curves in log-linear space suggests it would take at least approximately \(950\) billion tokens for these loss curves to intersect (see Appendix \Cref{app-fig:long-runs-with-lines}).
It is likely that this is an underestimate of the true number of tokens required for the models to achieve loss parity as the curves are not perfectly log-linear at the end of our data.

\subsection{Scheduling Recurrences}~\label{subsec:scheduling}
Using truncated backpropagation means the forward pass for our recurrent models consumes a larger share of runtime than it would for a non-recurrent model.
Hence, reducing the time spent on the forward pass for our models has a large impact on training time.
With this insight, we explore an efficient curriculum which schedules the mean of the Poisson-Lognormal distribution.
This curriculum is analogous to the gradual stacking technique \citep{gong2019efficient,reddi2023efficient,saunshi2024inductive,du2024stacking} which increases the depth of a non-recurrent model by duplicating layers within the model during training and then training them independently.
We visualize our curricula in Appendix \Cref{app-fig:schedule-mean-explainer}.

\begin{figure}[h!]
    \centering
    \includegraphics[width=\linewidth]{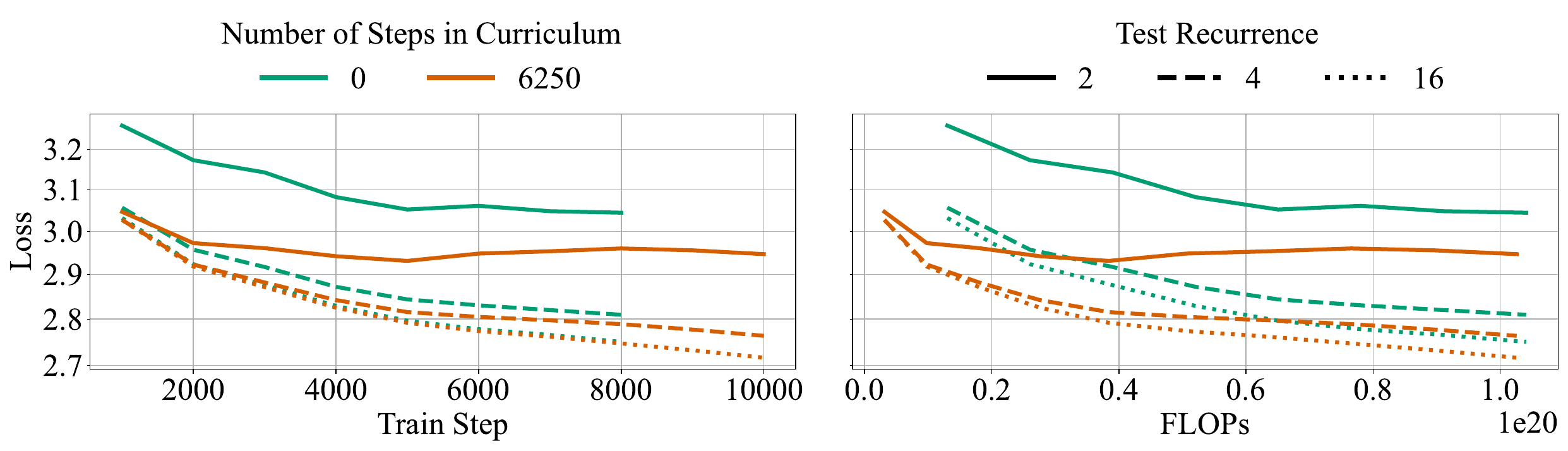}
    \caption{\textbf{Scheduling the mean of the depth distribution is efficient in terms of \textit{both} data and compute.} We report validation loss over multiple recurrent depths in terms on steps (i.e. data) on the left and in terms of FLOPs on the right.
    We see that linearly scheduling the number of recurrences up to the final mean \((32)\) over a long period of training decreases the validation loss, hence the curriculum is both data and compute efficient. 
    Alternative length curricula and more test recurrent depths are shown in Appendix \Cref{app-fig:schedule-mean-n-all}.}
    \label{fig:schedule-mean-n}
\end{figure}

\Cref{fig:schedule-mean-n} measures the validation loss on one million tokens taken every \(1000\) training steps for \((2,4,2)\) models initialized from \llama layers.
This is the same as in \Cref{fig:long-runs} but for a shorter time horizon of \(48\) hours on \(4\) MI300A GPUs which equates to approximately \(1e^{20}\) FLOPs.
In \Cref{fig:schedule-mean-n} (left), we see that linearly scheduling the recurrent depth has a small positive impact on the validation loss as a function of steps.
Furthermore, on the right, we see that linearly \textbf{scheduling greatly improves the efficiency in terms of loss improvement as a function of FLOPs spent during training}.
In \Cref{app-subsec:schedule-rec}, we show that scheduling the maximum backpropagation depth over training is better in terms of FLOPs but worse in terms of steps, and therefore less efficient overall than scheduling the mean depth but still valuable when trying to reach the lowest possible loss in a given period of time.
Finally, in Appendix \Cref{app-fig:1-sqrt-sched}, we show that a more efficient curricula where we schedule according to a \sqrtsched function (visulaized in \Cref{app-fig:schedule-mean-explainer}) are as good as linear curricula for \tinyllama.

\subsection{How to Retrofit Recurrence}~\label{subsec:tinyllama-llama}
Next, we investigate how to efficiently retrofit depth-recurrence into pretrained non-recurrent transformers.
First, we find Muon to be a better optimizer than AdamW when training recurrent models in \Cref{subsubsec:muon}.
In \Cref{subsubsec:tinyllama}, we analyze \tinyllama, \olmo and \llama models.
In all cases, under the same training FLOP budget, depth-recurrent models with fewer parameters can achieve higher accuracy on math problems than the non-recurrent parent model.
We extend these results for \tinyllama, \olmo and \llama in \Cref{app-subsec:retrofit}.

\begin{figure}[b!]
  \centering
  \begin{minipage}[t!]{0.475\linewidth}
    \includegraphics[width=\linewidth]{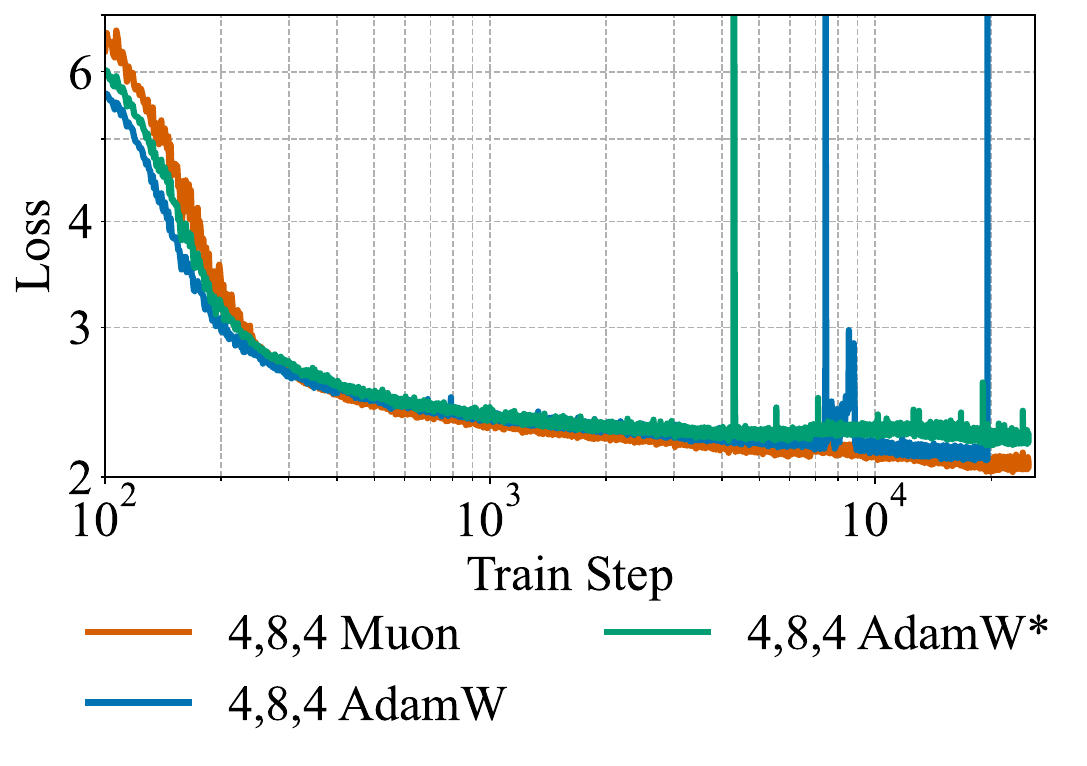}
  \end{minipage}\hfill
  \begin{minipage}[t!]{0.525\linewidth}
    \vspace{-1.5em}
    \includegraphics[width=\linewidth]{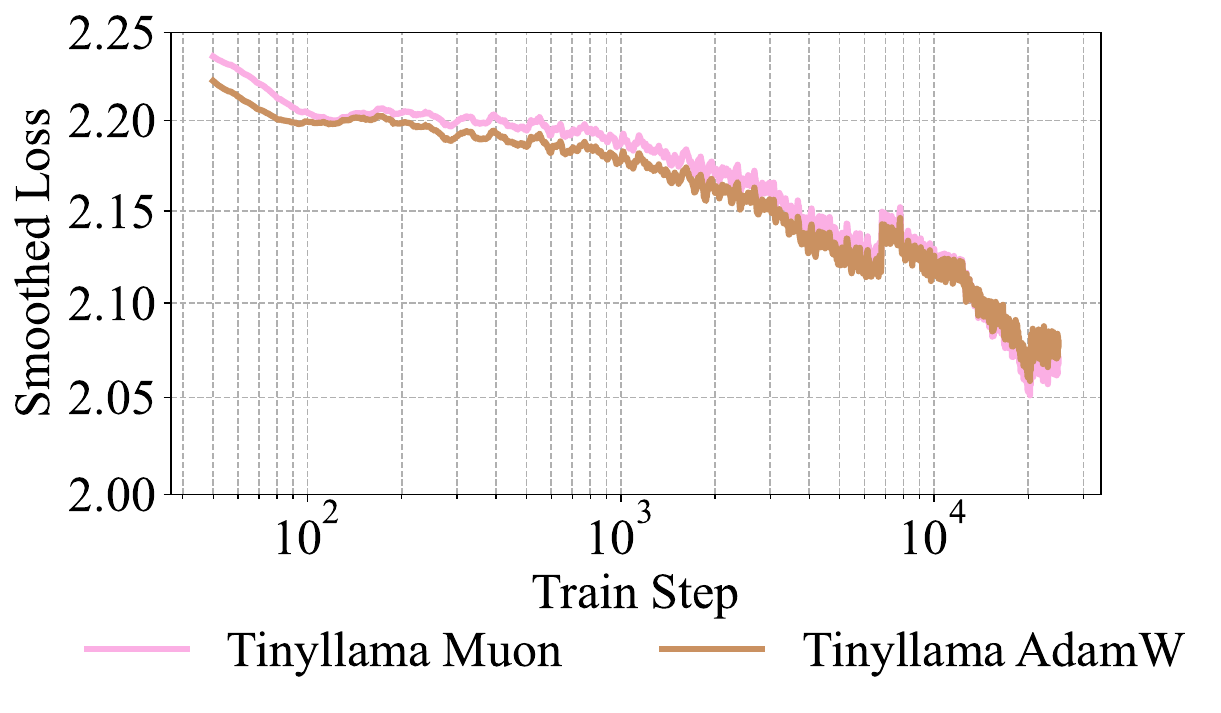}
  \end{minipage}
  \caption{\textbf{Muon improves over AdamW when training recurrent models.}
  \textbf{Left: }Loss vs. step for multiple training runs on the same data order with different optimizers, using a learning rate of \(5e^{-5}\) for AdamW and \(0.001\) for Muon. Muon is the most stable and achieves the lowest loss for recurrent models. Note, the AdamW line ends early as the loss spikes and becomes \texttt{NaN}.
  \textbf{Right: } Loss (smoothed over \(50\) steps) vs. step for AdamW and Muon. For the non-recurrent \tinyllama model there is minimal difference between optimizers.}
  \label{fig:muon}
\end{figure}

\subsubsection{Optimization}~\label{subsubsec:muon}
We begin by initializing models from \tinyllama-1.1B-intermediate-step-1431k-3T.
We consider a \((4,8,4)\) \tinyllama recurrent configuration, dropping out \(6\) layers (layers \(4,5,6,7,8\) and \(9\), using \(0\)-indexing) from the original model.
This yields approximately \(700\) million remaining parameters in this recurrent model, \(72.7\%\) of the parameters in the parent non-recurrent \tinyllama model.
We also compare our \((4,8,4)\) to a \((7,8,7)\) model in \Cref{app-fig:extend-prelude-coda}, finding that removing the layers is efficient, even from the prelude and coda.
Full parameter counts are provided in \Cref{app-sec:param-counts}.

In \Cref{fig:muon} (left), 
\textbf{the Muon optimizer improves over AdamW} for our recurrent models as it achieves lower loss and removes loss spikes during training.
For the non-recurrent \tinyllama models, the difference is much less pronounced, but we still see a small gain using the Muon optimizer.
We smooth the loss over \(50\) steps to make this more visible in the plot.
In \Cref{fig:muon}, we also compare to the variant of AdamW which is used by \citet{geiping2025scaling}, and we refer to this variant as AdamW*.
AdamW* modifies AdamW by including 
update clipping, removing the \(\varepsilon\) constant \citep{wortsman2023stable,everett2024scaling}, and using a different decoupling method than the PyTorch AdamW implementation \citep{adamw-decoupling-blog}.
In subsequent experiments, we optimize all models with Muon.

\subsubsection{Recurrent Models are Efficient to Train}~\label{subsubsec:tinyllama}
\looseness -1
In our next set of experiments, while we continue training our \((4,8,4)\) \tinyllama configuration, we build another set of models initialized from the weights of \olmo-2-0425-1B. 
For \olmo, we construct \((4,6,4)\) configurations, removing \(2\) layers (layers \(4\) and \(5\) with \(0\) indexing) from the pretrained model.
This leaves approximately \(900\) million remaining parameters in the recurrent model, which equates to \(87.5\%\) of the pretrained models parameters.

\begin{figure}[ht!]
    \centering
    \includegraphics[width=\linewidth]{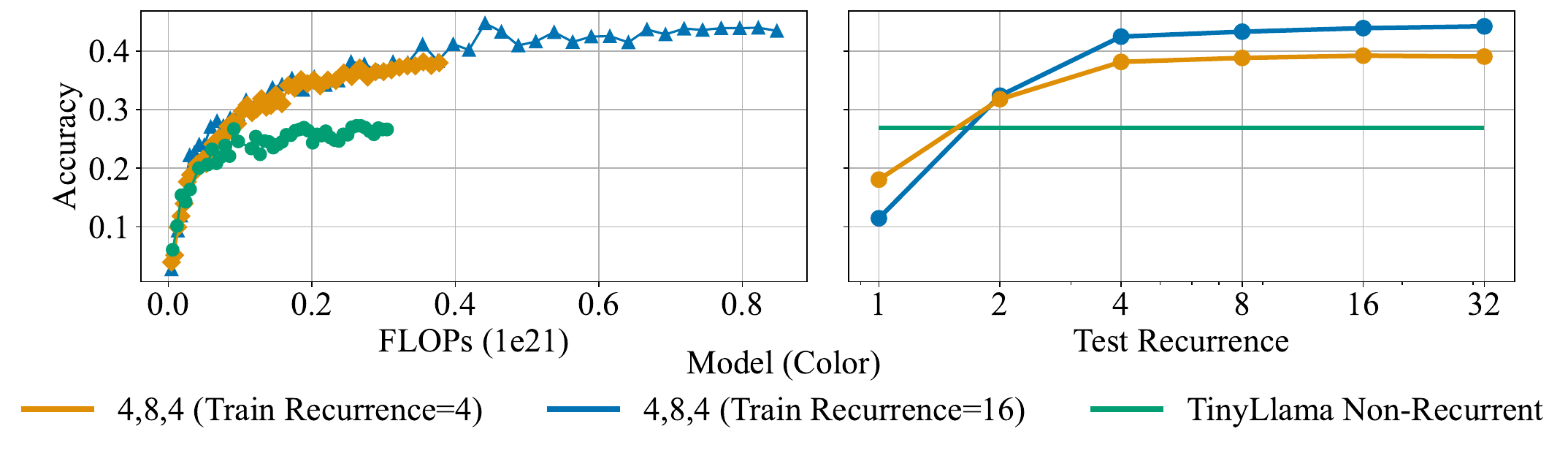}
    \caption{\textbf{Recurrence improves reasoning on GSM8K for \tinyllama, even when controlling for FLOPs.}
    We train \((4,8,4)\) and non-recurrent models for approximately \(50\) billion tokens of Nemotron-CC-Math-v1 data. 
    \textbf{Left: }We plot accuracy over the number of FLOPs used during training. We see that recurrent models, trained with scheduling, can efficiently outperform the non-recurrent baseline.
    \textbf{Right:} We plot accuracy over the number of recurrences for inference. We see the recurrent models are competitive with the fixed depth baseline and can outperform it by using more recurrences and therefore more FLOPs. We plot each individual models accuracy over training step and recurrence in full in \Cref{app-subsubsec:retrofit-tinyllama}, including for training recurrence \(8\) and \(32\). Evaluations on the final checkpoint over tasks shown in \Cref{tab:data-mix} are in Appendix \Cref{app-tab:tinyllama-all-evals}. We also provide identical experiments for \olmo and \llama in \Cref{app-subsec:retrofit}.}
    \label{fig:tinyllama-GSM8K}
\end{figure}

\begin{figure}[ht!]
    \centering
    \includegraphics[width=\linewidth]{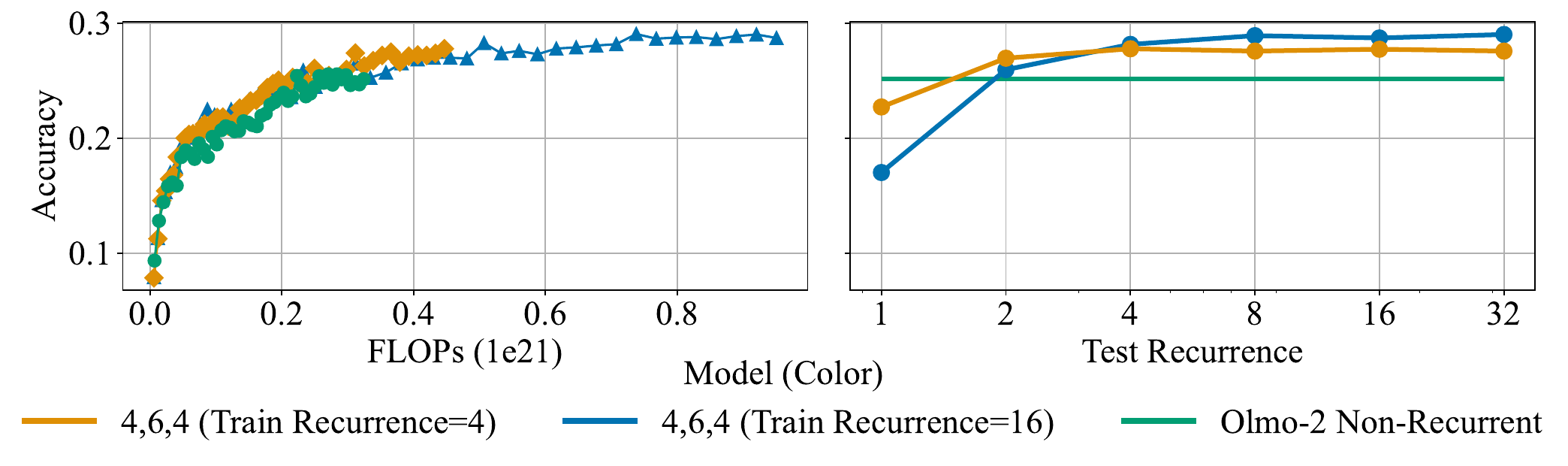}
    \caption{\textbf{Recurrence efficiently improves reasoning on MATH for \olmo.}
    \textbf{Left: }We plot accuracy over the number of FLOPs used during training. We see that recurrent models, trained with scheduling, can efficiently outperform the non-recurrent baseline when trained on the same tokens.
    \textbf{Right: }We plot accuracy over the number of recurrences used for inference. We see the recurrent models are competitive with the fixed depth baseline (green horizontal line) and can outperform it by using more recurrences and therefore more FLOPs. We plot each individual models accuracy over training and recurrence in full in \Cref{app-subsubsec:retrofit-olmo}, including for training recurrence \(8\) and \(32\). Evaluations on the final checkpoint over tasks shown in \Cref{tab:data-mix} are in Appendix \Cref{app-tab:olmo-all-evals}.We also provide identical experiments for \tinyllama and \llama in \Cref{app-subsec:retrofit}.}
    \label{fig:olmo-MATH}
\end{figure}

In Figures \ref{fig:tinyllama-GSM8K} and \ref{fig:olmo-MATH}, we train models on approximately \(50\) billion tokens of \texttt{Nemotron-CC-Math-v1-4plus} \citep{mahabadi2025nemotron} data, with a \sqrtsched curriculum for the first \(75\%\) of training and constant mean recurrences thereafter, and evaluate on GSM8K \citep{cobbe2021training} and MATH \citep{hendrycksmath2021}, for \tinyllama and \olmo, respectively.
For our GSM8K and MATH evaluations, we use a single shot example in context when evaluating. 
For GSM8K, we record the flexible extract accuracy to avoid formatting of the answer being a confounder, and for MATH we use the Minerva \citep{lewkowycz2022solving}  criteria with Math-Verify \citep{Kydlicek_MathVerify}.
Controlling for training FLOPs, both Figures \ref{fig:tinyllama-GSM8K} and \ref{fig:olmo-MATH} (left) show that it is efficient to train recurrent models.
The depth-recurrent models achieve comparable performance to the non-recurrent baseline when evaluated at smaller training budgets but continue to improve as more compute is invested while accuracy for the non-recurrent model plateaus.
We emphasize that all of these experiments utilize the same training dataset presented in the same order.
The differences in curve length come from the additional FLOPs required to train the recurrent models (which require more FLOPs per parameter) for the same number of steps.
The end of each line shown in Figures \ref{fig:tinyllama-GSM8K} and \ref{fig:olmo-MATH} (left) corresponds to the exact same number of tokens seen for each model.
For each recurrent model we visualize the accuracy using the test recurrence equal to the mean training recurrence.

In Figures \ref{fig:tinyllama-GSM8K} and \ref{fig:olmo-MATH} (right), we plot accuracy against number of recurrences used during inference for the models at the end of training. 
We see that recurrent models improve performance over the non-recurrent baseline significantly when utilizing more test-time compute.
Moreover, combining this with Appendix Figures \ref{app-fig:tinyllama-GSM8K-effective-params} and \ref{app-fig:olmo-MATH-effective-params}, we conclude that recurrent models are competitive on a per-FLOP basis for inference despite containing fewer trainable parameters at any FLOPs count. Overall, \textbf{depth-recurrent models are able to leverage compute to achieve a higher overall performance with fewer parameters than their non-recurrent counterparts}.

\begin{figure}[ht!]
    \centering
    \includegraphics[width=\linewidth]{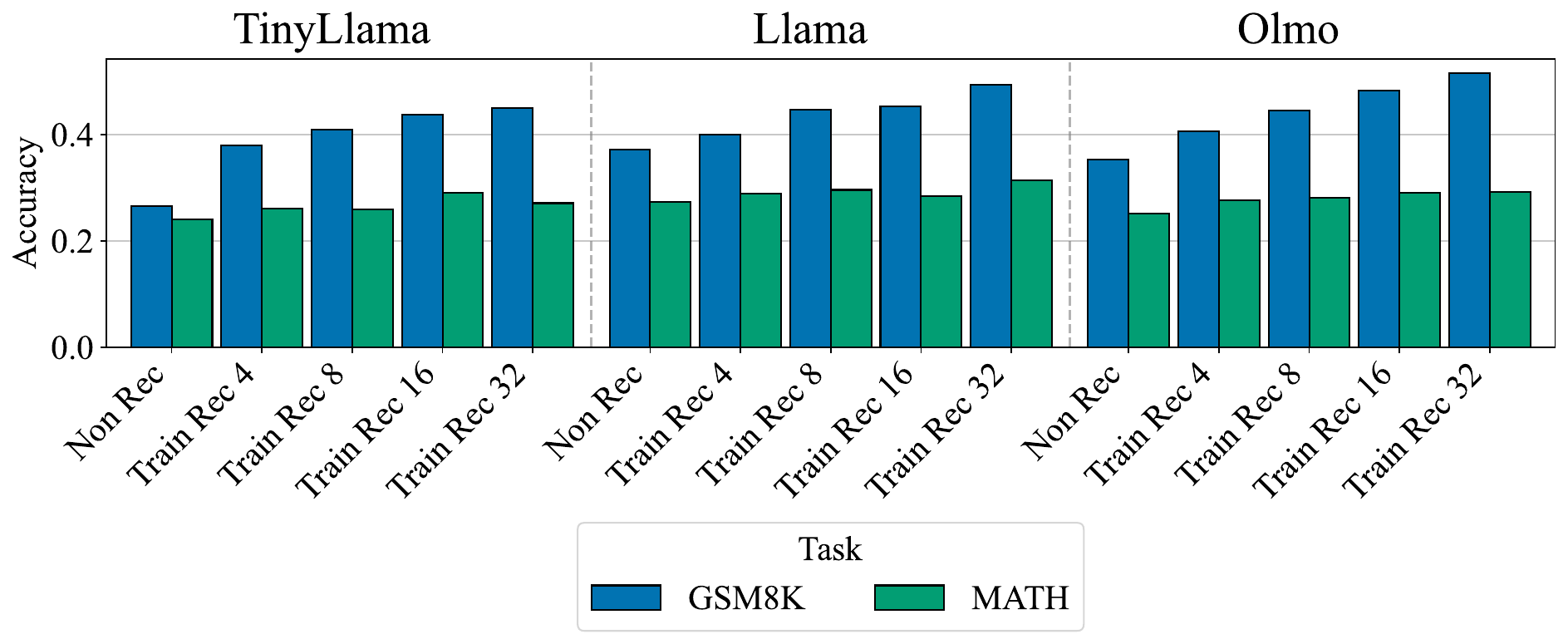}
    \caption{\textbf{Recurrent models achieve higher final checkpoint accuracy on GSM8K and MATH.} We plot the final checkpoint accuracy on GSM8K and MATH for the non-recurrent baseline and multiple training recurrences for \tinyllama, \llama and \olmo, using test recurrence \(32\) for all recurrent models. 
    Full accuracies including recurrences \(1,2,4,8\) and \(16\) can be seen in Tables \ref{app-tab:tinyllama-all-evals}, \ref{app-tab:olmo-all-evals} and \ref{app-tab:llama-all-evals}.}
    \label{fig:bar-plot}
\end{figure}

We construct our final set of models from the weights of \llama-3.2-1B. 
For \llama, we construct \((4,6,4)\) configurations, removing \(2\) layers (layers \(4\) and \(5\) with \(0\) indexing) from the pretrained model.
In \Cref{fig:bar-plot}, we visualize the final checkpoint accuracy for \tinyllama, \llama and \olmo on both GSM8K and MATH, seeing gains in all cases using recurrence.
We provide full visualizations of accuracy over train recurrences \(4,8,16\) and \(32\) for both training and inference on GSM8K and MATH for all three model families in \Cref{app-subsec:retrofit}.
While our results are generally congruous, we do note some  differences in the results across datasets and models.
For GSM8K, we see that larger training recurrences lead to a larger performance improvement on a per FLOP basis for \olmo and \llama, however this trend does not hold for \tinyllama (see \Cref{app-fig:tinyllama-GSM8K-all}).
For both tasks, we see that the accuracy achieved by the \llama and \olmo models is higher than that of the \tinyllama based models.
This suggests that \textbf{using a stronger set of pretrained weights transfers additional knowledge to the final depth-recurrent model.}

\subsection{Data Mixtures}~\label{subsec:data-mix}
In previous experiments, we observe that training strictly on math data slightly degrades performance on non-reasoning based evaluations such as Hellaswag, Arc and OpenbookQA (see Appendix Tables \ref{app-tab:tinyllama-all-evals}, \ref{app-tab:olmo-all-evals} and \ref{app-tab:llama-all-evals}).
To address this degradation, we train on an even mix of FineWeb-Edu \citep{penedo2024fineweb}, Nemotron-Pretraining-SFT-v1-General \citep{nvidia2025nvidianemotronnano2}, and Nemotron-Pretraining-SFT-v1-Math \citep{nvidia2025nvidianemotronnano2}.
We also specifically remove rows in the Nemotron-Pretraining-SFT-v1 dataset generated by reasoning models trained with reinforcement learning (e.g., DeepSeek-R1 \citep{guo2025deepseek}), as well as the user–assistant tags.

In our first experiment on data mixtures, we train \((4,8,4)\) \tinyllama models for \(26\) billion tokens on an even mix of the three datasets; we call this \textit{single phase} training.
Since we remove layers during recurrent retrofitting, we hypothesize that the depth-recurrent models must first recover their basic language modeling abilities before they can efficiently learn the high-quality Nemotron-Pretraining-SFT-v1 data.
To test this hypothesis, we then construct a simple two phase training procedure involving an initial ``healing'' period followed by a phase of high-quality training.
In \textit{two phase} training, we train for \(26\) billion tokens of FineWeb-Edu followed by the same data as seen in the single phase training, totaling \(52\) billion tokens.
For our recurrent models, we use  a linear curriculum  for \(25\%\) of total steps.
We note it is common to heal models after pruning to regain language modeling performance \citep{yang2024laco,men2024shortgpt}.

\begin{figure}[h!] 
    \centering
    \includegraphics[width=\linewidth]{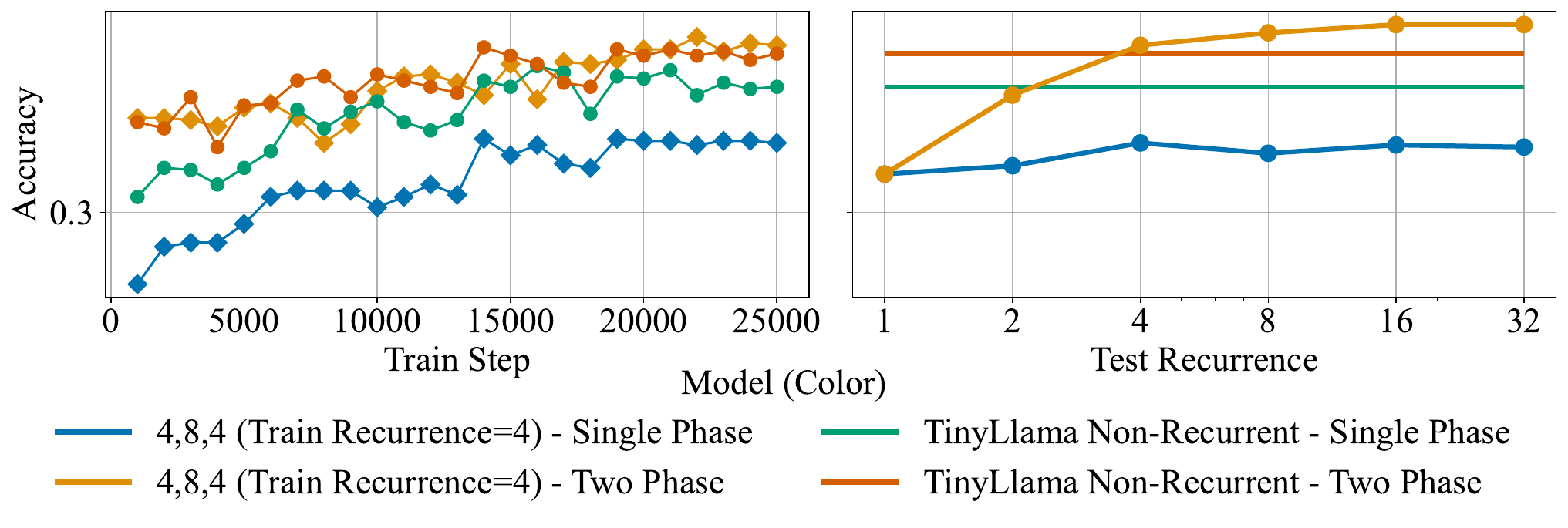}
    \caption{\textbf{High quality data and curricula improve recurrent model performance on non-reasoning benchmarks.} \textbf{Left: } We plot accuracy on Arc-Challenge over training for the \(26\) billion tokens on FineWeb-Edu and Nemotron-SFT data, i.e. after healing for two phase training. We see the training accuracy of the non-recurrent model does not differ significantly between single or two phase training. For the depth-recurrent model, two phase training outperforms single phase by \(5\%\) suggesting the healing period helps the model recover language modeling ability after surgery. \textbf{Right: } Accuracy over multiple recurrences at the end of training. We see the depth-recurrent model with two phase training can use recurrence to extend its accuracy to \(37.7\%\) by utilizing more FLOPs during inference. We repeat our Arc-Challenge accuracies in \Cref{tab:data-mix} for clarity at test recurrences \(1\) and \(32\).}
    \label{fig:tinyllama-data-mix}
\end{figure}

In \Cref{fig:tinyllama-data-mix}, we visualize accuracy on Arc-Challenge over training for the \(26\) billion tokens on the combination of FineWeb-Edu, Nemotron-Pretraining-SFT-v1-General, and Nemotron-Pretraining-SFT-v1-Math data, i.e. the secondary phase after healing for the two phase approach.
We see that under single phase training, when we directly train on the mix of all three datasets, the final recurrent model is worse than the non-recurrent model.
Next, we observe that during two phase training, the non-recurrent model sees a small increase in accuracy over single phase training.
Intuitively, this could be explained by the fact that the initial model is already trained for \(3\) trillion tokens of web text and as there is no model surgery performed on the non-recurrent baseline, there is nothing to explicitly ``heal.'' 
However, for our depth-recurrent model, two phase training provides an increase of \(5\%\) on Arc-C, demonstrating that the initial \(26\) billion token healing period is effective in helping the model to regain basic language modeling abilities.
Our results demonstrate that \textbf{a data curriculum designed to minimize initial distribution shift after model surgery helps depth-recurrent models maintain basic language modeling performance while improving on math problems.}

In \Cref{tab:data-mix}, we measure zero-shot accuracy on Arc-Easy \citep{clark2018think}, Arc-Challenge \citep{clark2018think}, Hellaswag (HS) \citep{zellers2019hellaswag}, Winogrande (WG) \citep{sakaguchi2021winogrande}, MMLU \citep{hendrycks2020measuring}, PIQA \citep{bisk2020piqa}, and OpenbookQA (OBQA) \citep{mihaylov2018can}.
We see that the depth-recurrent model achieves high scores across all benchmarks only outperformed by the non-recurrent model on MMLU by less than one standard error.
We include the Huginn-0125 evaluations from \citet{geiping2025scaling}, comparisons to the base \tinyllama, as well as evaluations using more test recurrences for our models in Appendix \Cref{app-tab:data-mix-full}.
Our models are competitive with the much larger \texttt{Huginn-0125} model released by \citet{geiping2025scaling} achieving an MMLU score over \(12\%\) higher and GSM8K performance \(10\%\) higher than their published evaluation results.
Overall, we find \textbf{depth-recurrent models can improve performance on math tasks while improving or maintaining performance across a broad range of language modeling benchmarks despite having fewer unique trainable parameters.}


\begin{table}[th!]
    \centering
    \caption{\textbf{High quality data and curricula improve recurrent model performance across benchmarks.} We see that depth-recurrence achieves better accuracy on non-math when using two phase training and confirm that the depth-recurrent models improve as a function of test-time recurrence. On the other hand, for the non-recurrent baseline we see single phase and two phase training perform similarly. Full results in Appendix \Cref{app-tab:data-mix-full}, including test recurrences \(2,4,8\) and \(16\).}
    \begin{tabular}{cccccccccc}
        \toprule
         Test Rec & Arc-E & Arc-C & HS & WG & MMLU & PIQA & OBQA  & GSM8K & MATH\\
        \midrule
        \multicolumn{10}{c}{4,8,4 (Train Recurrence=4) - Single Phase} \\
        \midrule
        1 & 50.0 & 31.6 & 50.8 & 58.0 & 35.7 & 69.3 & 38.8 & 25.6 & 8.8 \\
        32 & 52.7 & 32.7 & 58.2 & \textbf{61.1} & 39.4 & 71.4 & 38.6 & \textbf{52.0} & \textbf{14.5} \\
        \midrule
        \multicolumn{10}{c}{4,8,4 (Train Recurrence=4) - Two Phase} \\
        \midrule
        1 & 52.7 & 31.6 & 51.5 & 56.7 & 36.2 & 71.0 & 39.4 & 26.5 & 9.7 \\
        32 & \textbf{65.2} & \textbf{37.7} & \textbf{60.4} & 60.5 & 44.8 & \textbf{73.6} & \textbf{40.0} & 51.2 & 14.2 \\
        \midrule
        \multicolumn{10}{c}{TinyLlama-1.1b-3T
        Static Depth - Single Phase} \\
        \midrule
         & 61.2 & 35.2 & 58.9 & 60.5 & \textbf{45.1} & 71.4 & 39.2 & 46.2 & 14.4 \\
        \midrule
        \multicolumn{10}{c}{TinyLlama-1.1b-3T
        Static Depth - Two Phase} \\
        \midrule
         & 62.5 & 36.5 & 60.3 & 59.6 & 44.4 & 72.9 & 39.4 & 45.2 & 12.8 \\
         \bottomrule
    \end{tabular}
    \label{tab:data-mix}
\end{table}

\section{Discussion}~\label{sec:discuss}
Our work demonstrates that depth-recurrent language models are parameter efficient and highlights their flexibility in decoupling train-time and test-time compute.  However, investing more FLOPs per parameter makes the training process for depth-recurrent models more expensive. Our work makes significant progress towards ameliorating this issue by leveraging pretrained initializations, recurrence scheduling during training, and a data curriculum.

\looseness -1
Here we identify several promising avenues for future work.
One unsolved problem is how to most effectively build depth-recurrent models that can recur deeper at test time to solve harder problems than were seen during training. A related goal is how to imbue recurrent models with native adaptivity that automatically assigns the right amount of compute (recurrence) to a given problem based on how difficult it is. 
Such built-in stopping criteria would in principle allow models to think deeply on hard problems while solving easy problems quickly.
\Cref{fig:which-layers} and \Cref{fig:which-layers-shortgpt} present our search process on selecting which layers to keep and which ones to discard, but future work could identify a more optimal method for layer choice during model surgery.
While our experiments are at the 1B parameter and 50B token scales, more experimentation is required to verify that our method generalizes to much larger model and data scales.
Finally, we primarily focus on strengthening a model's mathematical capabilities via depth-recurrence, and future work should extend this to other reasoning-intensive domains.

\section*{Acknowledgments}
This work was made possible by DARPA TIAMAT and the NSF TRAILS Institute (2229885). Additional support was provided by awards from Capital One Bank, Open Philanthropy, and the Center For AI and Responsible Financial Innovation.

Prepared by LLNL under Contract DE-AC52-07NA27344 and supported by the LLNL-LDRD Program under Project No. 24-ERD-010 (LLNL-CONF-2012308). This manuscript has been authored by Lawrence Livermore National Security, LLC under Contract No. DE-AC52-07NA27344 with the U.S. Department of Energy. The United States Government retains, and the publisher, by accepting the article for publication, acknowledges that the United States Government retains a non-exclusive, paid-up, irrevocable, world-wide license to publish or reproduce the published form of this manuscript, or allow others to do so, for United States Government purposes.

\newpage
\bibliography{refs}
\bibliographystyle{conf/iclr2026_conference}

\appendix
\section*{Appendix}

\section{Extended Related Works}~\label{app-sec:app-rel-work}
The field of methods for leveraging adaptive test-time computation with architectural modifications \citep[e.g.][]{von2025mesanet} and additional training methodologies \citep[e.g.][]{guo2025deepseek} is vast and we refer the reader to \citet{zhu2025survey} for a detailed survey.

Recurrent models have been a cornerstone of machine learning for many years \citep{amari1972learning,hopfield1982neural,gers2000recurrent,sutskever2008recurrent}.
depth-recurrent architectures can all be viewed as learning the gradient of an energy based model \citep{lecun2005loss}. \citet{gladstone2025energy} show energy based models can be scaled effectively.
Recurrent mechanisms are shown to learn generalizable solutions to problems using ResNet \citep{he2015deep} based architectures \citep{schwarzschild2021can,bansal2022end,anil2022path,schwarzschild2023deep,bear2024rethinking}. 

\citet{yang2023looped,giannou2023looped,gatmiry2024can} and \citet{fan2024looped} study the potential theoretical benefits of recurrence at small scales.
Many works study the impact of depth for transformers both theoretically and practically \citep{levine2020depth,merrill2022saturated,mcleish2025gemstones,zuo2025falcon,merrill2025little,csordas2025language}, it is still an open question how recurrent depth impacts the performance of transformers.
\citet{saunshi2025reasoning} demonstrate the power of recurrence by showing chain of thought \citep{wei2022chain} steps can be implicitly simulated in latent space using recurrence.
Similar to latent thinking is continuous chain of thought \citep{hao2024training}, a finetuning method to add recurrent behavior to pretrained language models, but training is limited as it requires sequential computations.

Prior work on model surgery has heavily studied converting pretrained transformer language models into linear complexity attention models \citep{kasai2021finetuning,zhang2024lolcat,mercat2024linearizing,wang2024mamba}.

\section{Additional Technical Details}~\label{app-sec:app-details}
\paragraph{Optimization}
Similarly to \citet{geiping2025scaling}, we train all models with truncated backpropagation \citep{williams1990efficient,mikolov2011extensions}, only recording gradients for at most the last \(8\) uses of the recurrent block.
We train in \texttt{bfloat16} mixed precision \citep{zamirai2020revisiting}, with Flash Attention \citep{dao2023flashattention} and compile the model when training.
Notably, to compile the model at scale we observe repeating the prebuilt inductor cache on each individual node removes deadlock errors and improves speed.
We train all models on AMD MI300A accelerators \citep{amd_amd_2023}, using distributed data parallel training.
We use a warmup-stable-decay learning rate scheduler \citep{zhai2022scaling,geiping2023cramming}, adjusting the warmup and decay periods to be appropriate for each experiment.
We optimize with the official implementation of Muon\footnote{\url{https://github.com/KellerJordan/Muon}}.
Muon shards the Newton-Schulz calculations between all accelerators and then communicates them, overcoming some of the efficiency degradations compared to Adam.
Combined with the fact that the models we are optimizing are smaller language models, we do not observe a degradation in step time when using Muon.

\section{Additional Experiments}
\subsection{Model Surgery Ablations}~\label{app-subsec:model-surgery}
\begin{figure}[ht!]
    \centering
    \includegraphics[width=\linewidth]{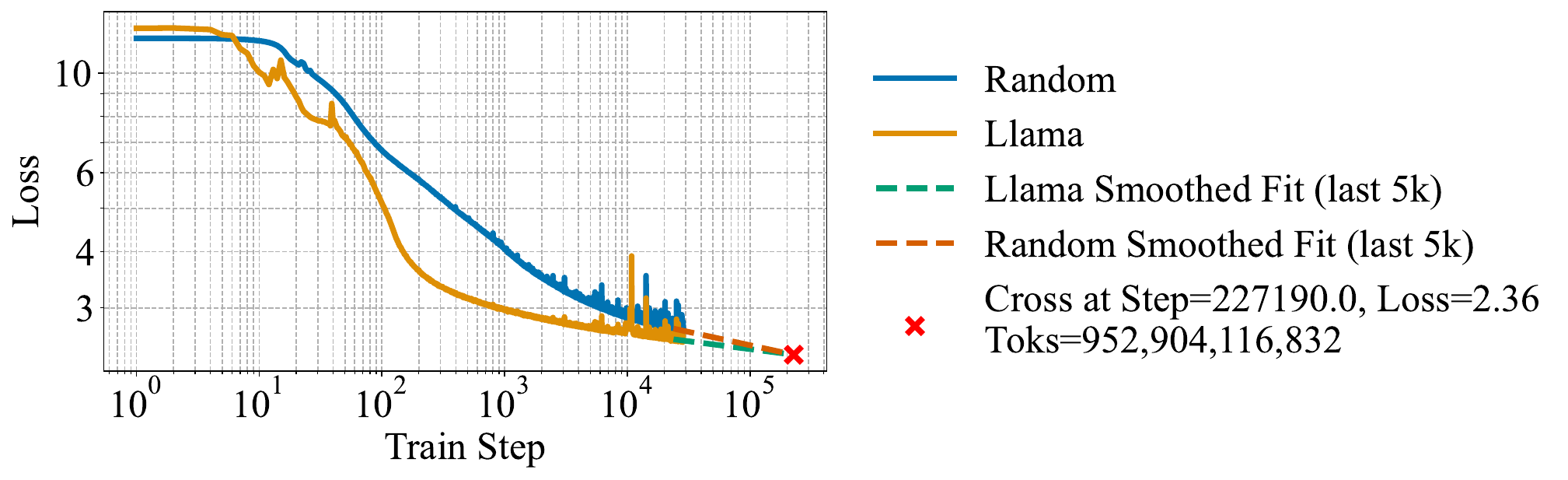}
    \caption{\textbf{Training loss for models initialized from \llama layers and Randomly.} Here we extend \Cref{fig:long-runs} including the linear extrapolations in log-log space. We note this is more than likely an underestimate of the point of intersection as there is still curvature in the loss curves.}
    \label{app-fig:long-runs-with-lines}
\end{figure}

\begin{figure}[ht!]
    \centering
    \includegraphics[width=\linewidth]{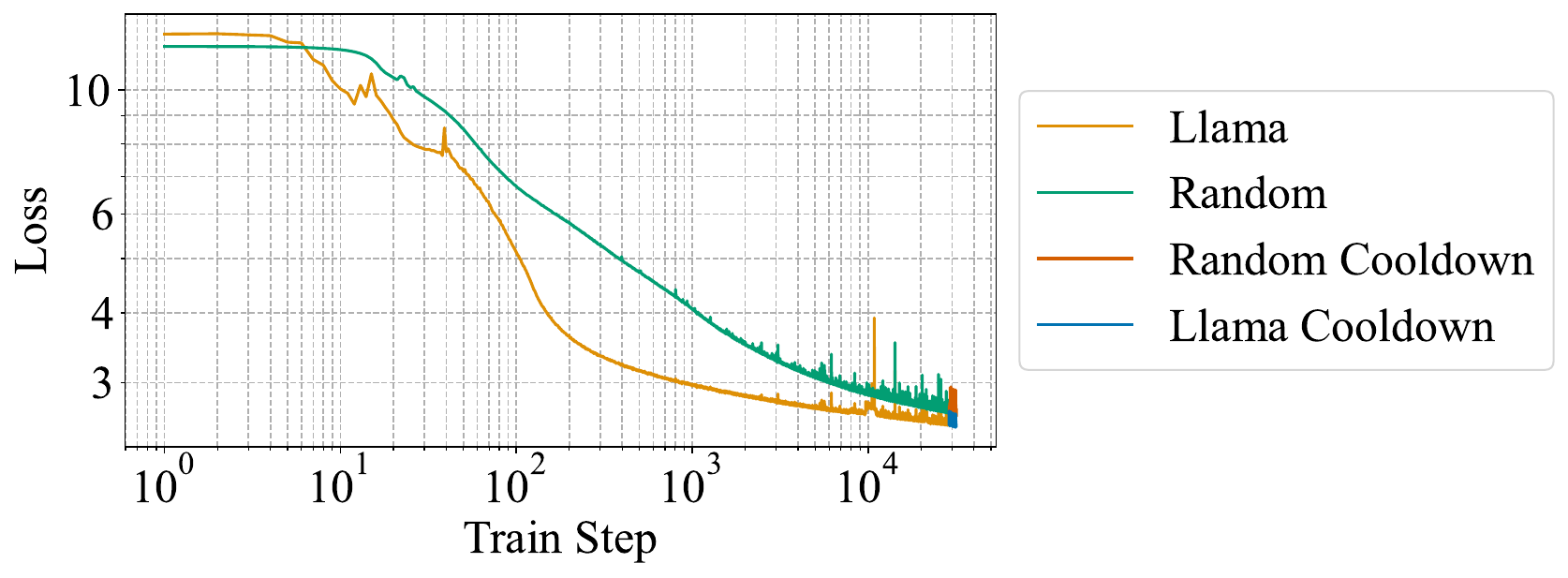}
    \caption{\textbf{Training loss for models initialized from \llama layers and Randomly.} Here, we extend \Cref{fig:long-runs} by including a cooldown for \(12b\) additional tokens, taking this to a total of \(132b\) tokens.}
    \label{app-fig:long-runs-with-cooldown}
\end{figure}

\begin{figure}[ht!]
    \centering
    \includegraphics[width=\linewidth]{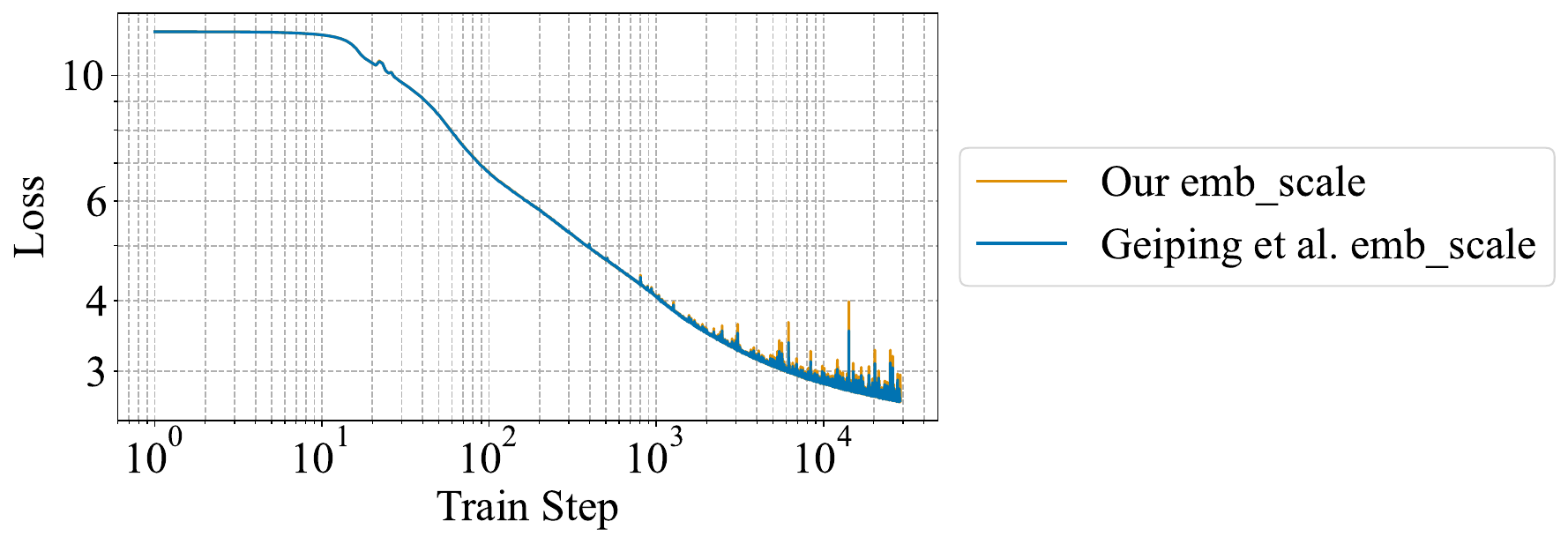}
    \caption{\textbf{Training loss for models initialized Randomly with different embedding scales over \(120\) billion tokens.} We follow \citet{geiping2025scaling} when initializing models with scaled embeddings. However, we also ablate how much the scale impacts by using the same embedding scale from Huginn-0125 in this much smaller model, we find there to be minimal impact.}
    \label{app-fig:long-runs-with-mp}
\end{figure}

In \Cref{app-fig:long-runs-with-lines}, we perform a linear extrapolation of the loss curves shown in \Cref{fig:long-runs}, seeing the extrapolations intersect at approximately \(950\) billion tokens.
We note this is more than likely an underestimate as there is still curvature in the loss curves.
In \Cref{app-fig:long-runs-with-cooldown}, we continue training the models from \Cref{fig:long-runs}, cooling the learning rate down over an additional \(12\) billion tokens.
In \Cref{app-fig:long-runs-with-mp}, we vary the \texttt{emb\_scale} hyperparameter used by \citet{geiping2025scaling}. ``Ours'' is using the \texttt{emb\_scale} from the \texttt{Huginn-0125} model, where as the line for \citet{geiping2025scaling} has been adjusted for this specific model shape.
We see a negligible difference.
In \Cref{app-tab:random_init_benchmarks_full}, we extend \Cref{fig:long-runs} with additional test recurrences for other language modelling tasks.

\begin{table}[ht!]
    \centering
    \caption{\textbf{Initializing from pretrained model weights yields consistent gains across benchmarks.} We evaluate our models trained for \(120\) billion tokens in a zero shot setting seeing clear advantages to initializing from pretrained weights.}
    \begin{tabular}{cccccccc}
        \toprule
         Test Recurrence & Arc-E & Arc-C & HS & WG & MMLU & PIQA & OBQA \\
         \midrule
         \multicolumn{8}{c}{Random} \\
         \midrule
          & 25 & 25 & 25 & 50 & 25 & 50 & 25 \\
         \midrule
         \multicolumn{8}{c}{Takase init} \\
         \midrule
         1 & 36.1 & 23.4 & 28.6 & 50.5 & 22.9 & 55.2 & 26.6 \\
         2 & 41.2 & 22.4 & 31.6 & 50.2 & 23.0 & 58.3 & 28.4 \\
         4 & 50.7 & 26.7 & 37.8 & 48.8 & 23.4 & 63.4 & 31.2 \\
         8 & 54.5 & 29.4 & 45.4 & 53.3 & 24.4 & 67.9 & 35.8 \\
         16 & 55.8 & 30.0 & 47.8 & 53.7 & 24.8 & 68.7 & 36.6 \\
         32 & 56.1 & 29.5 & 48.3 & 54.3 & 25.0 & 68.9 & 36.8 \\
         \midrule
         \multicolumn{8}{c}{Llama init} \\
         \midrule
         1 & 41.6 & 23.8 & 34.8 & 51.3 & 22.9 & 62.5 & 27.2 \\
         2 & 48.4 & 26.6 & 41.2 & 51.4 & 23.2 & 65.9 & 30.6 \\
         4 & 54.5 & 30.8 & 49.4 & 53.2 & 24.0 & 69.7 & 35.4 \\
         8 & 59.2 & 34.0 & 54.9 & 55.6 & \textbf{25.4} & 72.3 & 38.4 \\
         16 & 60.2 & \textbf{35.1} & 55.4 & 55.7 & 25.3 & \textbf{73.1} & 38.4 \\
         32 & \textbf{60.4} & 35.0 & \textbf{55.6} & \textbf{56.1} & 25.3 & 72.9 & \textbf{38.6} \\
         \bottomrule
    \end{tabular}
    \label{app-tab:random_init_benchmarks_full}
\end{table}

\FloatBarrier
\subsubsection{Which Layers to take?}
\begin{figure}[ht!]
    \centering
    \includegraphics[width=\linewidth]{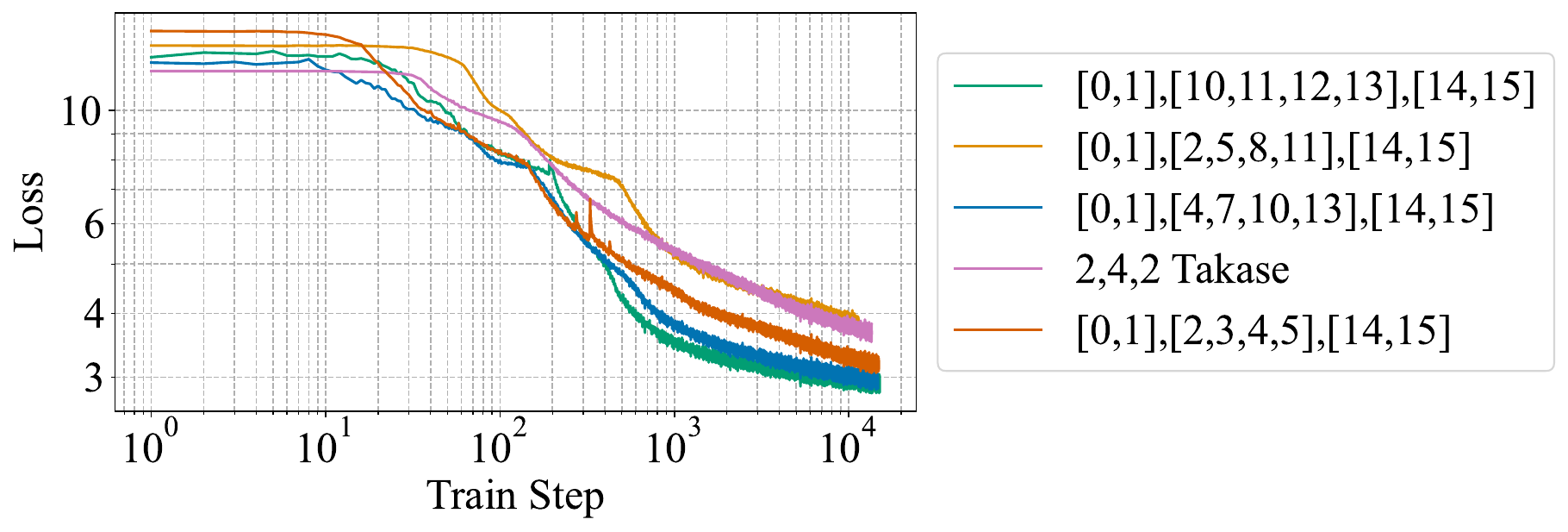}
    \caption{\textbf{We ablate which layers to select from \llama-1b.} We measure the training loss on Fineweb-Edu with different layer selections from \llama-1b. We find taking early layers for the prelude, and later layers for the recurrent block and coda to be best.}
    \label{fig:which-layers}
\end{figure}

\begin{figure}[ht!]
    \centering
    \includegraphics[width=\linewidth]{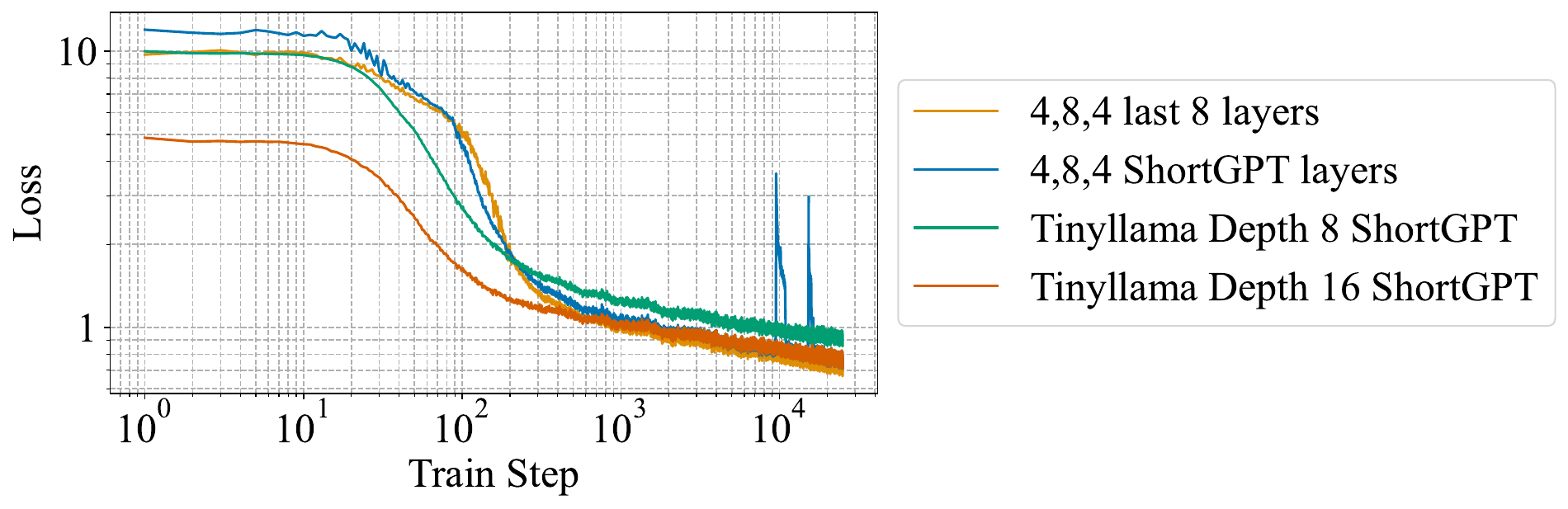}
    \caption{\textbf{Comparison to prior methods for decreasing depth.} We use the ShortGPT pruning method proposed by \citet{men2024shortgpt} to decrease the depth of the \tinyllama model. We train two non-recurrent models with this pruning method, reducing \tinyllama's depth to \(8\) and \(16\). We also train a \((4,8,4)\) model using our layer selection (See \Cref{app-tab:recur_layers} and a model using the layers prescribed by ShortGPT. We train on the nemotron dataset for approximately \(25\) billion tokens and find our layer selection to be better in terms of loss.}
    \label{fig:which-layers-shortgpt}
\end{figure}

\begin{figure}[ht!]
    \centering
    \includegraphics[width=\linewidth]{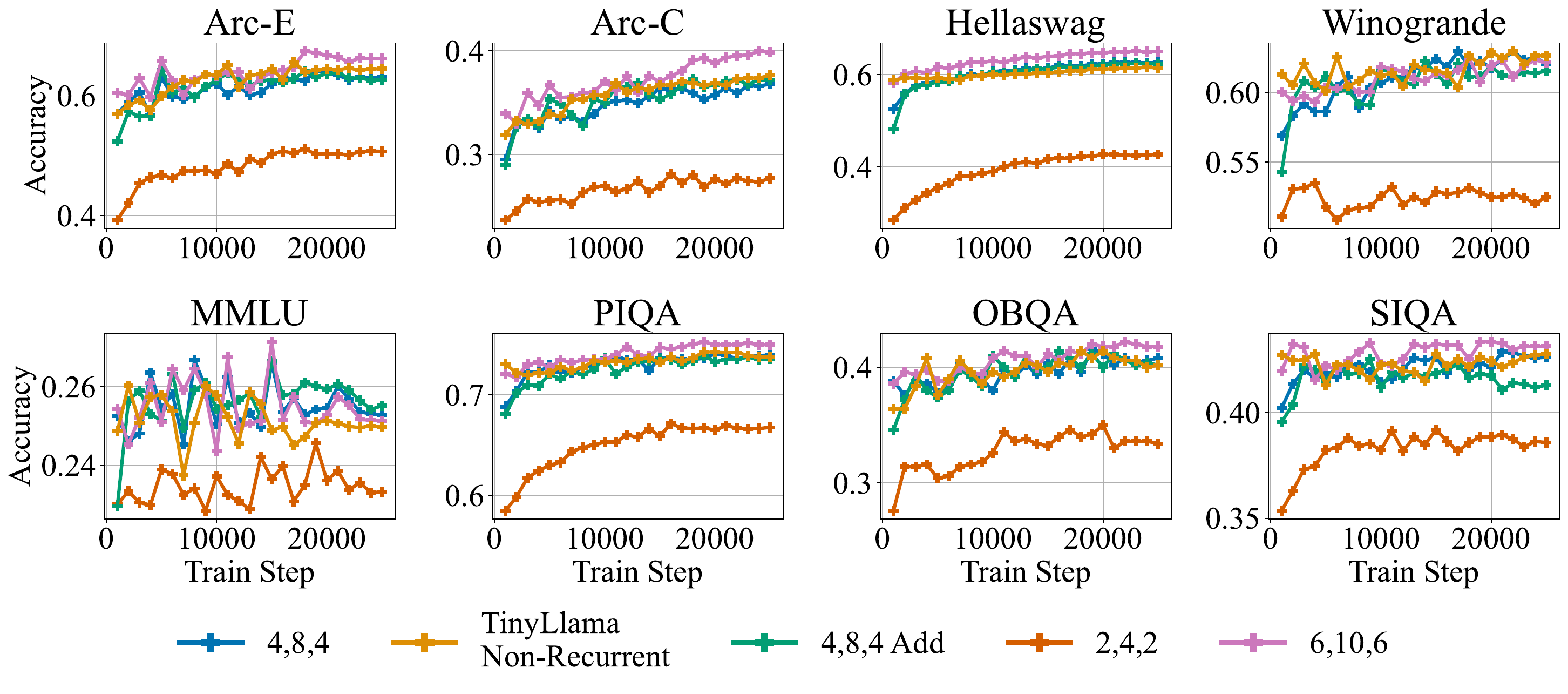}
    \caption{\textbf{We ablate different layer selections and architectural choices for \tinyllama.} We show the accuracy on evaluations at \(32\) recurrences after training on Fineweb-Edu. We ablate \((2,4,2)\), \((4,8,4)\) and \((6,10,6)\) models with \((6,10,6)\) keeping all of the layers of the depth \(22\) \tinyllama model. It is clear to see that increasing the number of layers in the recurrent block allows for the model to achieve higher accuracy, consistently beating the fixed depth model. However, having a larger recurrent block does significantly increase the FLOPs used by the model. We also ablate swapping the linear adapter used by \citet{geiping2025scaling} and in our main results for an addition adapter (``Add''). We find that although training loss is higher the evaluation accuracy is approximately the same.}
    \label{fig:which-layers-all-layers}
\end{figure}

\begin{figure}[ht!]
    \centering
    \includegraphics[width=0.49\linewidth]{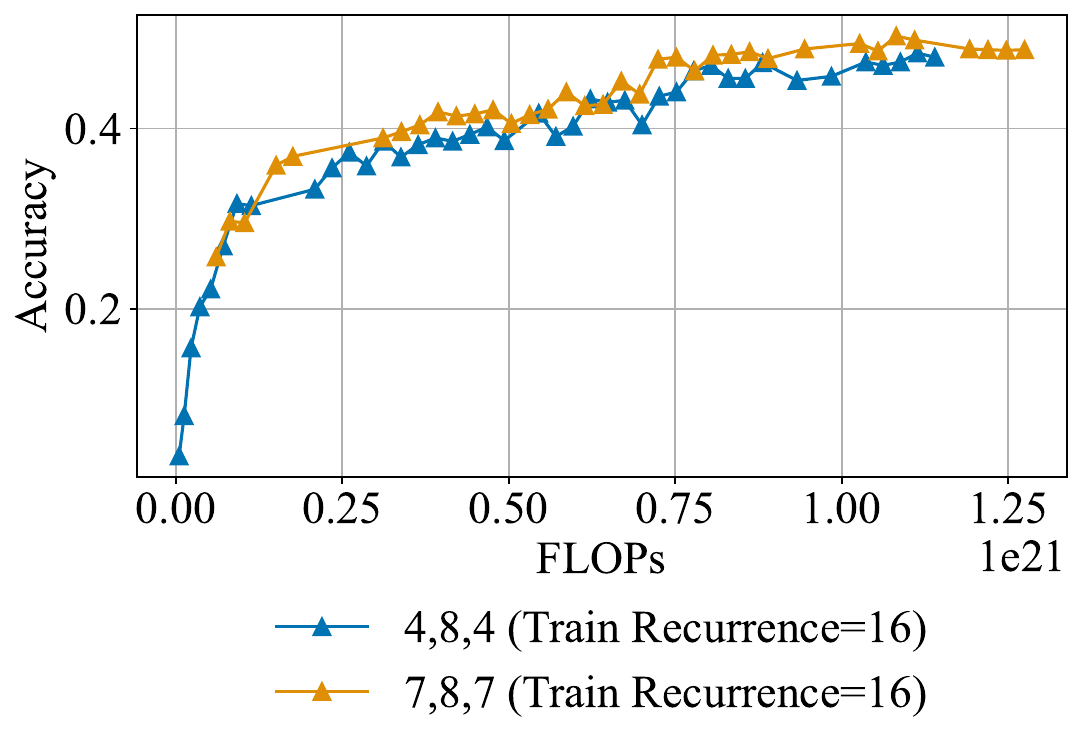}
    \includegraphics[width=0.49\linewidth]{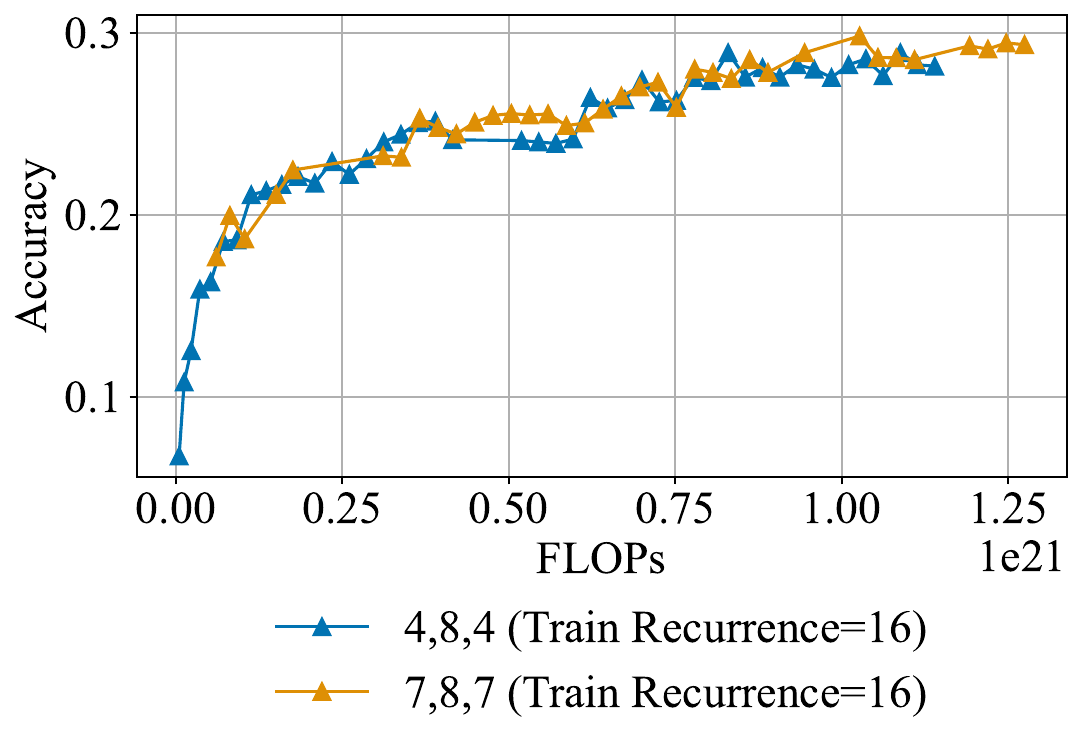}
    \caption{\textbf{Extending the prelude and coda leads to minimal performance improvements.} We train \((4,8,4)\) and \((7,8,7)\) \tinyllama models with a \(25\%\) linear curriculum. We see that adding an additional \(6\) layers leads to minimal final performance improvement for the additional FLOPs used. \textbf{Left: }Performance over training FLOPs on GSM8K. \textbf{Right: }Performance over training FLOPs on MATH.}
    \label{app-fig:extend-prelude-coda}
\end{figure}

In \Cref{fig:which-layers}, we perform a small search over which layers to select when forming a recurrent model and removing layers.
In \Cref{fig:which-layers-shortgpt}, we compare the layers we found to be optimal to dropping the least impactful layers using the ShortGPT method \citep{men2024shortgpt}.
We find for training depth-recurrent models our selection is better.
In \Cref{fig:which-layers-all-layers}, we show additional results for models with more varied shapes.
In \Cref{app-fig:extend-prelude-coda}, we extend the prelude and coda to leverage all layers of the parent model, this yields only negligible improvement for the increased FLOPs.

\FloatBarrier
\subsection{Scheduling Recurrences Ablations}~\label{app-subsec:schedule-rec}
\begin{figure}[ht!]
    \centering
    \includegraphics[width=0.6\linewidth]{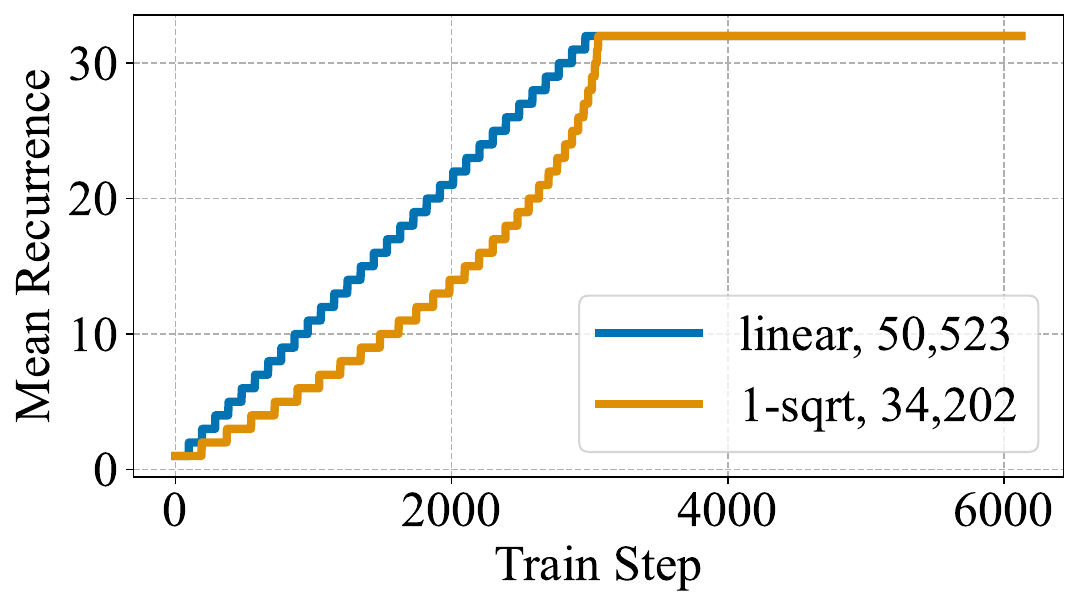}
    \caption{\textbf{Visualization of our curriculum over training steps.} We visualize a curriculum with \(3125\) steps over a training period of \(6250\) steps with a final mean recurrence of \(32\). We show both a \textit{linear} and \textit{\sqrtsched} schedules.\\
    \(f_{\text{linear}}(\text{tgt\_depth},\text{current\_step}) = ceil(\text{tgt\_depth} * (\text{current\_step} / \text{num\_warmup\_steps}))\)\\
    \(f_{\text{1-sqrt}}(\text{tgt\_depth},\text{current\_step}) = ceil(\text{tgt\_depth} * (1- sqrt(1- \text{current\_step} / \text{num\_warmup\_steps})))\)\\
    In the legend we include the number of recurrences used during the curriculum period, seeing the \sqrtsched schedule uses fewer recurrences.}
    \label{app-fig:schedule-mean-explainer}
\end{figure}

\begin{figure}[ht!]
    \centering
    \includegraphics[width=\linewidth]{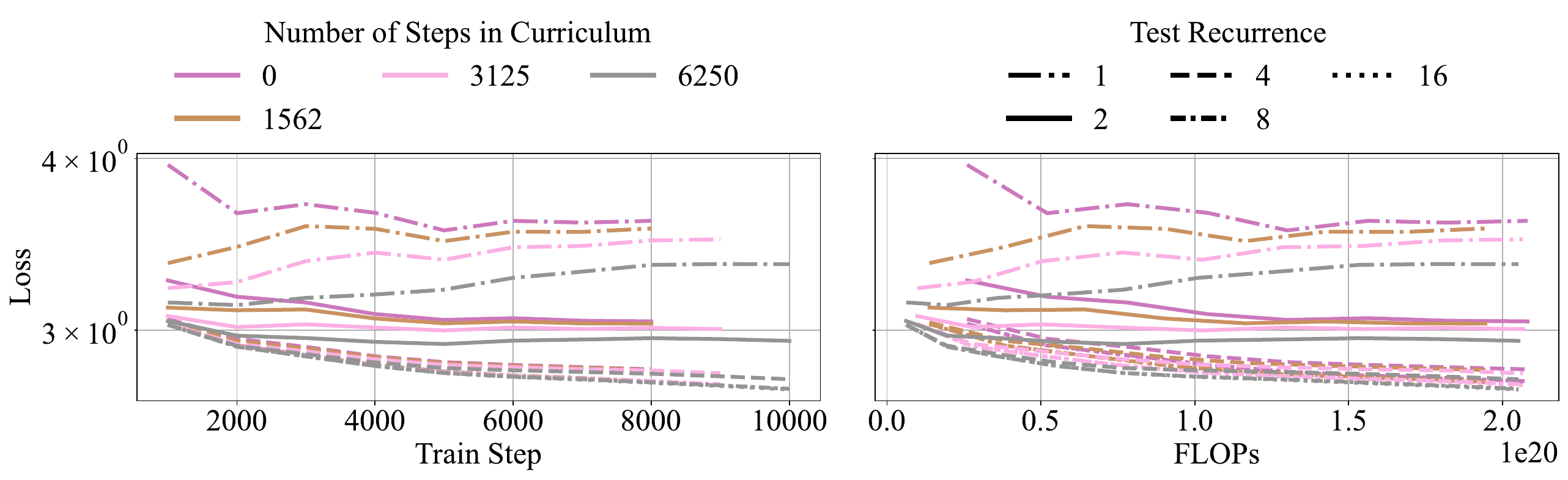}
    \caption{\textbf{Scheduling the mean of the depth distribution is efficient in terms of data and compute.} We extend \Cref{fig:schedule-mean-n}, showing more curriculum lengths on the left and more test recurrences on the right. We see the same as in \Cref{fig:schedule-mean-n}, that it is efficient in terms of data (steps) and compute to schedule the mean of the depth distribution.}
    \label{app-fig:schedule-mean-n-all}
\end{figure}

\begin{figure}[ht!]
    \centering
    \includegraphics[width=\linewidth]{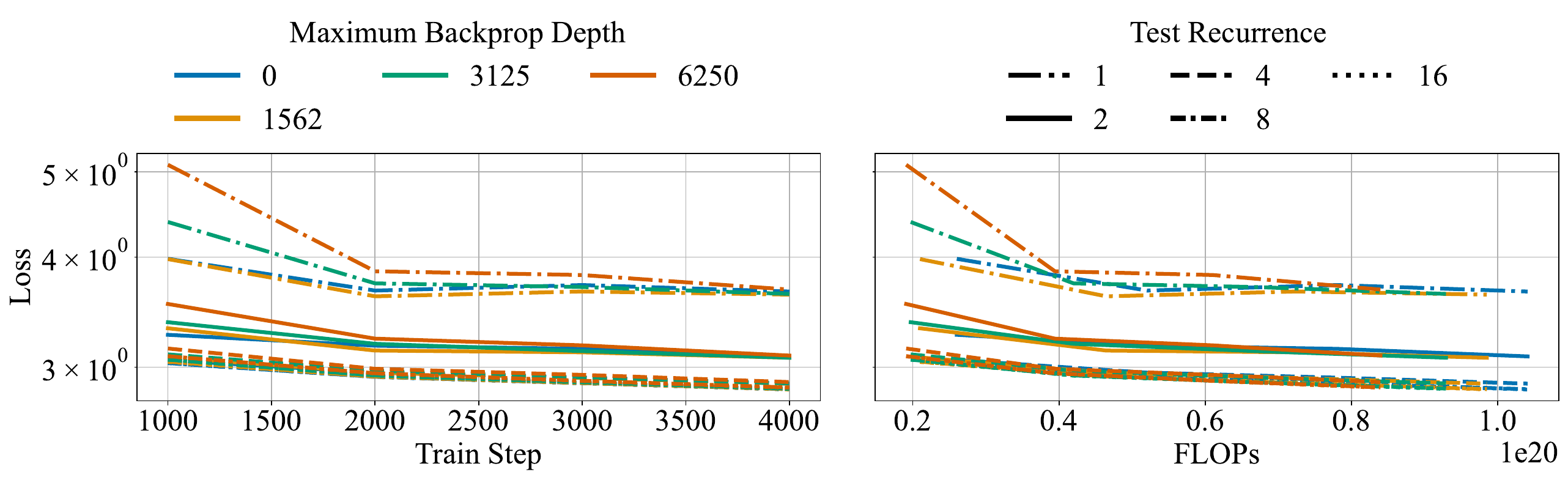}
    \caption{\textbf{Validation loss of models with schedules maximum backpropagation depth.} We see that scheduling the maximum backpropagation is efficient in terms of FLOPs spent but does lead to worse models in terms of steps.}
    \label{app-fig:schedule-mean-k}
\end{figure}

In \Cref{app-fig:schedule-mean-explainer}, we visually show the values our curriculum takes, looking like a staircase from \(1\) to the maximum value over the curriculum period.
In \Cref{app-fig:schedule-mean-n-all}, we extend \Cref{fig:schedule-mean-n}, showing more curriculum lengths and more test recurrences.
In \Cref{app-fig:schedule-mean-k}, we show the result of scheduling the backpropagation depth over training.

\begin{figure}[ht!]
    \centering
    \includegraphics[width=0.49\linewidth]{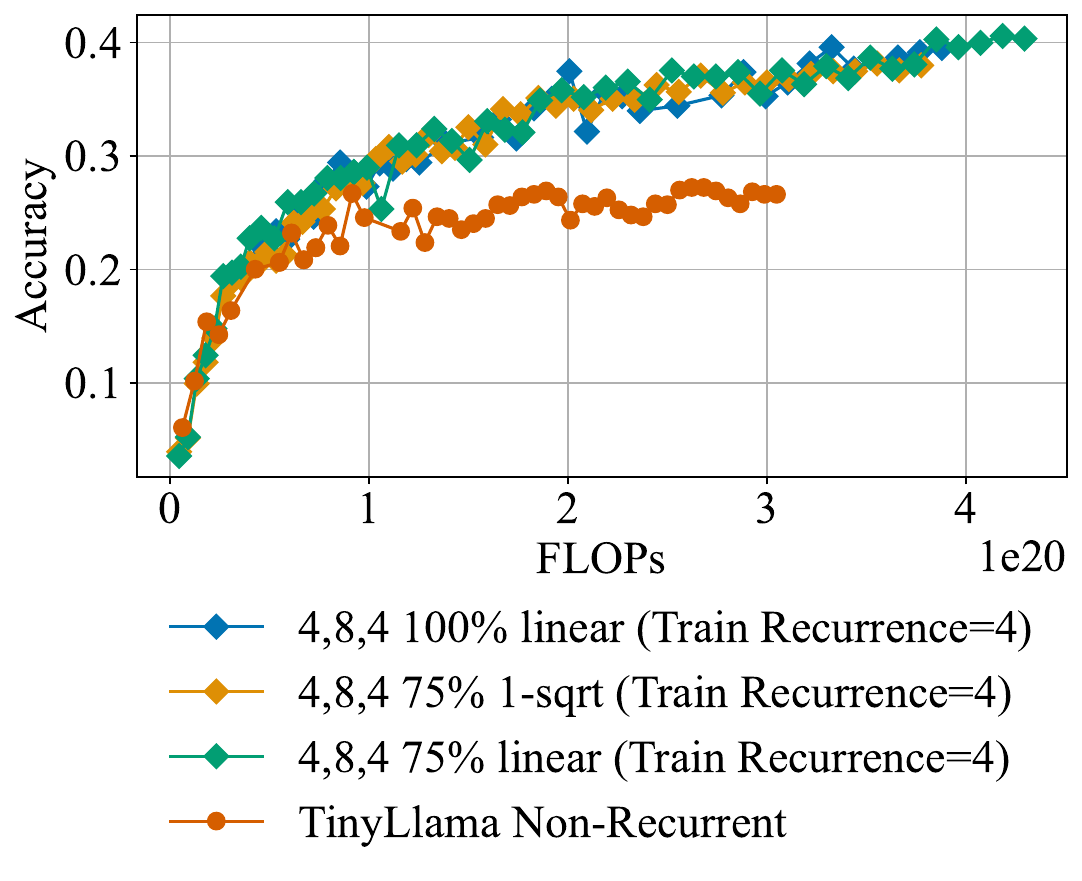}
    \includegraphics[width=0.49\linewidth]{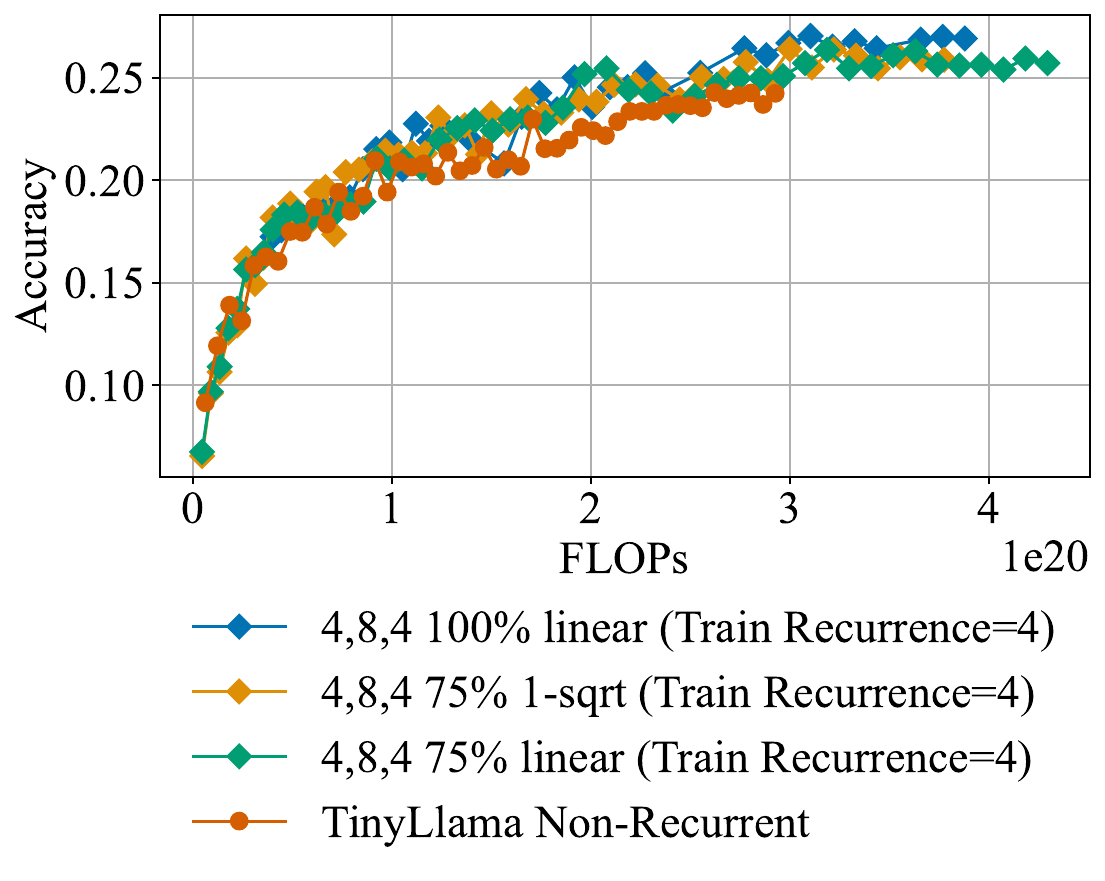}
    \caption{\textbf{\sqrtsched vs. linear curricula for \tinyllama}. We see there is little separating the curricula on a per FLOP basis and therefor choose \(75 \%\) \sqrtsched for our experiments in Figures \ref{fig:tinyllama-GSM8K} and \ref{fig:olmo-MATH} as it efficient while spending \(25\%\) of training at the maximum mean number of recurrences. \textbf{Left: }We plot accuracy over FLOPs for GSM8K. \textbf{Right: }We plot accuracy over FLOPs for MATH.}
    \label{app-fig:1-sqrt-sched}
\end{figure}

\FloatBarrier
\subsection{How to Retrofit Recurrence Ablations}~\label{app-subsec:retrofit}
\subsubsection{\tinyllama}~\label{app-subsubsec:retrofit-tinyllama}
\begin{figure}[ht!]
    \centering
    \includegraphics[width=\linewidth]{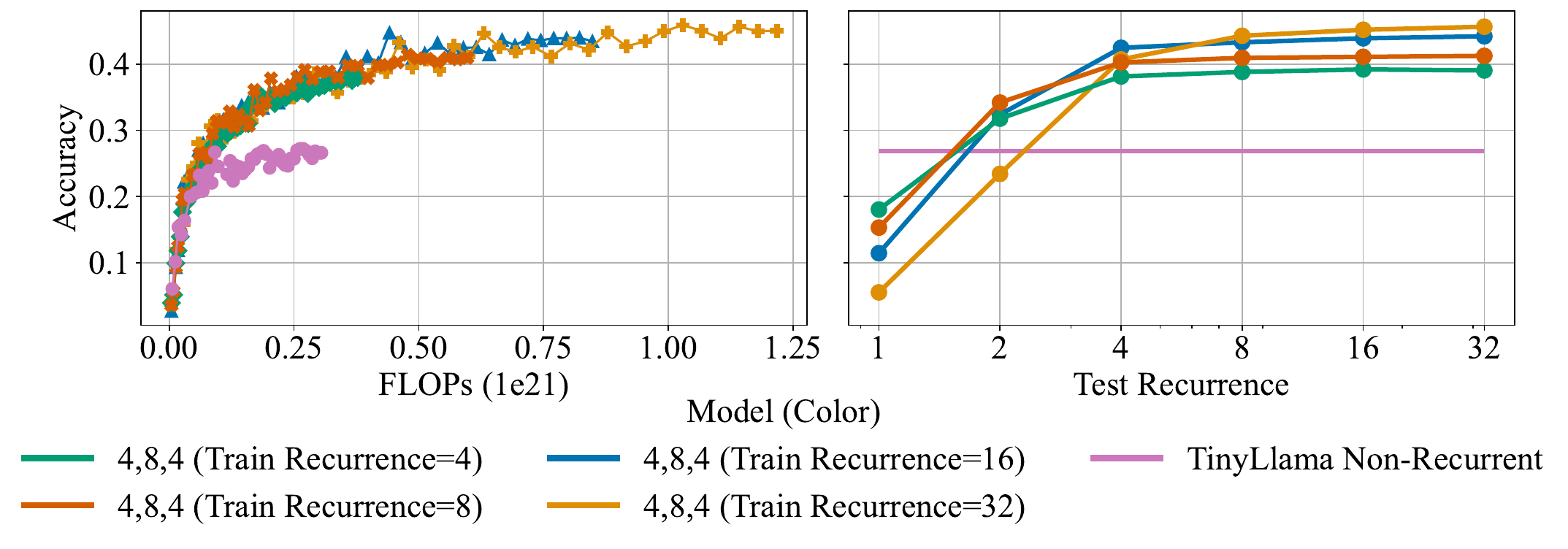}
    \caption{\textbf{Recurrence efficiently improves reasoning on GSM8K for \tinyllama.}
    We train \((4,8,4)\) and non-recurrent models for approximately \(50\) billion tokens of Nemotron-CC-Math-v1 data, extending \Cref{fig:tinyllama-GSM8K}.
    \textbf{Left: }We plot accuracy over the number of FLOPs used during training. We see that recurrent models can efficiently outperform the non-recurrent baseline.
    \textbf{Right: } We plot accuracy over the number of recurrences for inference. We see the recurrent models are competitive with the fixed depth baseline and can outperform it by using more FLOPs.\\ We plot each individual models accuracy over training and recurrence in full in \Cref{app-fig:tinyllama-GSM8K-4-8} and \Cref{app-fig:tinyllama-GSM8K-16-32}. Evaluations on the final checkpoint over tasks shown in \Cref{tab:data-mix} are in Appendix \Cref{app-tab:tinyllama-all-evals}.}
    \label{app-fig:tinyllama-GSM8K-all}
\end{figure}

\begin{figure}[ht!]
    \centering
    \includegraphics[width=0.7\linewidth]{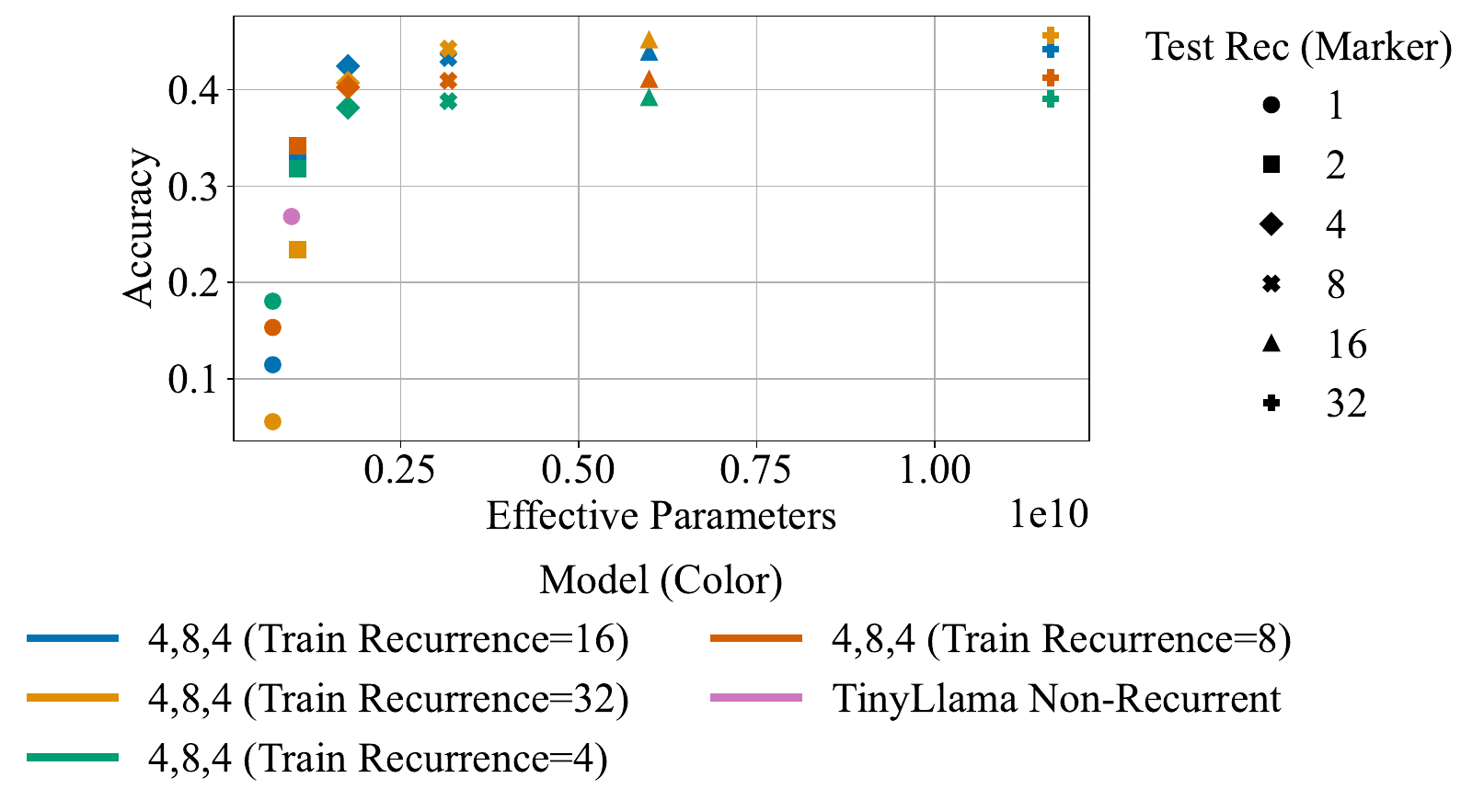}
    \caption{\textbf{Recurrent models are competitive in terms of inference FLOPs for GSM8K.} This is the same data as in Right of Figures \ref{fig:tinyllama-GSM8K} and \ref{app-fig:tinyllama-GSM8K-all} but replotted with an effective parameters x-axis, which can be viewed as proportional to FLOPs required for inference.}
    \label{app-fig:tinyllama-GSM8K-effective-params}
\end{figure}

\begin{figure}[ht!]
    \centering
    \includegraphics[width=0.48\linewidth]{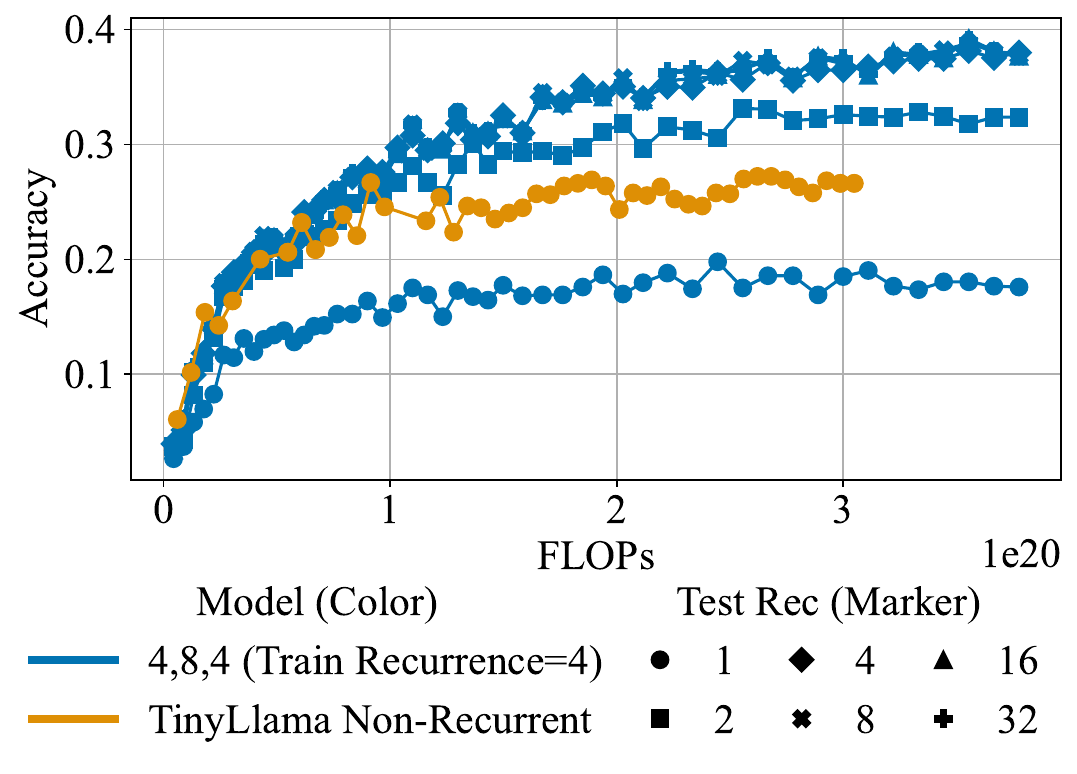}
    \includegraphics[width=0.48\linewidth]{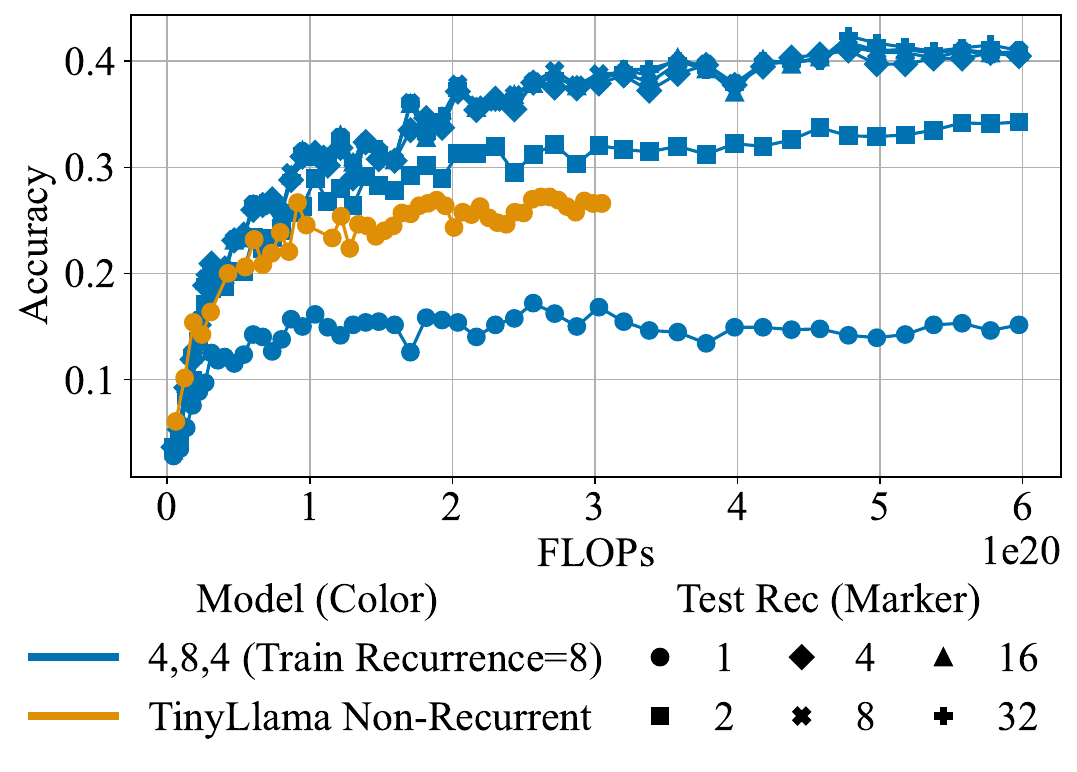}
    \caption{\textbf{Recurrence efficiently improves reasoning.} \textbf{Left}: GSM8K accuracy over training step for train recurrence equal to $4$ model. \textbf{Right}: GSM8K accuracy over training step for train recurrence equal to $8$ model.}
    \label{app-fig:tinyllama-GSM8K-4-8}
\end{figure}

\begin{figure}[ht!]
    \centering
    \includegraphics[width=0.48\linewidth]{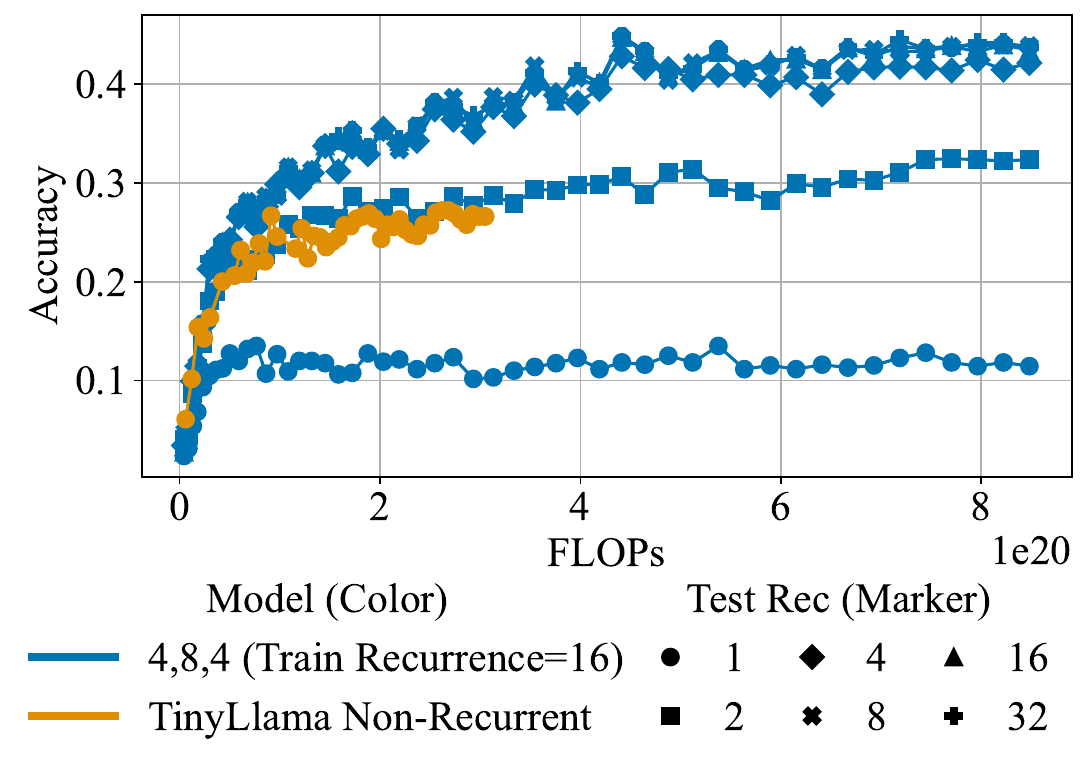}
    \includegraphics[width=0.48\linewidth]{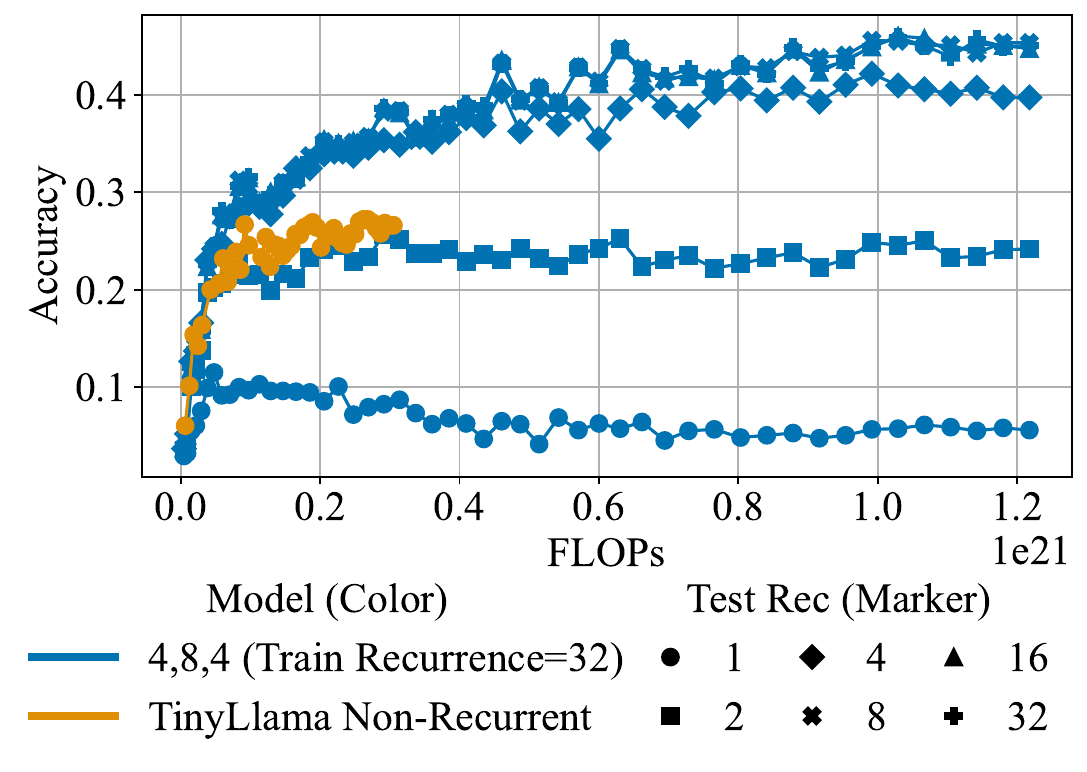}
    \caption{\textbf{Recurrence efficiently improves reasoning. }\textbf{Left}: GSM8K accuracy over training step for train recurrence equal to $16$ model. \textbf{Right}: GSM8K accuracy over training step for train recurrence equal to \(32\) model.}
    \label{app-fig:tinyllama-GSM8K-16-32}
\end{figure}

In \Cref{app-fig:tinyllama-GSM8K-all} we extend \Cref{fig:tinyllama-GSM8K}, showing more train recurrences.
In \Cref{app-fig:tinyllama-GSM8K-effective-params}, we plot Right of Figures \ref{fig:tinyllama-GSM8K} and \ref{app-fig:tinyllama-GSM8K-all} with an effective parameters x-axis, which can be viewed as proportional to FLOPs required for inference. 
In Figures \ref{app-fig:tinyllama-GSM8K-4-8} and \ref{app-fig:tinyllama-GSM8K-16-32} we show the GSM8K accuracy over training step for train recurrences \(4,8,16\) and \(32\).

\begin{figure}[ht!]
    \centering
    \includegraphics[width=\linewidth]{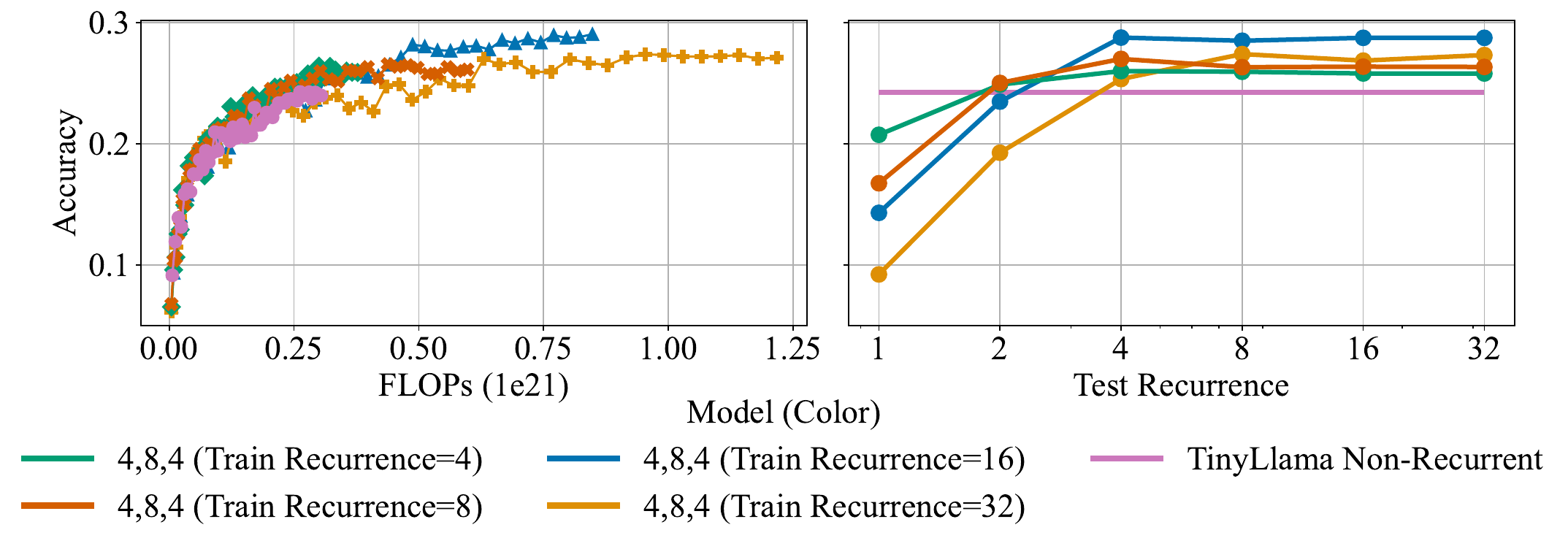}
    \caption{\textbf{Recurrence efficiently improves reasoning on MATH for \tinyllama.} \textbf{Left: }We plot accuracy over the number of FLOPs used during training. We see that recurrent models can efficiently outperform the non-recurrent baseline.
    \textbf{Right: } We plot accuracy over the number of recurrences for inference. We see the recurrent models are competitive with the fixed depth baseline and can outperform it by using more FLOPs.\\ We plot each individual models accuracy over training and recurrence in full in \Cref{app-fig:tinyllama-MATH-4-8} and \Cref{app-fig:tinyllama-MATH-16-32}. Evaluations on the final checkpoint over tasks shown in \Cref{tab:data-mix} are in Appendix \Cref{app-tab:tinyllama-all-evals}.}
    \label{app-fig:tinyllama-MATH-all}
\end{figure}

\begin{figure}[ht!]
    \centering
    \includegraphics[width=0.7\linewidth]{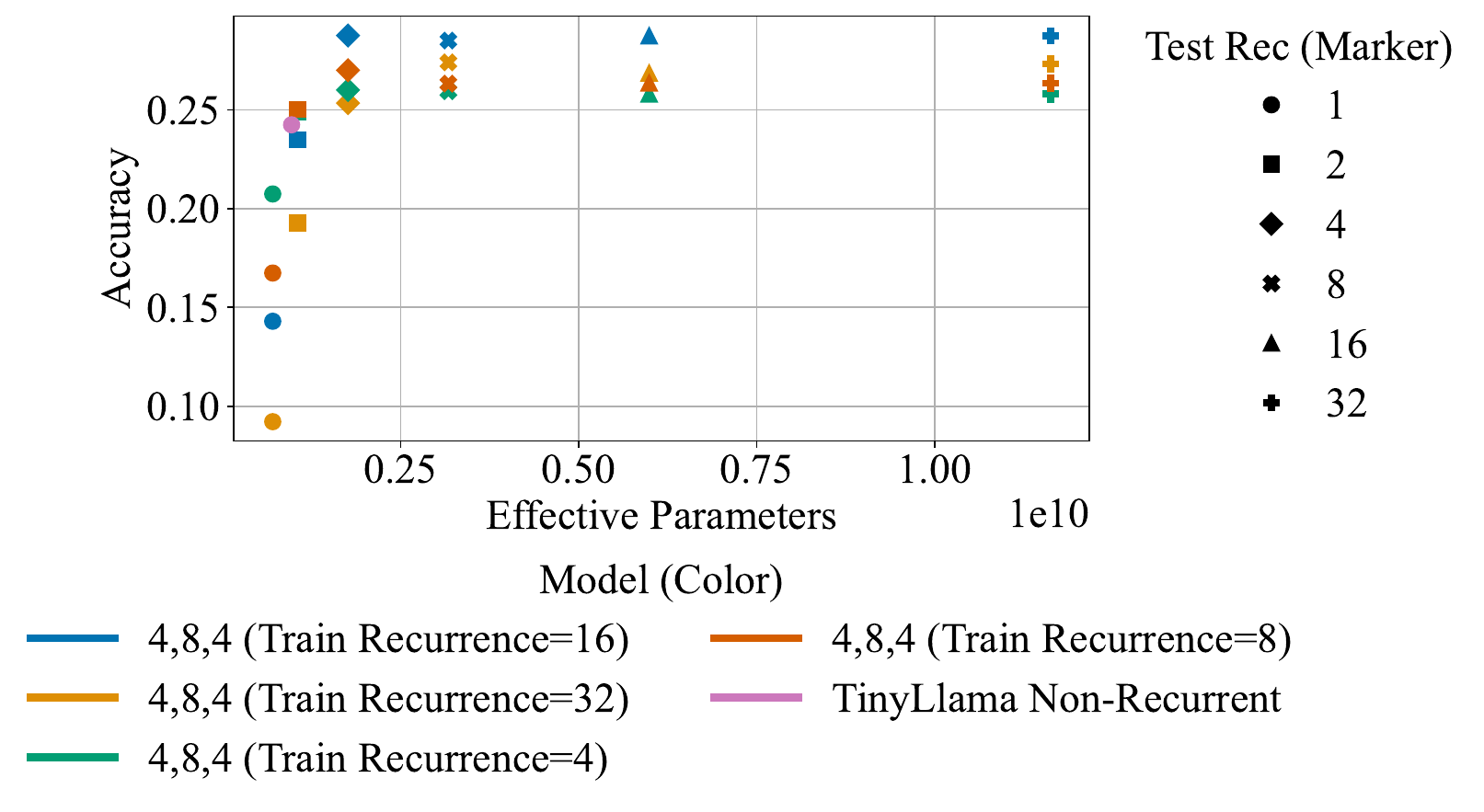}
    \caption{\textbf{Recurrent models are competitive in terms of inference FLOPs for MATH.} This is the same data as in \ref{app-fig:tinyllama-MATH-all} but replotted with an effective parameters x-axis, which can be viewed as proportional to FLOPs required for inference.}
    \label{app-fig:tinyllama-MATH-effective-params}
\end{figure}

\begin{figure}[ht!]
    \centering
    \includegraphics[width=0.48\linewidth]{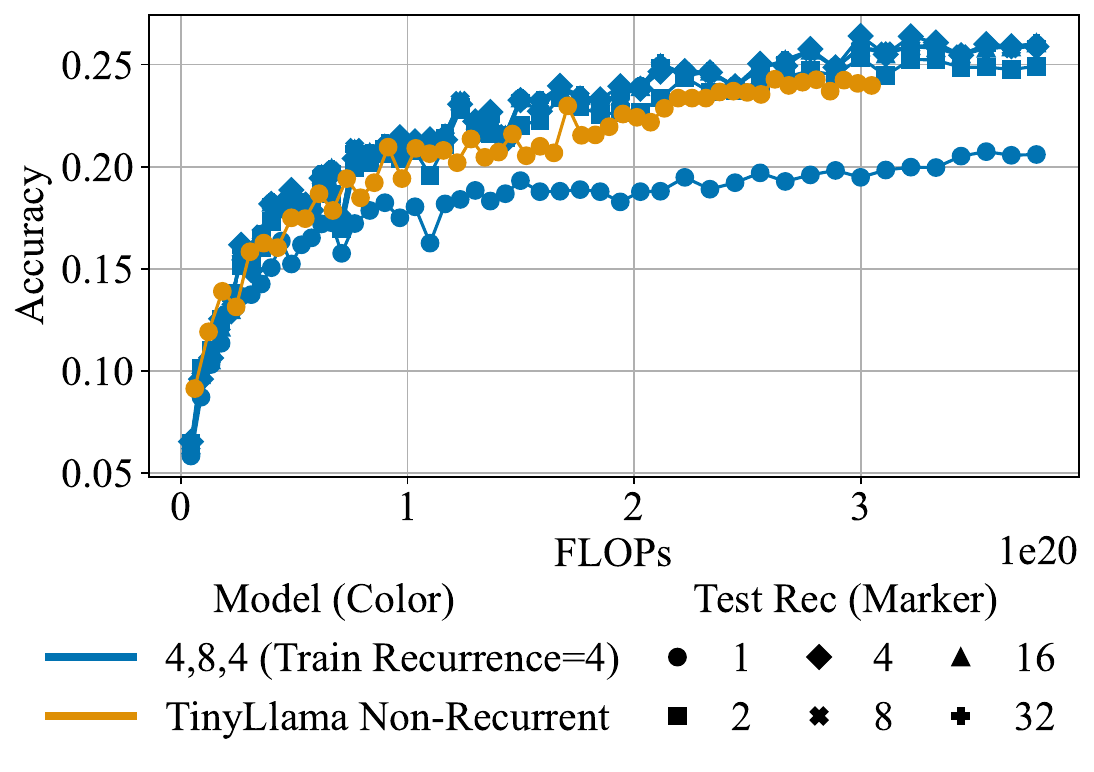}
    \includegraphics[width=0.48\linewidth]{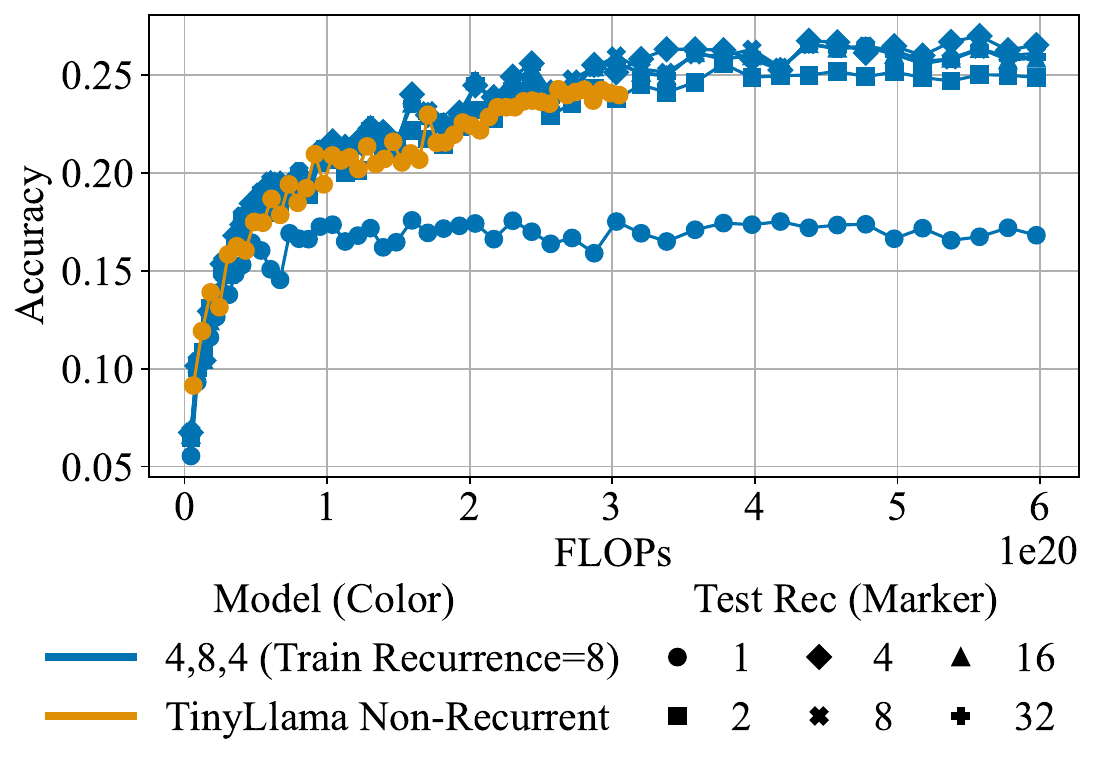}
    \caption{\textbf{Recurrence efficiently improves reasoning.} \textbf{Left}: MATH accuracy over training step for train recurrence equal to $4$ model. \textbf{Right}: MATH accuracy over training step for train recurrence equal to $8$ model.}
    \label{app-fig:tinyllama-MATH-4-8}
\end{figure}

\begin{figure}[ht!]
    \centering
    \includegraphics[width=0.48\linewidth]{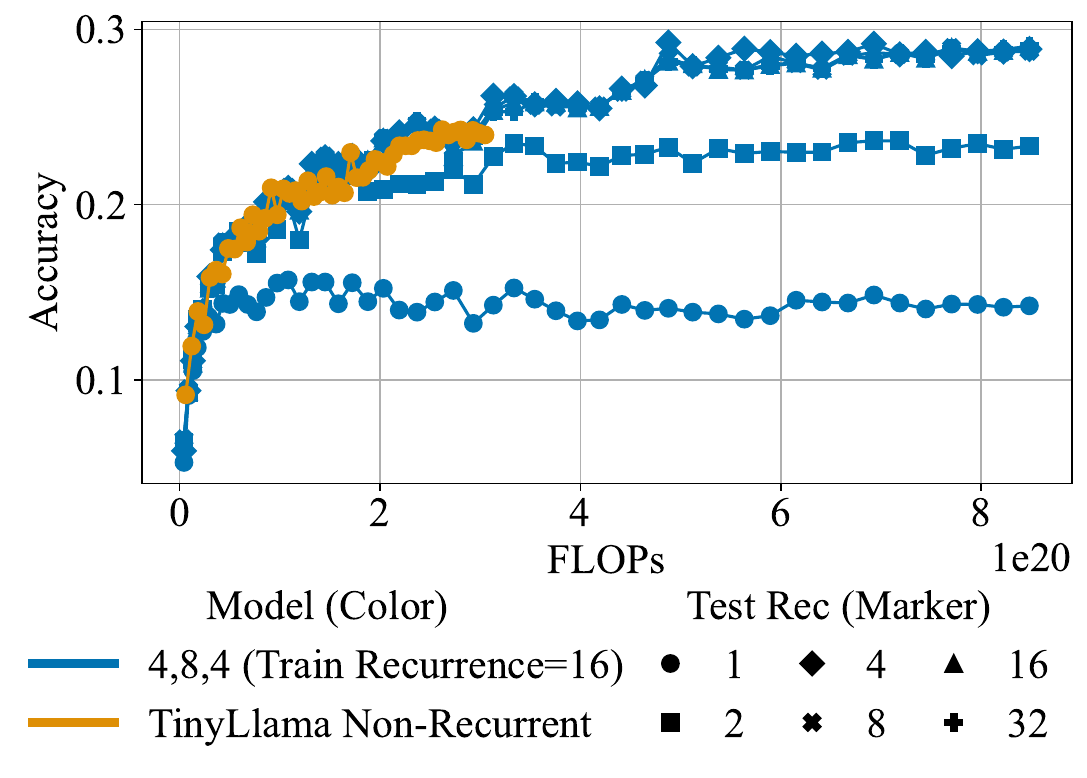}
    \includegraphics[width=0.48\linewidth]{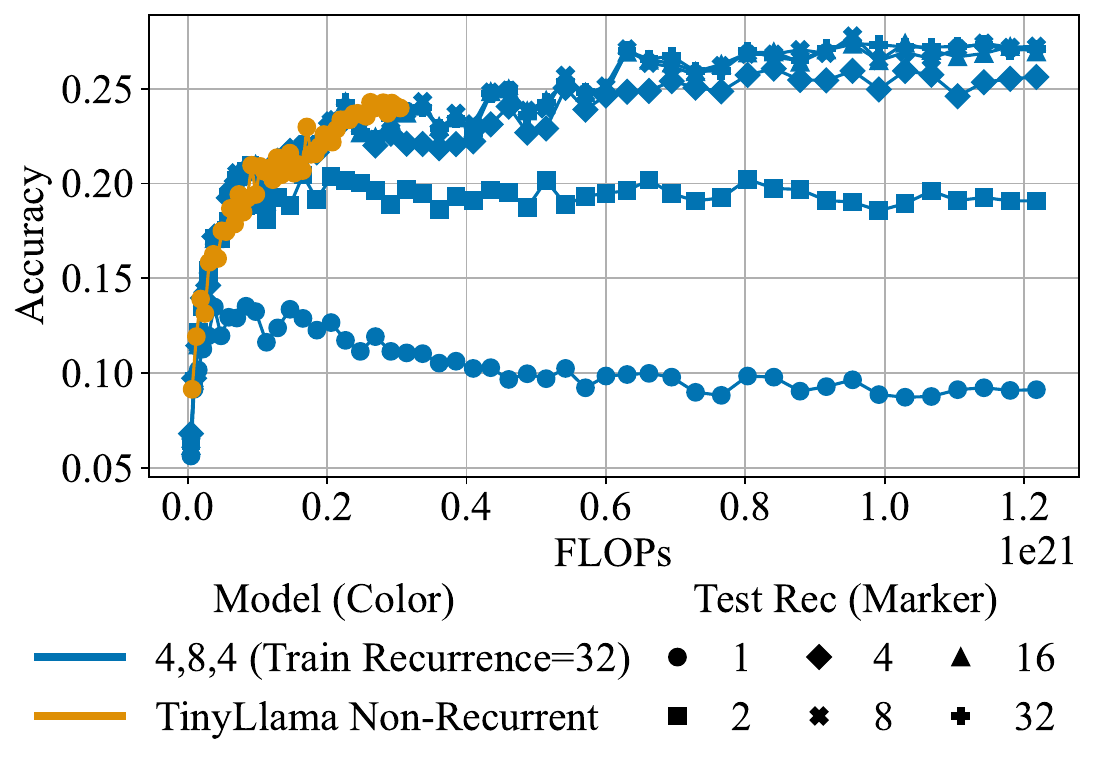}
    \caption{\textbf{Recurrence efficiently improves reasoning. }\textbf{Left}: MATH accuracy over training step for train recurrence equal to $16$ model. \textbf{Right}: MATH accuracy over training step for train recurrence equal to \(32\) model.}
    \label{app-fig:tinyllama-MATH-16-32}
\end{figure}

In \Cref{app-fig:tinyllama-MATH-all}, we show evaluation results over FLOPs for MATH.
In \Cref{app-fig:tinyllama-MATH-effective-params}, we plot Right of \Cref{app-fig:tinyllama-MATH-all} with an effective parameters x-axis, which can be viewed as proportional to FLOPs required for inference.
In Figures \ref{app-fig:tinyllama-MATH-4-8} and \ref{app-fig:tinyllama-MATH-16-32} we show the MATH accuracy over training step for train recurrences \(4,8,16\) and \(32\).

In \Cref{app-tab:tinyllama-all-evals}, we show a broad range of evaluations for the models in Figures \ref{fig:tinyllama-GSM8K} and \ref{app-fig:tinyllama-GSM8K-all}.

\begin{table}
    \centering
    \caption{\textbf{Final step accuracy for models shown in \Cref{fig:tinyllama-GSM8K} on a broad range of evaluations.} We also include \textit{\tinyllama-1.1b-3T Hugging Face} which is our evaluations of the \tinyllama-1.1b-3T model downloaded from Hugging Face, i.e. the step 0 accuracy of the non-recurrent \tinyllama model.}
    \begin{tabular}{cccccccccc}
        \toprule
         Test Rec  & Arc-E & Arc-C & HS & WG & MMLU & PIQA & OBQA & GSM8K & MATH\\
        \midrule
        \multicolumn{10}{c}{4,8,4 (Train Recurrence=4)} \\
        \midrule
        1 & 54.5 & 33.2 & 39.7 & 55.2 & 28.5 & 64.0 & 32.4 & 17.6 & 20.6 \\
        2 & 57.8 & 34.8 & 42.3 & 54.8 & 30.1 & 65.3 & 33.6 & 32.4 & 24.9 \\
        4 & \underline{58.5} & 34.6 & 42.9 & 55.3 & 32.1 & 65.7 & 33.4 & 38.0 & 25.9 \\
        8 & \textbf{58.6} & 34.0 & 43.2 & 54.6 & 32.0 & 65.8 & 33.6 & 37.7 & 25.9 \\
        16 & \underline{58.5} & 33.8 & 43.2 & 54.8 & 32.0 & 65.9 & 33.6 & 37.7 & 26.1 \\
        32 & \textbf{58.6} & 33.8 & 43.2 & 54.9 & 32.0 & 65.9 & 33.6 & 37.9 & 26.0 \\
        \midrule
        \multicolumn{10}{c}{4,8,4 (Train Recurrence=8)} \\
        \midrule
        1 & 51.6 & 32.8 & 38.9 & 51.8 & 26.8 & 64.3 & 32.0 & 15.2 & 16.8 \\
        2 & 55.5 & 34.3 & 42.1 & 55.0 & 31.2 & 65.7 & 32.0 & 34.3 & 24.9 \\
        4 & 57.1 & 34.2 & 43.2 & 56.4 & 34.0 & 66.1 & 33.2 & 40.5 & 26.5 \\
        8 & 57.5 & 34.1 & 43.4 & 56.9 & 34.1 & 66.2 & 32.6 & 40.9 & 26.1 \\
        16 & 57.9 & 34.3 & 43.5 & 56.7 & 34.1 & 66.4 & 32.8 & 40.9 & 25.9 \\
        32 & 58.0 & 34.4 & 43.5 & 57.0 & 34.1 & 66.2 & 32.8 & 41.0 & 25.9 \\
        \midrule
        \multicolumn{10}{c}{4,8,4 (Train Recurrence=16)} \\
        \midrule
        1 & 50.6 & 31.3 & 39.4 & 52.4 & 25.0 & 61.5 & 32.4 & 11.4 & 14.2 \\
        2 & 54.7 & 34.3 & 42.7 & 54.6 & 30.2 & 65.0 & 32.2 & 32.4 & 23.3 \\
        4 & 56.5 & \textbf{35.8} & 44.6 & 55.2 & 34.0 & 65.5 & \underline{34.4} & 42.2 & \underline{28.9} \\
        8 & 56.8 & \underline{35.6} & 44.8 & 55.7 & \underline{35.0} & 66.0 & \underline{34.4} & 43.9 & 28.8 \\
        16 & 57.1 & \underline{35.6} & 44.9 & 56.3 & 34.9 & 65.9 & 34.2 & 43.4 & \textbf{29.0} \\
        32 & 57.2 & 35.4 & 44.9 & 56.2 & 34.8 & 65.9 & 34.2 & 43.7 & \textbf{29.0} \\
        \midrule
        \multicolumn{10}{c}{4,8,4 (Train Recurrence=32)} \\
        \midrule
        1 & 46.8 & 28.7 & 37.5 & 50.9 & 25.4 & 62.6 & 31.6 & 5.6 & 9.1 \\
        2 & 53.3 & 32.1 & 41.8 & 54.4 & 28.6 & 64.7 & 31.8 & 24.2 & 19.1 \\
        4 & 57.6 & 33.8 & 44.5 & 57.5 & 32.8 & 65.6 & 33.4 & 39.7 & 25.6 \\
        8 & 58.4 & 35.0 & 44.9 & \textbf{59.0} & \textbf{35.2} & 66.5 & 32.4 & \textbf{45.3} & 27.2 \\
        16 & \textbf{58.6} & \underline{35.6} & 45.0 & 58.2 & 34.8 & 66.3 & 32.0 & 44.7 & 26.9 \\
        32 & \textbf{58.6} & \underline{35.6} & 45.1 & 57.6 & 34.6 & 66.4 & 32.2 & \underline{45.0} & 27.1 \\
        \midrule
        \multicolumn{10}{c}{TinyLlama Non-Recurrent} \\
        \midrule
         & 57.5 & 34.9 & \underline{45.3} & 55.8 & 33.4 & \underline{68.8} & 32.8 & 26.6 & 24.0 \\
        \midrule
        \multicolumn{10}{c}{TinyLlama-1.1b-3T Hugging Face} \\
        \midrule
         & 55.7 & 31.0 & \textbf{59.1} & \underline{58.9} & 25.4 & \textbf{73.0} & \textbf{35.0} & 1.6 & 2.3 \\
         \bottomrule
    \end{tabular}
    \label{app-tab:tinyllama-all-evals}
\end{table}

\FloatBarrier
\subsubsection{\olmo}~\label{app-subsubsec:retrofit-olmo}

\begin{figure}[ht!]
    \centering
    \includegraphics[width=\linewidth]{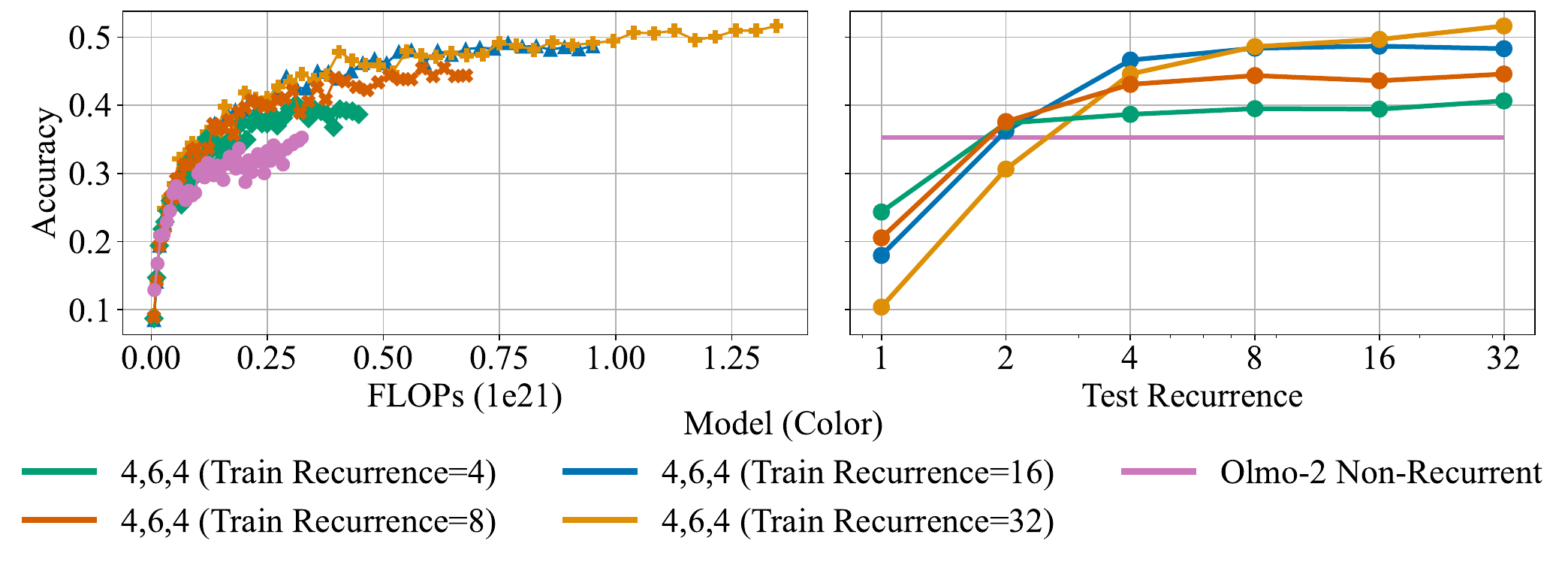}
    \caption{\textbf{Recurrence efficiently improves reasoning on GSM8K for \olmo.}
    We train \((4,8,4)\) and non-recurrent models for approximately \(50\) billion tokens of Nemotron-CC-Math-v1 data.
    \textbf{Left: }We plot accuracy over the number of FLOPs used during training. We see that recurrent models can efficiently outperform the non-recurrent baseline.
    \textbf{Right: } We plot accuracy over the number of recurrences for inference. We see the recurrent models are competitive with the fixed depth baseline and can outperform it by using more FLOPs.\\ We plot each individual models accuracy over training and recurrence in full in \Cref{app-fig:olmo-GSM8K-4-8} and \Cref{app-fig:olmo-GSM8K-16-32}. Evaluations on the final checkpoint over tasks shown in \Cref{tab:data-mix} are in Appendix \Cref{app-tab:olmo-all-evals}.}
    \label{app-fig:olmo-GSM8K-all}
\end{figure}

\begin{figure}[ht!]
    \centering
    \includegraphics[width=0.7\linewidth]{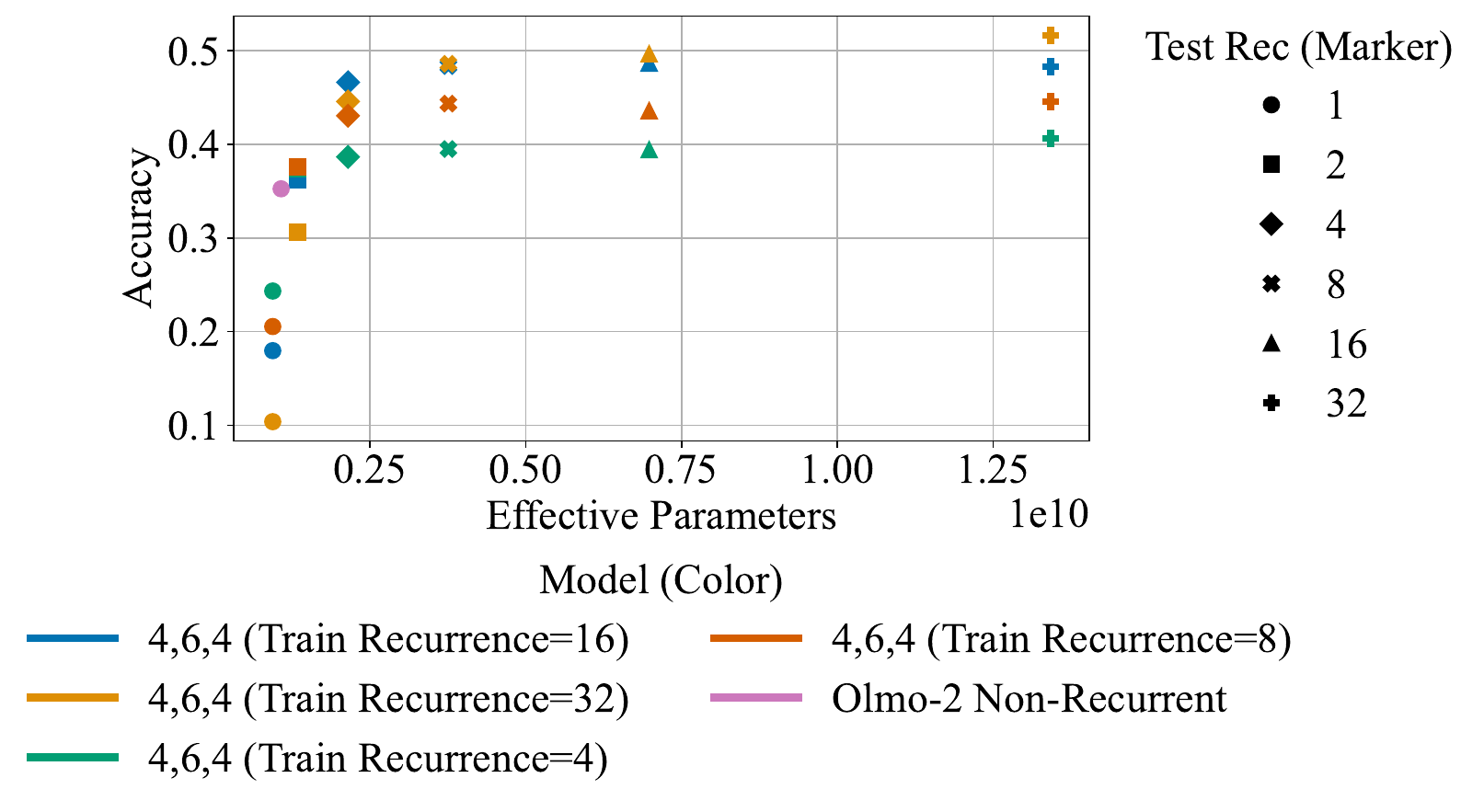}
    \caption{\textbf{Recurrent models are competitive in terms of inference FLOPs for GSM8K.} This is the same data as in \ref{app-fig:olmo-GSM8K-all} but replotted with an effective parameters x-axis, which can be viewed as proportional to FLOPs required for inference.}
    \label{app-fig:olmo-GSM8K-effective-params}
\end{figure}

\begin{figure}[ht!]
    \centering
    \includegraphics[width=0.48\linewidth]{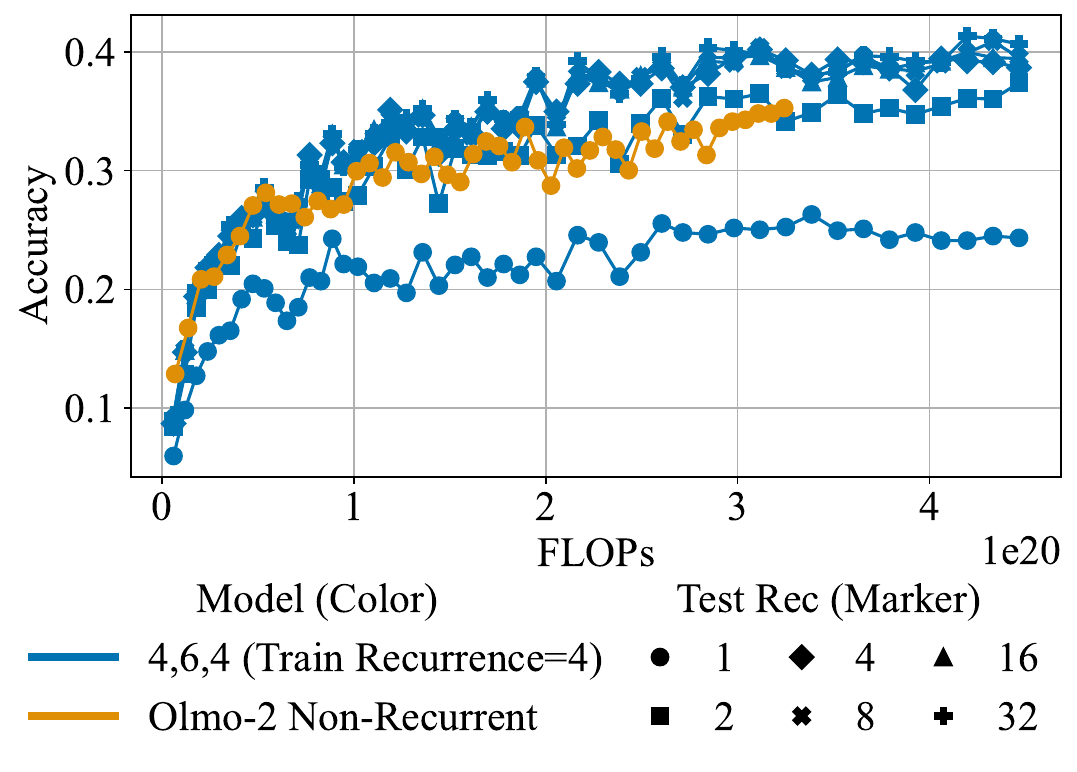}
    \includegraphics[width=0.48\linewidth]{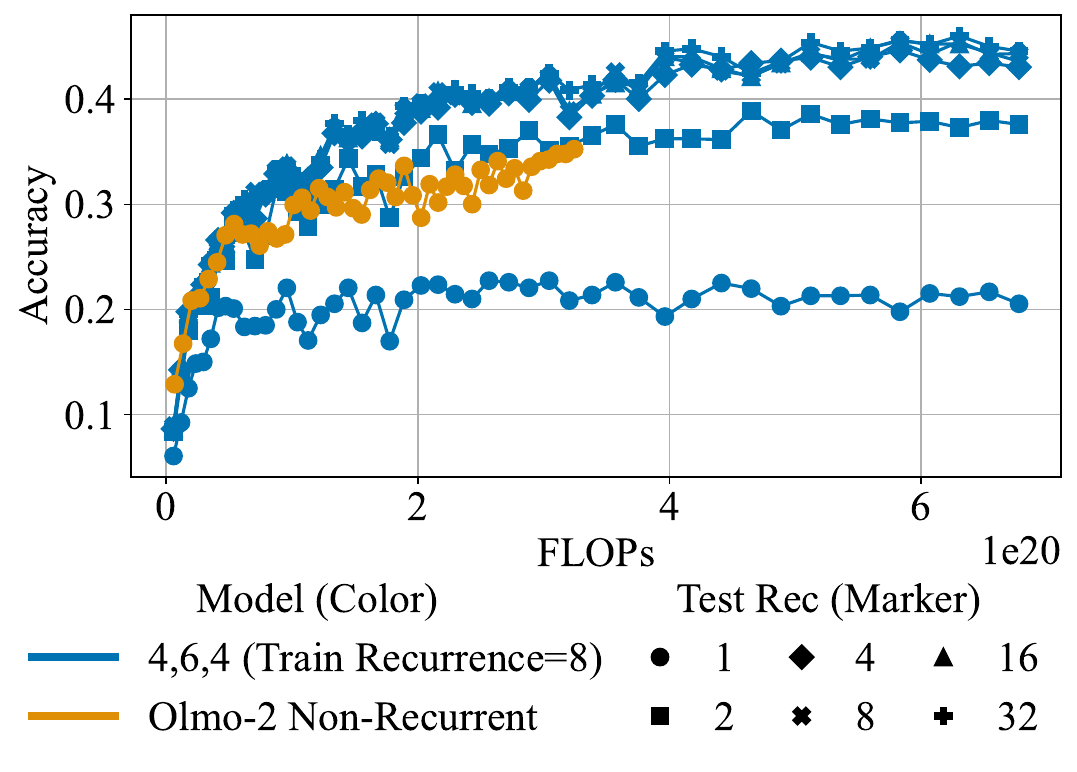}
    \caption{\textbf{Recurrence efficiently improves reasoning.} \textbf{Left}: GSM8K accuracy over training step for train recurrence equal to $4$ model. \textbf{Right}: GSM8K accuracy over training step for train recurrence equal to $8$ model.}
    \label{app-fig:olmo-GSM8K-4-8}
\end{figure}

\begin{figure}[ht!]
    \centering
    \includegraphics[width=0.48\linewidth]{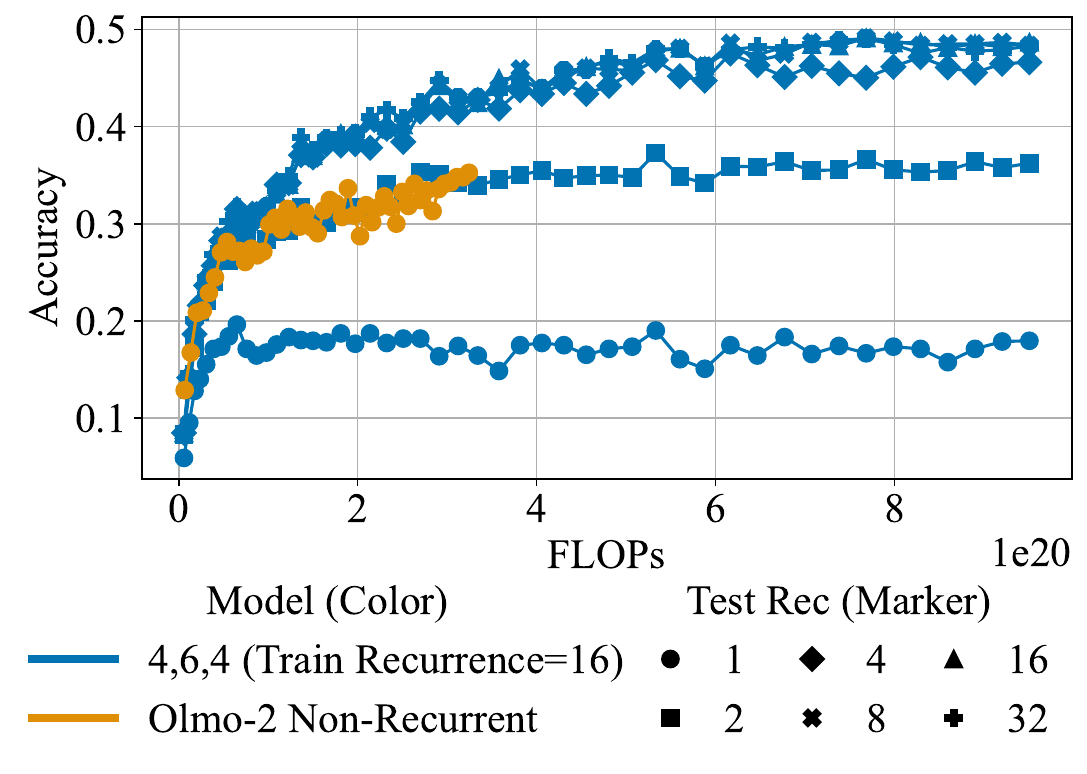}
    \includegraphics[width=0.48\linewidth]{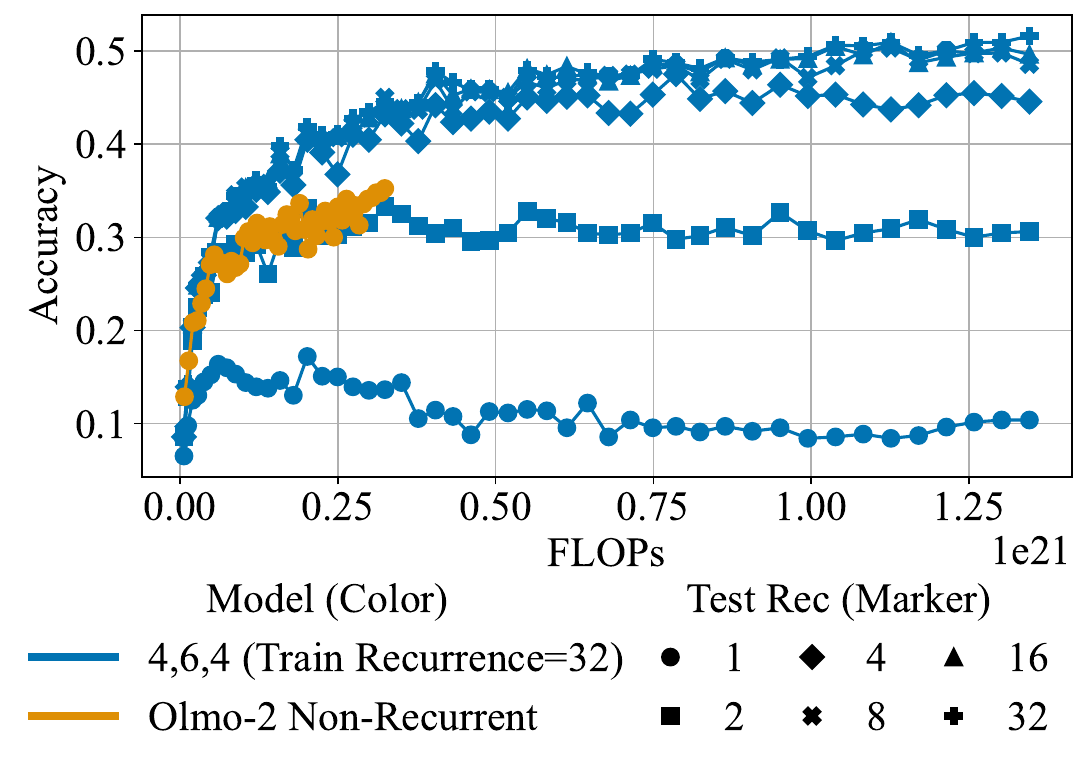}
    \caption{\textbf{Recurrence efficiently improves reasoning. }\textbf{Left}: GSM8K accuracy over training step for train recurrence equal to $16$ model. \textbf{Right}: GSM8K accuracy over training step for train recurrence equal to \(32\) model.}
    \label{app-fig:olmo-GSM8K-16-32}
\end{figure}

In \Cref{app-fig:olmo-GSM8K-all} we show evaluation results for \olmo on GSM8k.
In \Cref{app-fig:olmo-GSM8K-effective-params}, we plot Right of \Cref{app-fig:olmo-GSM8K-all} with an effective parameters x-axis, which can be viewed as proportional to FLOPs required for inference. 
In Figures \ref{app-fig:olmo-GSM8K-4-8} and \ref{app-fig:olmo-GSM8K-16-32} we show the GSM8K accuracy over training step for train recurrences \(4,8,16\) and \(32\).

\begin{figure}[ht!]
    \centering
    \includegraphics[width=\linewidth]{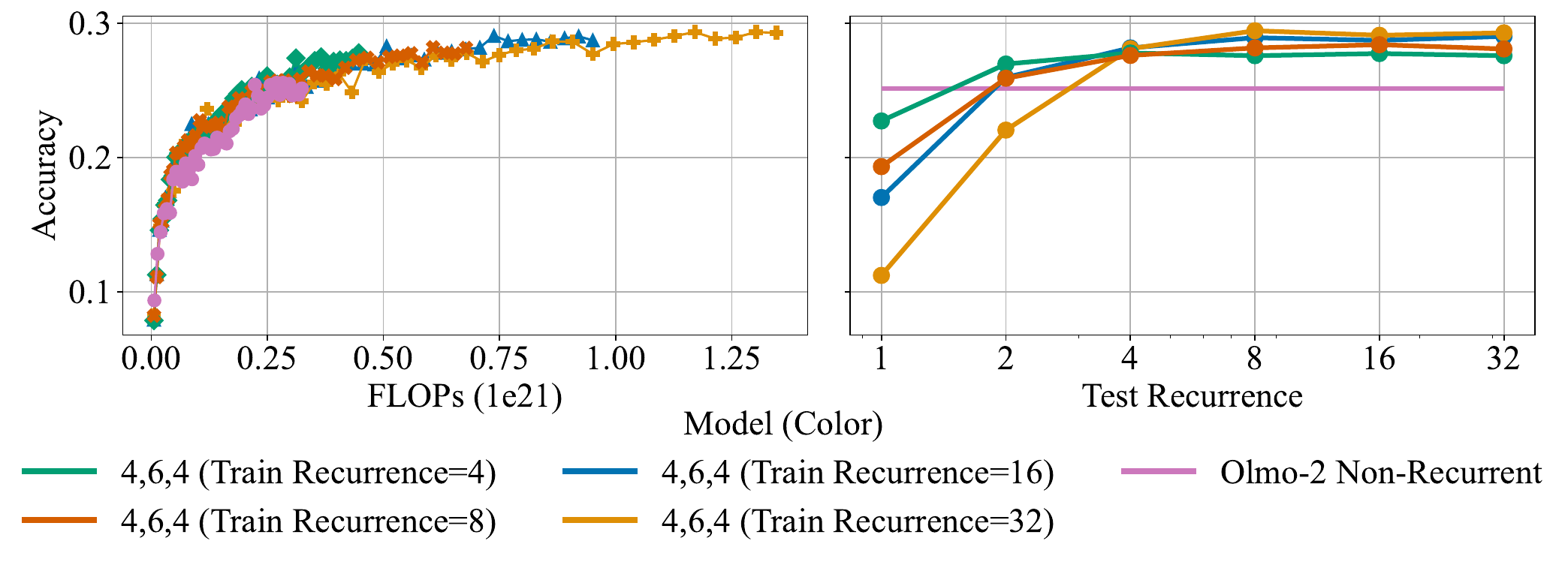}
    \caption{\textbf{Recurrence efficiently improves reasoning on MATH for \olmo.}
    We train \((4,8,4)\) and non-recurrent models for approximately \(50\) billion tokens of Nemotron-CC-Math-v1 data, extending \Cref{fig:olmo-MATH}.
    \textbf{Left: }We plot accuracy over the number of FLOPs used during training. We see that recurrent models can efficiently outperform the non-recurrent baseline.
    \textbf{Right: } We plot accuracy over the number of recurrences for inference. We see the recurrent models are competitive with the fixed depth baseline and can outperform it by using more FLOPs.\\ We plot each individual models accuracy over training and recurrence in full in \Cref{app-fig:olmo-MATH-4-8} and \Cref{app-fig:olmo-MATH-16-32}. Evaluations on the final checkpoint over tasks shown in \Cref{tab:data-mix} are in Appendix \Cref{app-tab:olmo-all-evals}.}
    \label{app-fig:olmo-MATH-all}
\end{figure}

\begin{figure}[ht!]
    \centering
    \includegraphics[width=0.7\linewidth]{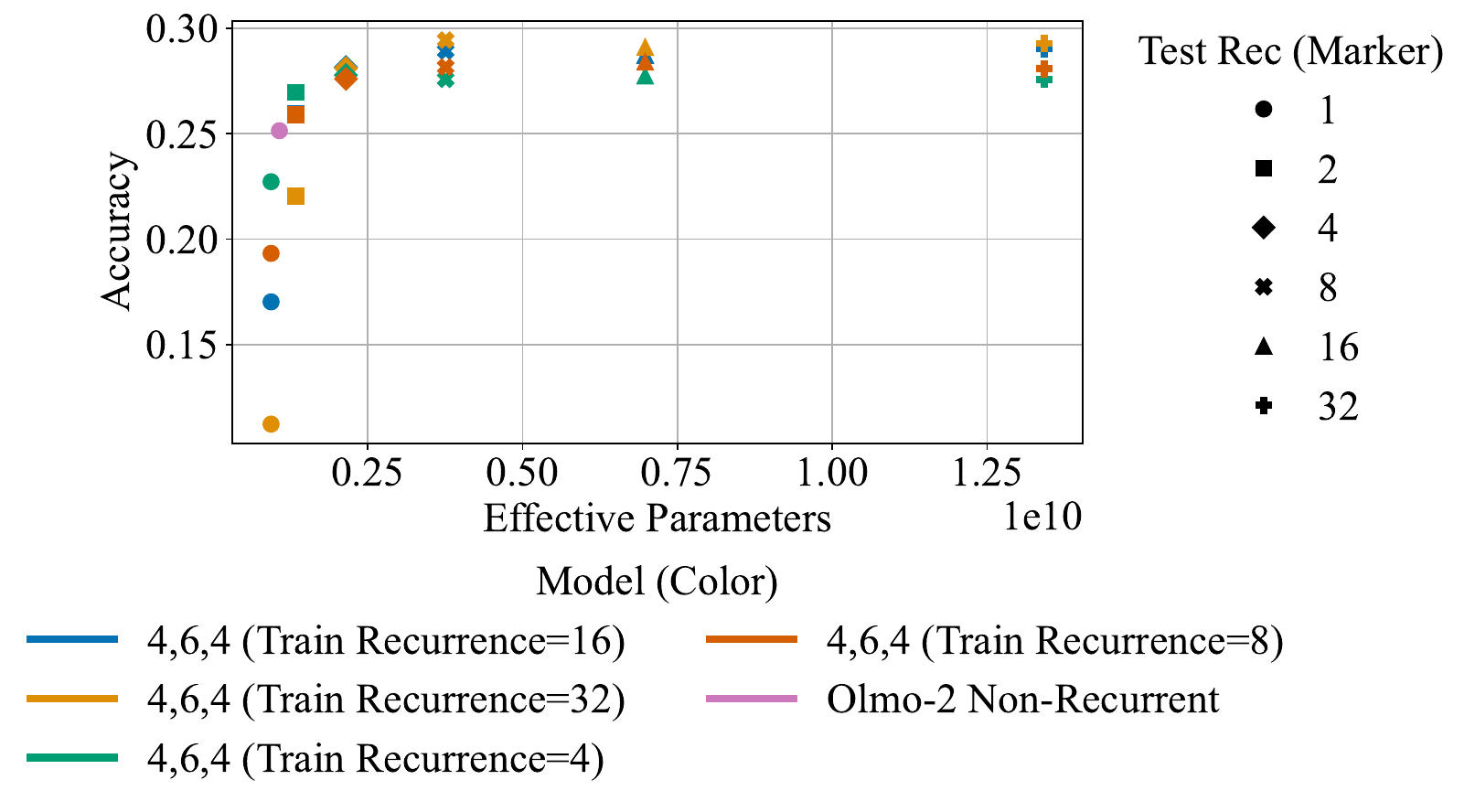}
    \caption{\textbf{Recurrent models are competitive in terms of inference FLOPs for MATH.} This is the same data as in \ref{app-fig:olmo-MATH-all} but replotted with an effective parameters x-axis, which can be viewed as proportional to FLOPs required for inference.}
    \label{app-fig:olmo-MATH-effective-params}
\end{figure}

\begin{figure}[ht!]
    \centering
    \includegraphics[width=0.48\linewidth]{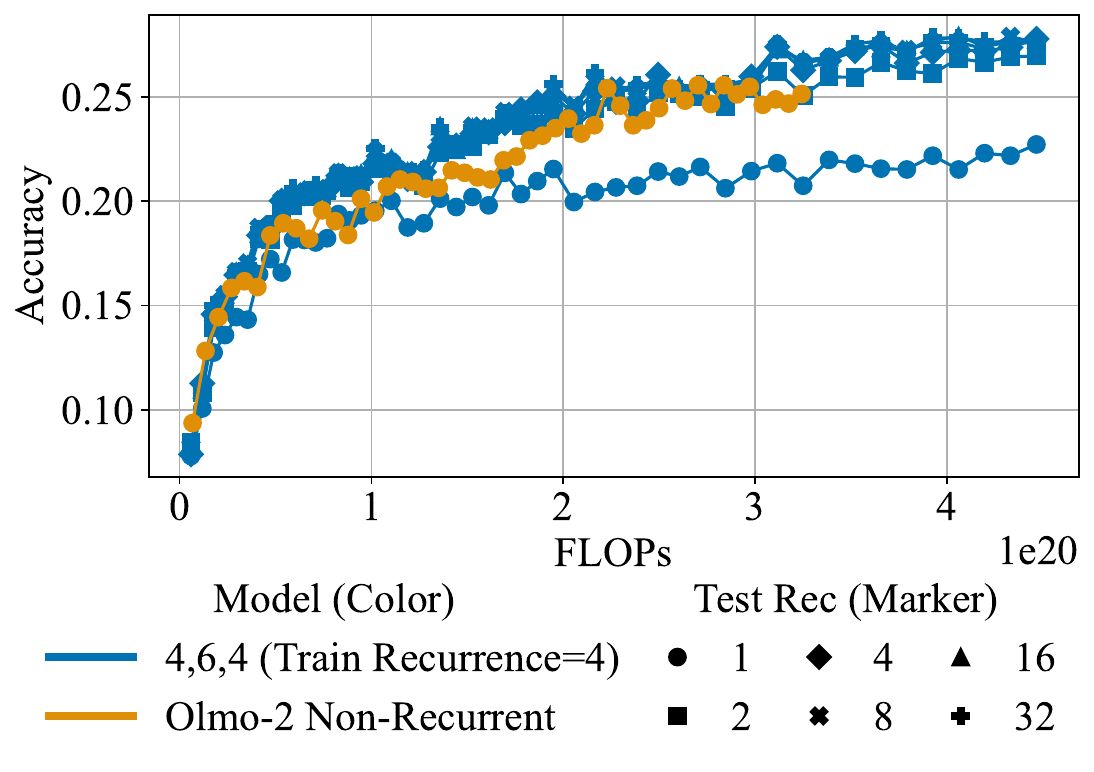}
    \includegraphics[width=0.48\linewidth]{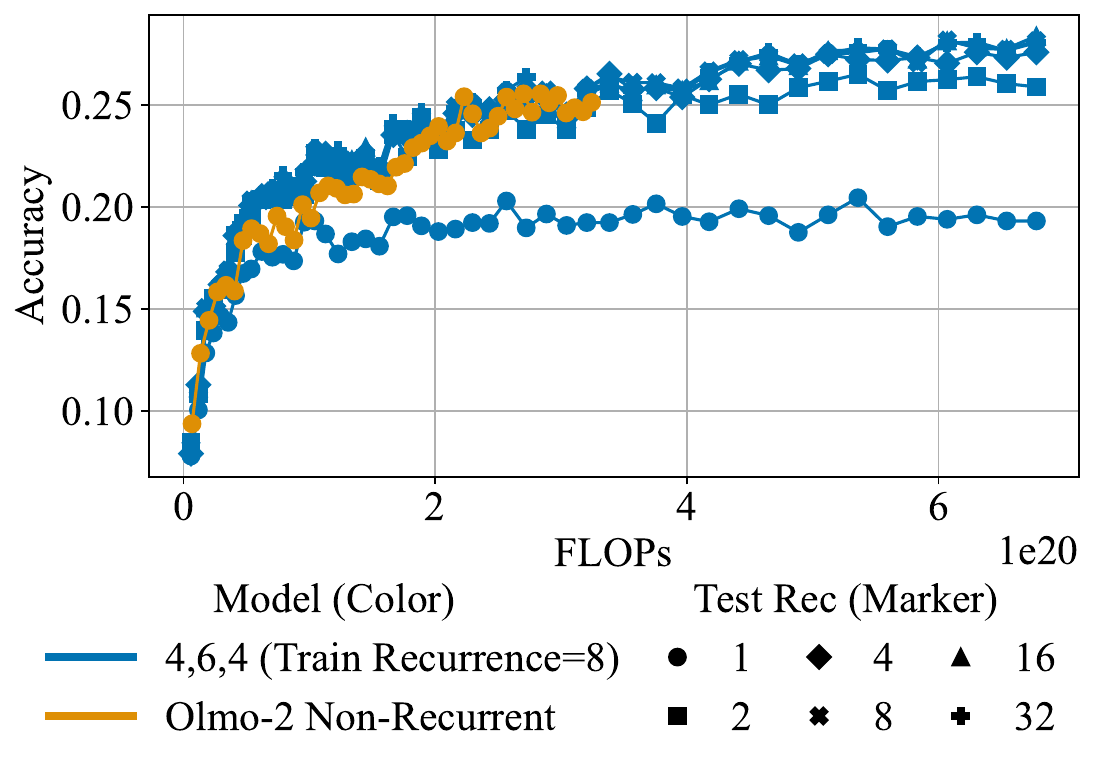}
    \caption{\textbf{Recurrence efficiently improves reasoning.} \textbf{Left}: MATH accuracy over training step for train recurrence equal to $4$ model. \textbf{Right}: MATH accuracy over training step for train recurrence equal to $8$ model.}
    \label{app-fig:olmo-MATH-4-8}
\end{figure}

\begin{figure}[ht!]
    \centering
    \includegraphics[width=0.48\linewidth]{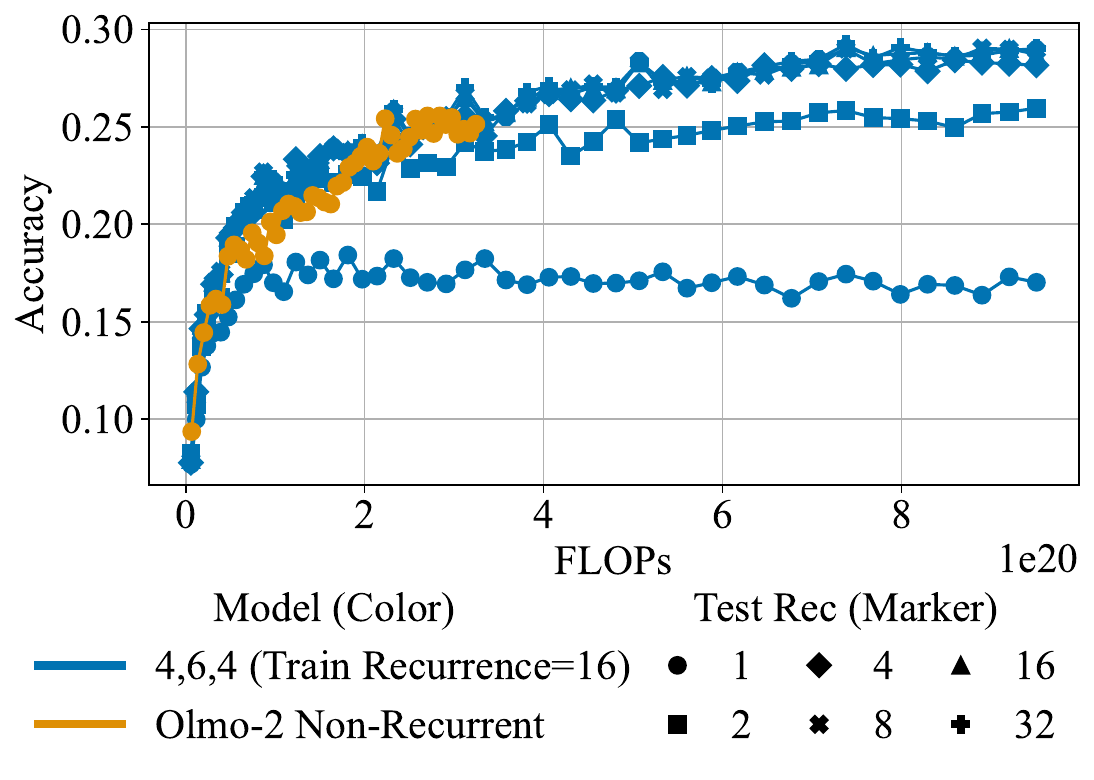}
    \includegraphics[width=0.48\linewidth]{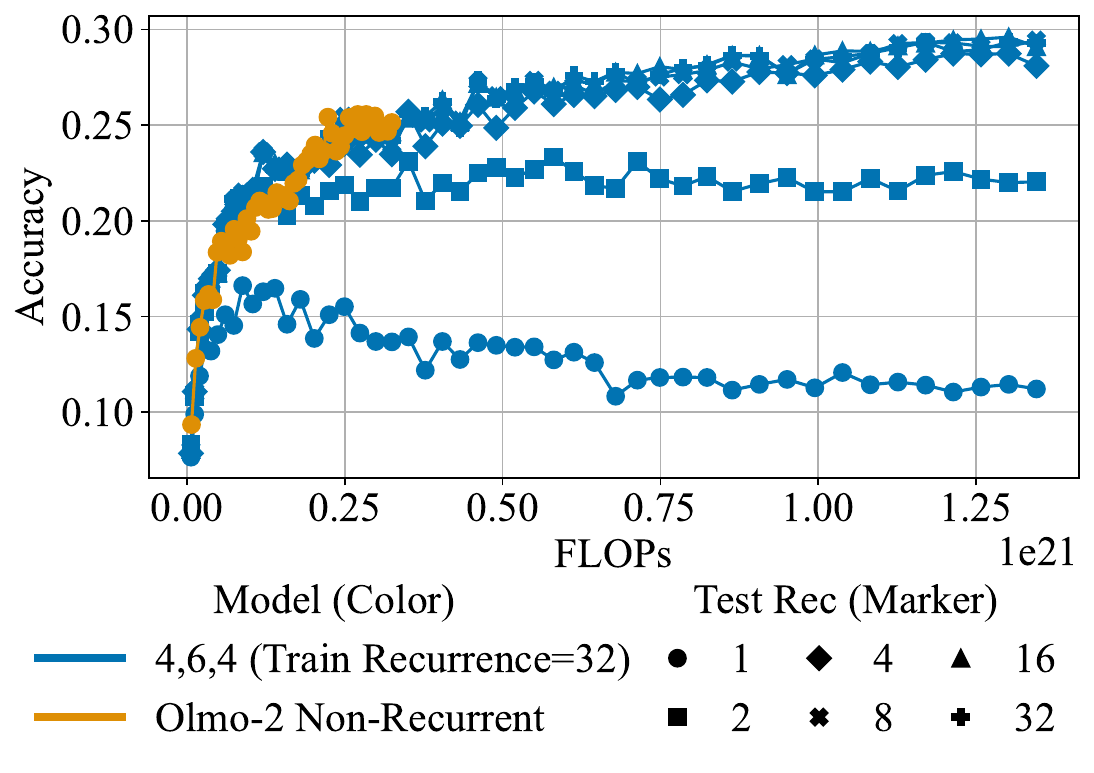}
    \caption{\textbf{Recurrence efficiently improves reasoning. }\textbf{Left}: MATH accuracy over training step for train recurrence equal to $16$ model. \textbf{Right}: MATH accuracy over training step for train recurrence equal to \(32\) model.}
    \label{app-fig:olmo-MATH-16-32}
\end{figure}

In \Cref{app-fig:olmo-MATH-all}, we extend \Cref{fig:olmo-MATH}, showing more training recurrences.
In \Cref{app-fig:olmo-MATH-effective-params}, we plot Right of \Cref{app-fig:olmo-MATH-all} with an effective parameters x-axis, which can be viewed as proportional to FLOPs required for inference.
In Figures \ref{app-fig:olmo-MATH-4-8} and \ref{app-fig:olmo-MATH-16-32} we show the MATH accuracy over training step for train recurrences \(4,8,16\) and \(32\).

In \Cref{app-tab:olmo-all-evals}, we show a broad range of evaluations for the models in \Cref{app-fig:olmo-GSM8K-all}.

\begin{table}
    \centering
    \caption{\textbf{Final step accuracy for models shown in \Cref{fig:olmo-MATH} on a broad range of evaluations.} We also include \textit{\olmo-2-0425-1B-step1907359 Hugging Face} which is our evaluations of the \olmo-2-0425-1B-step1907359 model downloaded from Hugging Face, i.e. the step 0 accuracy of the non-recurrent \olmo model.}
    \begin{tabular}{cccccccccc}
        \toprule
         Test Rec  & Arc-E & Arc-C & HS & WG & MMLU & PIQA & OBQA & GSM8K & MATH\\
        \midrule
        \multicolumn{10}{c}{4,6,4 (Train Recurrence=4)} \\
        \midrule
        1 & 61.6 & 36.3 & 46.4 & 56.8 & 36.4 & 68.4 & 33.6 & 24.3 & 22.7 \\
        2 & 63.8 & 37.7 & 48.3 & 58.2 & 37.9 & 69.5 & 35.6 & 37.4 & 27.0 \\
        4 & 63.8 & 37.4 & 48.8 & 57.7 & 38.3 & 69.6 & 36.2 & 38.7 & 27.8 \\
        8 & 63.7 & 37.4 & 49.0 & 57.2 & 38.2 & 69.9 & 35.8 & 39.5 & 27.6 \\
        16 & 63.6 & 37.3 & 49.0 & 57.1 & 38.1 & 70.0 & 35.8 & 39.4 & 27.7 \\
        32 & 63.6 & 37.3 & 49.0 & 57.2 & 38.1 & 70.0 & 35.8 & 40.6 & 27.6 \\
        \midrule
        \multicolumn{10}{c}{4,6,4 (Train Recurrence=8)} \\
        \midrule
        1 & 60.9 & 37.0 & 45.9 & 55.8 & 35.2 & 69.2 & 32.0 & 20.5 & 19.3 \\
        2 & 64.0 & 39.1 & 48.4 & 58.4 & 37.5 & 69.7 & 34.4 & 37.6 & 25.9 \\
        4 & 64.8 & 39.6 & 49.2 & 59.9 & 39.0 & 70.3 & 34.4 & 43.1 & 27.6 \\
        8 & 65.0 & 39.7 & 49.4 & 59.3 & 39.2 & \underline{70.6} & 34.4 & 44.4 & 28.2 \\
        16 & 65.0 & 39.4 & 49.5 & 59.4 & 39.1 & 70.5 & 34.2 & 43.6 & 28.4 \\
        32 & 65.0 & 39.4 & 49.5 & 59.4 & 39.1 & 70.5 & 34.2 & 44.6 & 28.1 \\
        \midrule
        \multicolumn{10}{c}{4,6,4 (Train Recurrence=16)} \\
        \midrule
        1 & 57.6 & 35.6 & 45.3 & 56.0 & 33.7 & 68.1 & \underline{36.8} & 18.0 & 17.0 \\
        2 & 62.3 & 38.7 & 48.7 & 59.3 & 37.1 & 67.8 & 34.0 & 36.2 & 26.0 \\
        4 & 64.1 & \underline{40.2} & 49.8 & 58.6 & 39.4 & 69.4 & 34.2 & 46.6 & 28.2 \\
        8 & 65.2 & 39.5 & 49.8 & 58.6 & 39.9 & 69.8 & 34.6 & 48.4 & 28.9 \\
        16 & 65.2 & 39.8 & 49.8 & 58.0 & 39.8 & 70.0 & 34.8 & 48.7 & 28.7 \\
        32 & 65.2 & 39.8 & 49.8 & 58.0 & 39.8 & 70.0 & 34.8 & 48.3 & 29.0 \\
        \midrule
        \multicolumn{10}{c}{4,6,4 (Train Recurrence=32)} \\
        \midrule
        1 & 56.9 & 33.2 & 44.4 & 54.1 & 30.9 & 66.8 & 33.4 & 10.4 & 11.2 \\
        2 & 61.7 & 37.8 & 47.9 & 55.5 & 36.5 & 68.1 & 34.2 & 30.6 & 22.0 \\
        4 & 65.2 & 38.9 & 49.4 & 59.2 & 39.3 & 68.8 & 33.2 & 44.6 & 28.1 \\
        8 & 66.0 & 39.8 & 49.6 & 58.2 & \underline{40.4} & 69.6 & 34.4 & 48.6 & \textbf{29.4} \\
        16 & \underline{66.1} & \underline{40.2} & 49.8 & 57.8 & \textbf{40.5} & 69.9 & 34.4 & \underline{49.7} & 29.1 \\
        32 & 66.0 & \underline{40.2} & 49.8 & 57.5 & \textbf{40.5} & 70.0 & 34.4 & \textbf{51.6} & \underline{29.3} \\
        \midrule
        \multicolumn{10}{c}{Olmo-2 Non-Recurrent} \\
        \midrule
         & 65.2 & \textbf{40.8} & \underline{50.7} & \underline{60.0} & 40.0 & 70.0 & 35.4 & 35.3 & 25.1 \\
        \midrule
        \multicolumn{10}{c}{OLMo-2-0425-1B-step1907359 Hugging Face} \\
        \midrule
         & \textbf{67.6} & 39.2 & \textbf{67.0} & \textbf{65.3} & 24.6 & \textbf{76.3} & \textbf{39.2} & 3.6 & 3.4 \\
         \bottomrule
    \end{tabular}
    \label{app-tab:olmo-all-evals}
\end{table}

\FloatBarrier
\subsubsection{\llama}~\label{app-subsubsec:retrofit-llama}
We build another set of models initialized from the weights of \llama-3.2-1B. 
For \llama, we construct \((4,6,4)\) configurations, removing \(2\) layers (layers \(4\) and \(5\) with \(0\) indexing) from the pretrained model.
This leaves approximately \(850\) million remaining parameters in the recurrent model, which equates to \(87.5\%\) of the pretrained models parameters.

\begin{figure}[ht!]
    \centering
    \includegraphics[width=\linewidth]{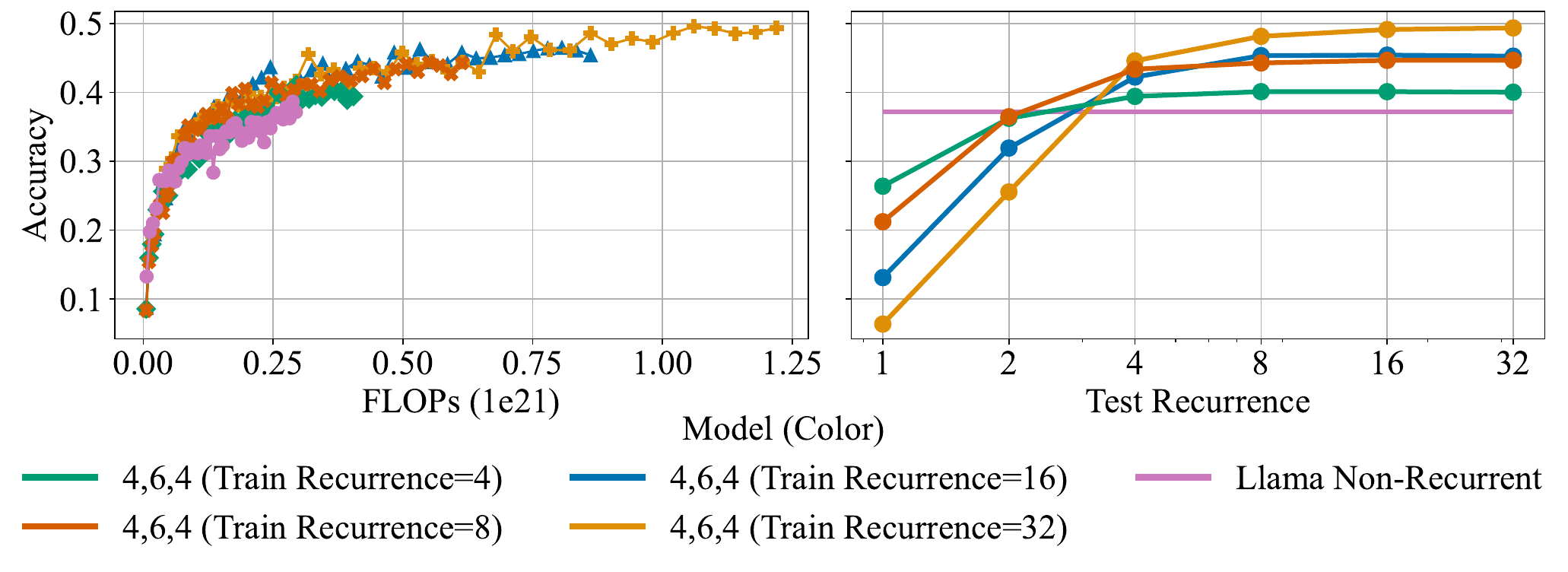}
    \caption{\textbf{Recurrence efficiently improves reasoning on GSM8K for \llama.}
    We train \((4,6,4)\) and non-recurrent models for approximately \(50\) billion tokens of Nemotron-CC-Math-v1 data.
    \textbf{Left: }We plot accuracy over the number of FLOPs used during training. We see that recurrent models can efficiently outperform the non-recurrent baseline.
    \textbf{Right: } We plot accuracy over the number of recurrences for inference. We see the recurrent models are competitive with the fixed depth baseline and can outperform it by using more FLOPs.\\ We plot each individual models accuracy over training and recurrence in full in \Cref{app-fig:llama-GSM8K-4-8} and \Cref{app-fig:llama-GSM8K-16-32}. Evaluations on the final checkpoint over tasks shown in \Cref{tab:data-mix} are in Appendix \Cref{app-tab:llama-all-evals}.}
    \label{app-fig:llama-GSM8K-all}
\end{figure}

\begin{figure}[ht!]
    \centering
    \includegraphics[width=0.7\linewidth]{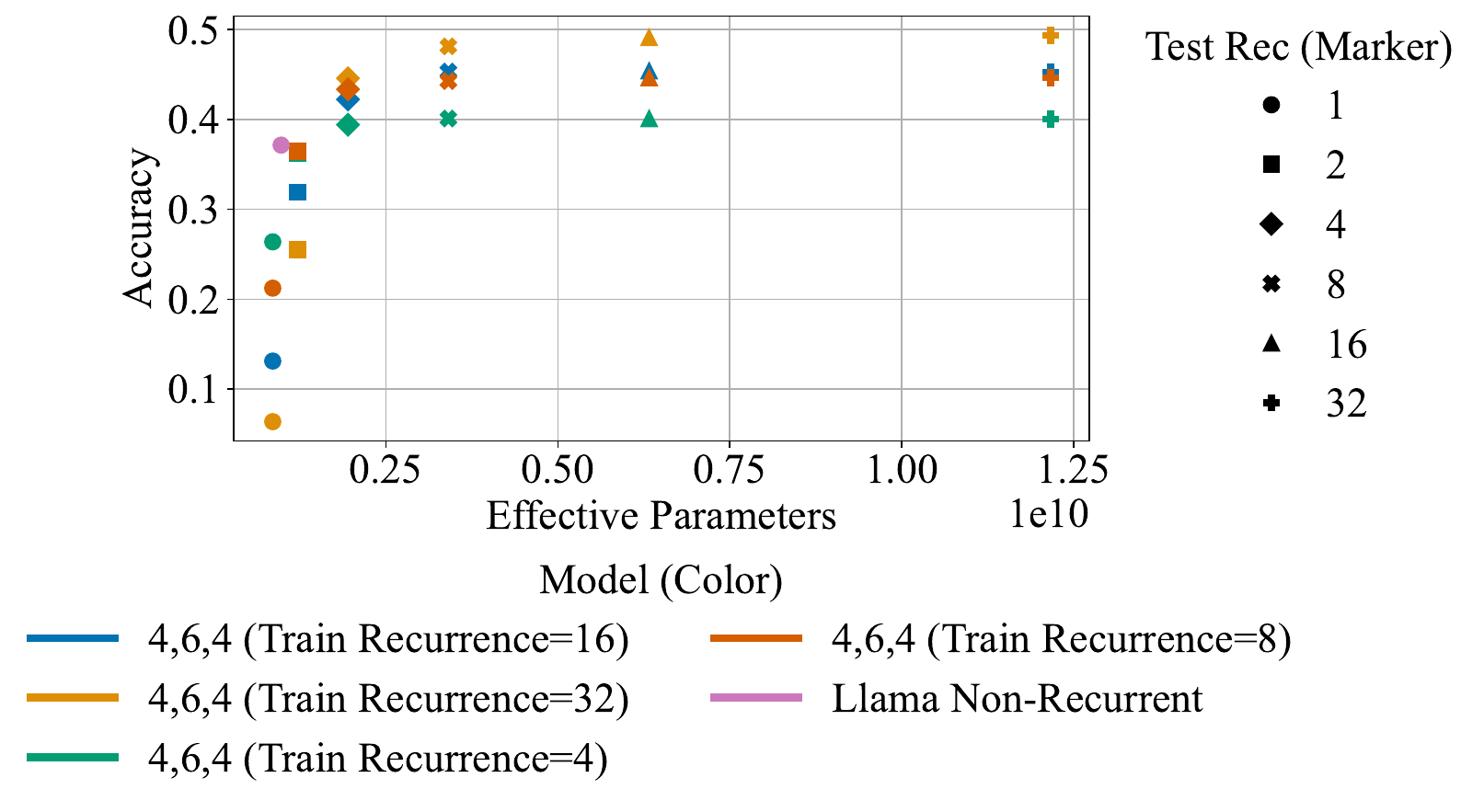}
    \caption{\textbf{Recurrent models are competitive in terms of inference FLOPs for GSM8K.} This is the same data as in Right of \Cref{app-fig:llama-GSM8K-all} but replotted with an effective parameters x-axis, which can be viewed as proportional to FLOPs required for inference. }
    \label{app-fig:llama-GSM8K-effective-params}
\end{figure}

\begin{figure}[ht!]
    \centering
    \includegraphics[width=0.48\linewidth]{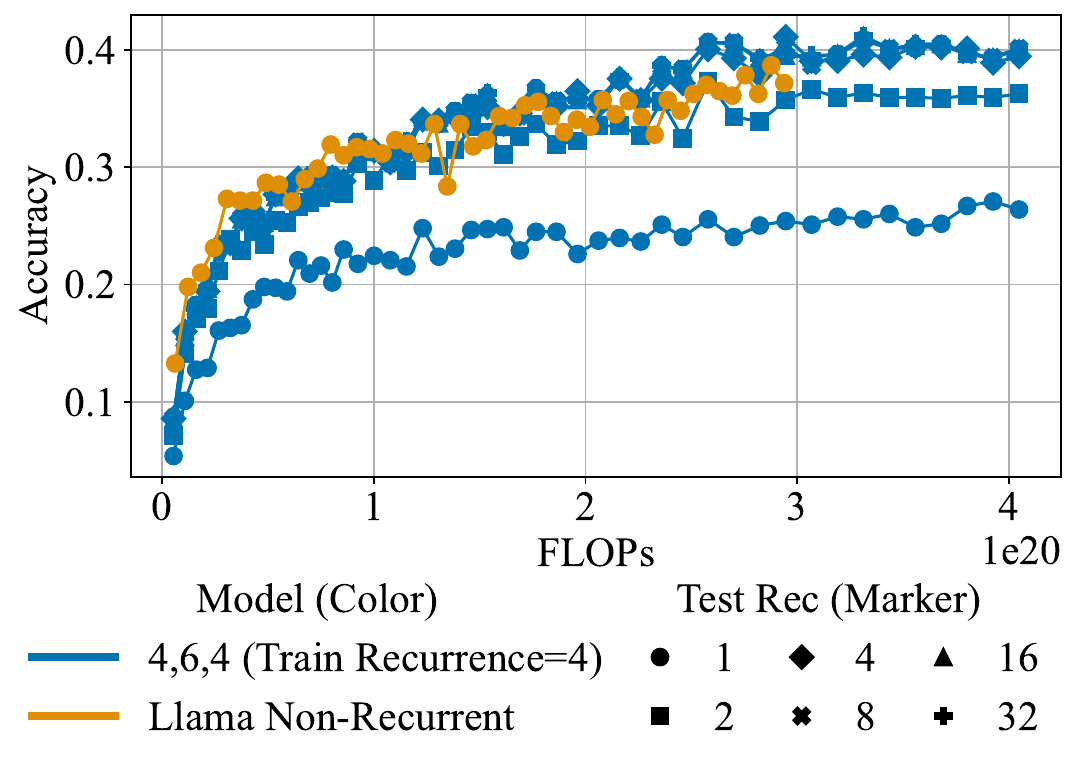}
    \includegraphics[width=0.48\linewidth]{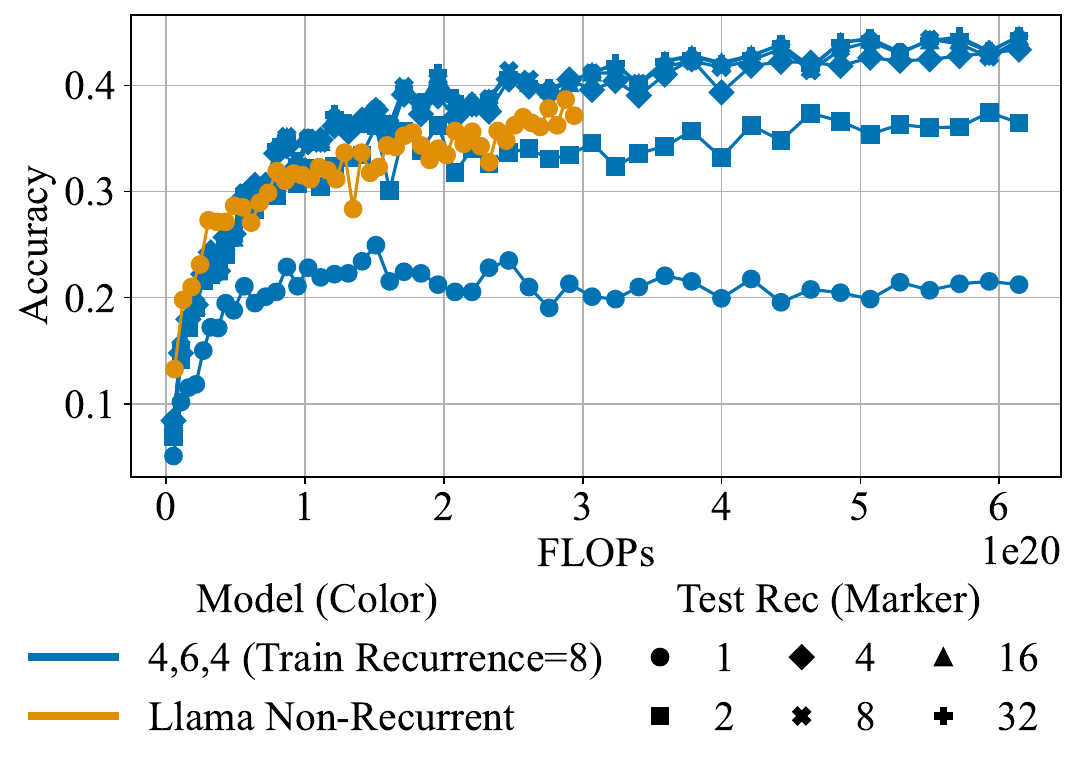}
    \caption{\textbf{Recurrence efficiently improves reasoning.} \textbf{Left}: GSM8K accuracy over training step for train recurrence equal to $4$ model. \textbf{Right}: GSM8K accuracy over training step for train recurrence equal to \(8\) model.}
    \label{app-fig:llama-GSM8K-4-8}
\end{figure}

\begin{figure}[ht!]
    \centering
    \includegraphics[width=0.48\linewidth]{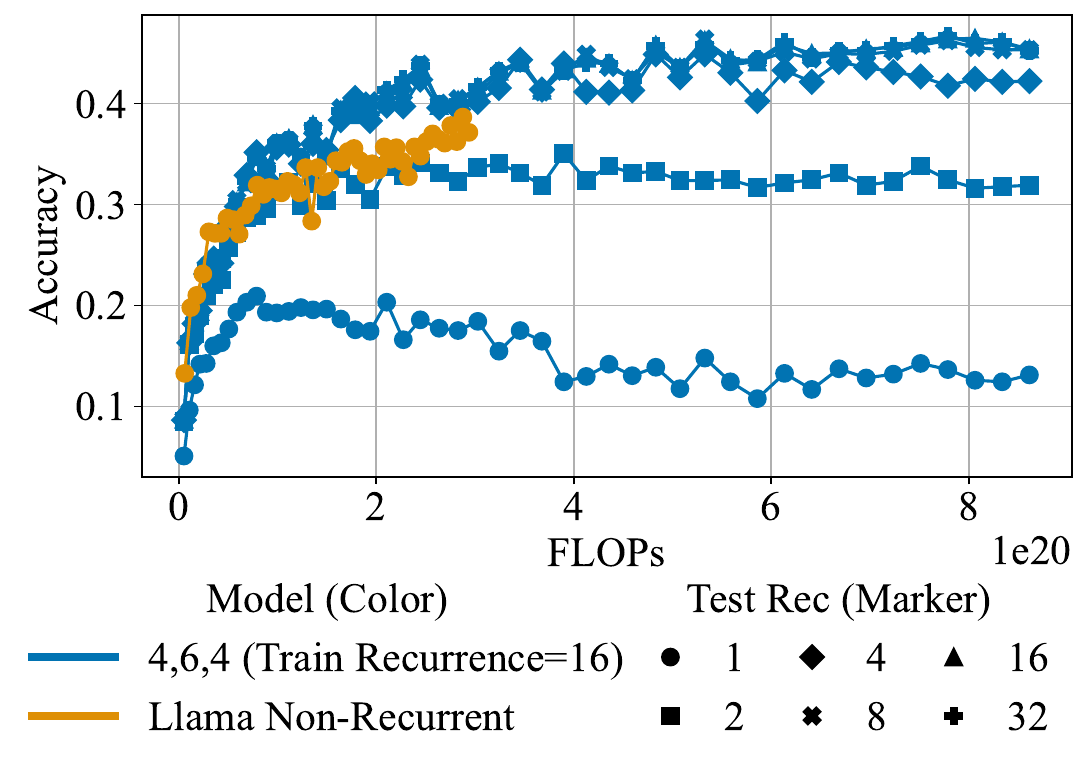}
    \includegraphics[width=0.48\linewidth]{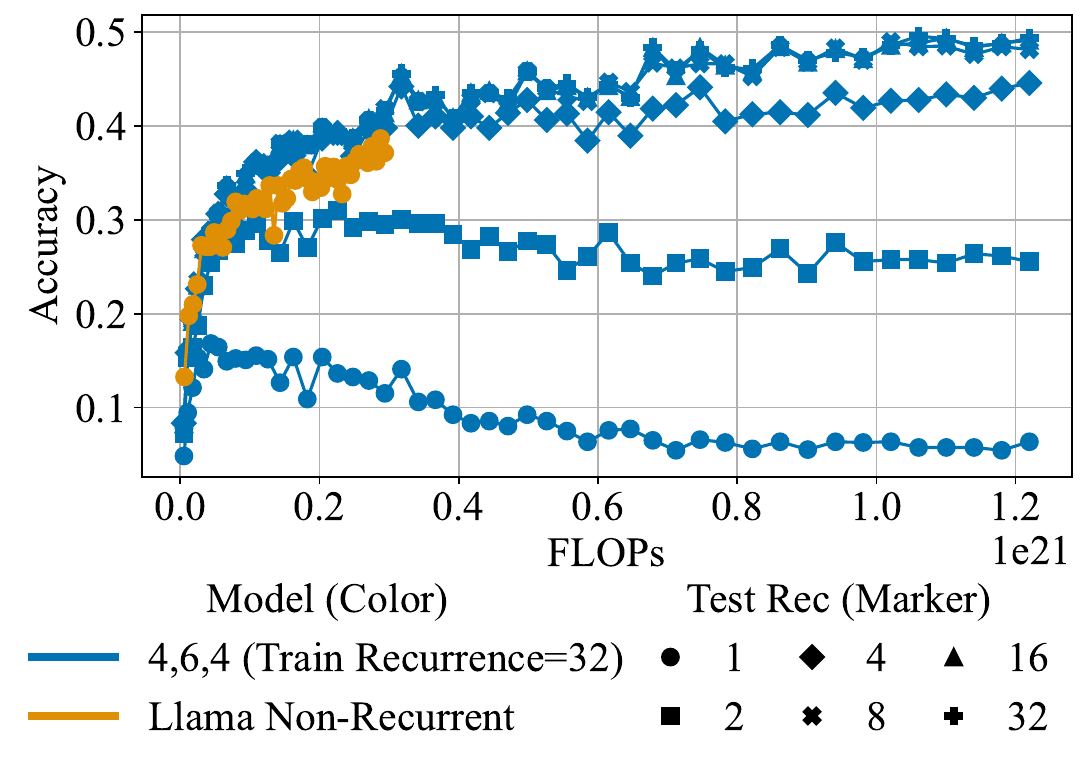}
    \caption{\textbf{Recurrence efficiently improves reasoning.} \textbf{Left}: GSM8K accuracy over training step for train recurrence equal to $16$ model. \textbf{Right}: GSM8K accuracy over training step for train recurrence equal to \(32\) model.}
    \label{app-fig:llama-GSM8K-16-32}
\end{figure}

In \Cref{app-fig:llama-GSM8K-all} we show evaluation results for \llama on GSM8k.
In \Cref{app-fig:llama-GSM8K-effective-params}, we plot Right of \Cref{app-fig:llama-GSM8K-all} with an effective parameters x-axis, which can be viewed as proportional to FLOPs required for inference. 
In Figures \ref{app-fig:llama-GSM8K-4-8} and \ref{app-fig:llama-GSM8K-16-32} we show the GSM8K accuracy over training step for train recurrences \(4,8,16\) and \(32\).

\begin{figure}[ht!]
    \centering
    \includegraphics[width=\linewidth]{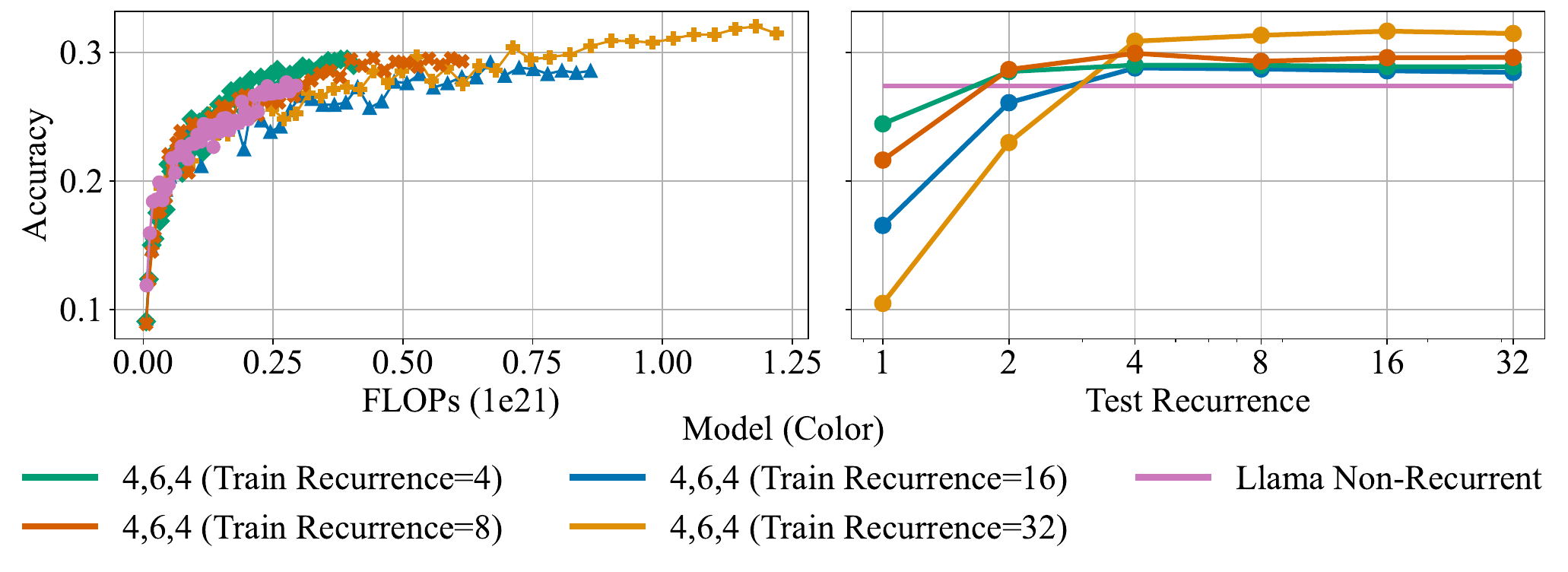}
    \caption{\textbf{Recurrence efficiently improves reasoning on MATH for \llama.}
    We train \((4,6,4)\) and non-recurrent models for approximately \(50\) billion tokens of Nemotron-CC-Math-v1 data.
    \textbf{Left: }We plot accuracy over the number of FLOPs used during training. We see that recurrent models can efficiently outperform the non-recurrent baseline.
    \textbf{Right: } We plot accuracy over the number of recurrences for inference. We see the recurrent models are competitive with the fixed depth baseline and can outperform it by using more FLOPs.\\ We plot each individual models accuracy over training and recurrence in full in \Cref{app-fig:llama-MATH-4-8} and \Cref{app-fig:llama-MATH-16-32}. Evaluations on the final checkpoint over tasks shown in \Cref{tab:data-mix} are in Appendix \Cref{app-tab:llama-all-evals}.}
    \label{app-fig:llama-MATH-all}
\end{figure}

\begin{figure}[ht!]
    \centering
    \includegraphics[width=0.7\linewidth]{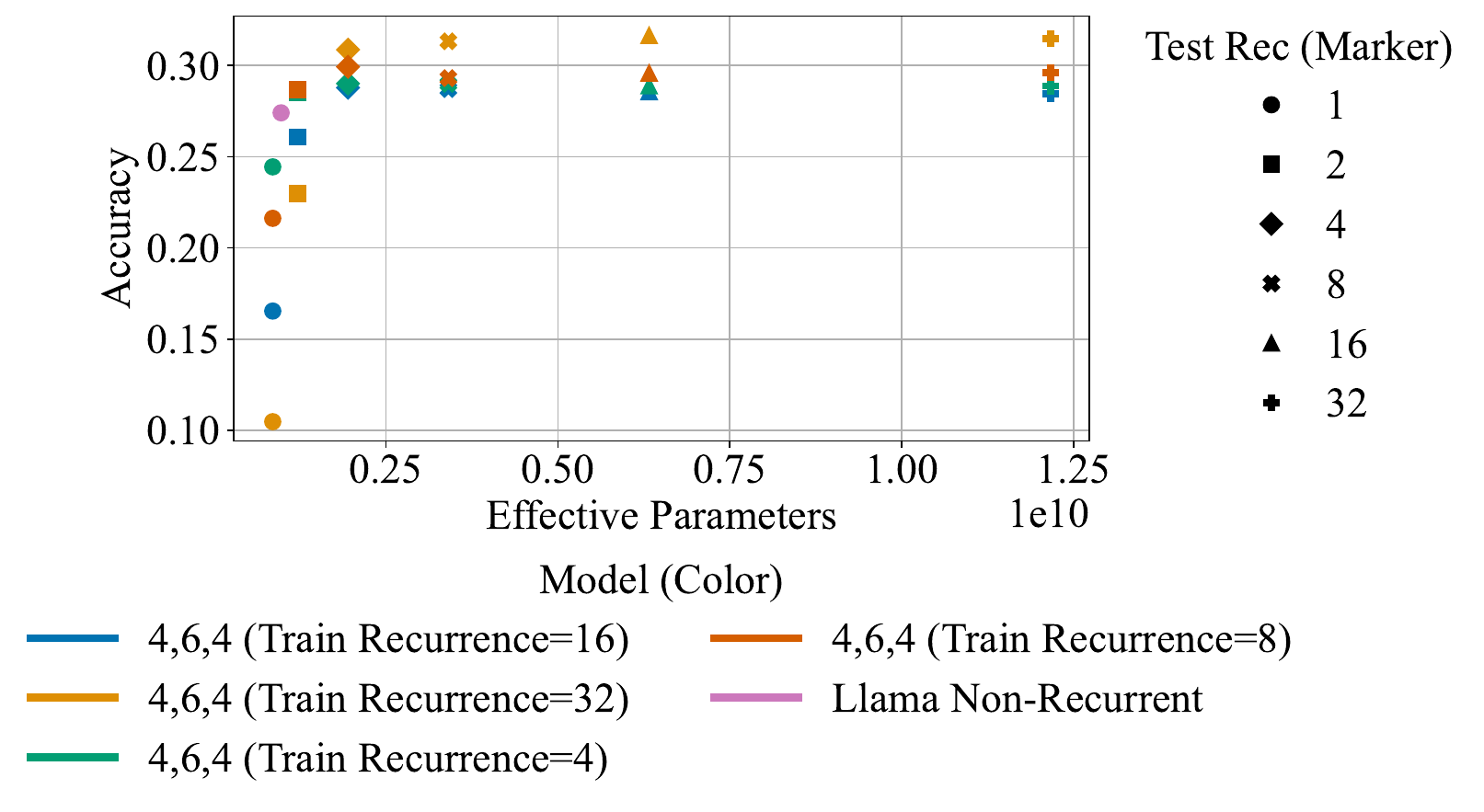}
    \caption{\textbf{Recurrent models are competitive in terms of inference FLOPs for MATH.} This is the same data as in \ref{app-fig:llama-MATH-all} but replotted with an effective parameters x-axis, which can be viewed as proportional to FLOPs required for inference. }
    \label{app-fig:llama-MATH-effective-params}
\end{figure}

\begin{figure}[ht!]
    \centering
    \includegraphics[width=0.48\linewidth]{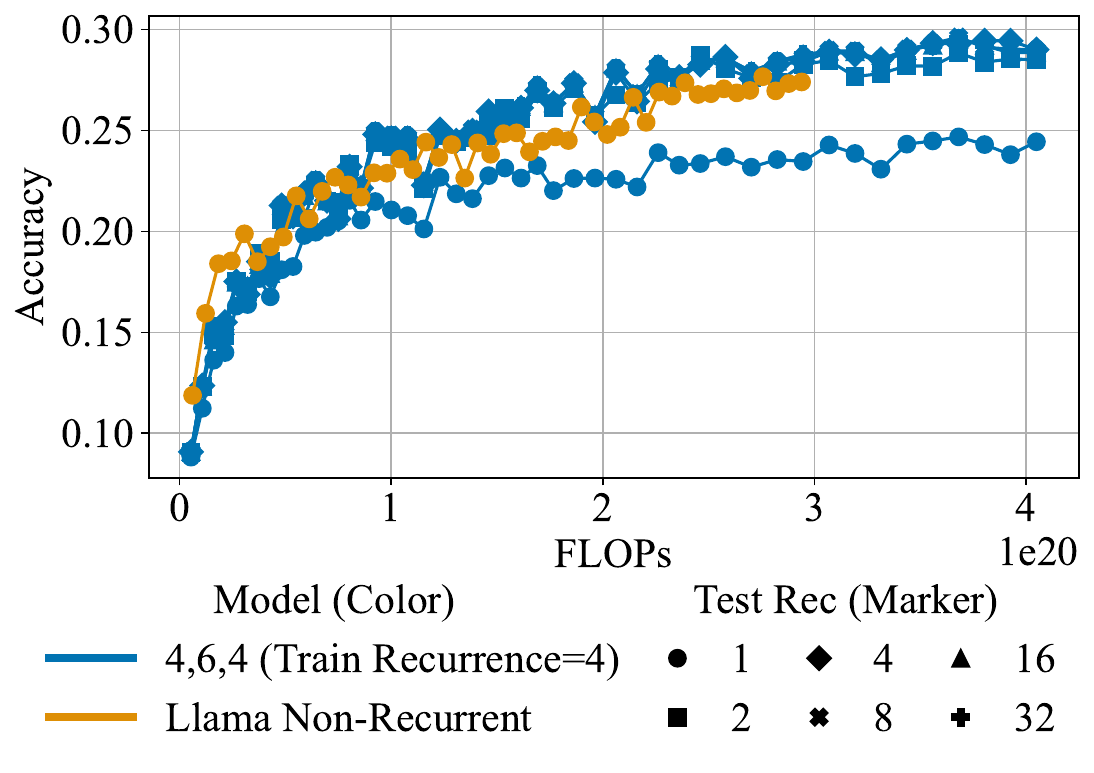}
    \includegraphics[width=0.48\linewidth]{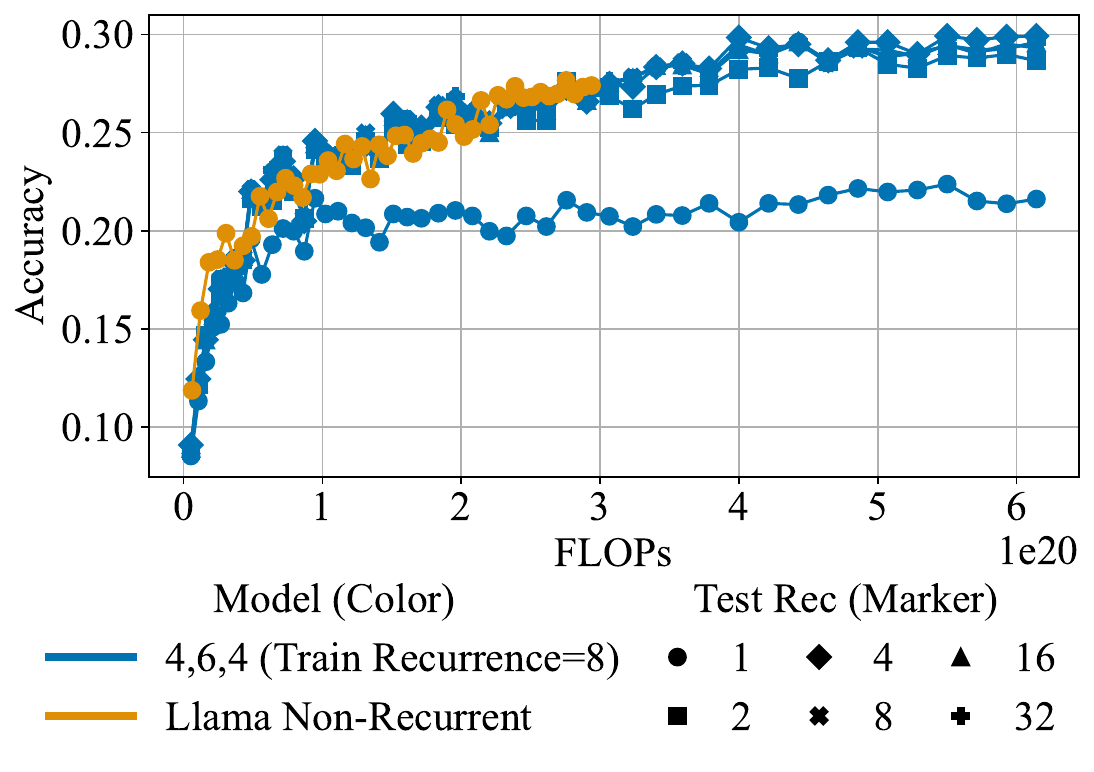}
    \caption{\textbf{Recurrence efficiently improves reasoning.} \textbf{Left}: MATH accuracy over training step for train recurrence equal to $4$ model. \textbf{Right}: MATH accuracy over training step for train recurrence equal to \(8\) model.}
    \label{app-fig:llama-MATH-4-8}
\end{figure}

\begin{figure}[ht!]
    \centering
    \includegraphics[width=0.48\linewidth]{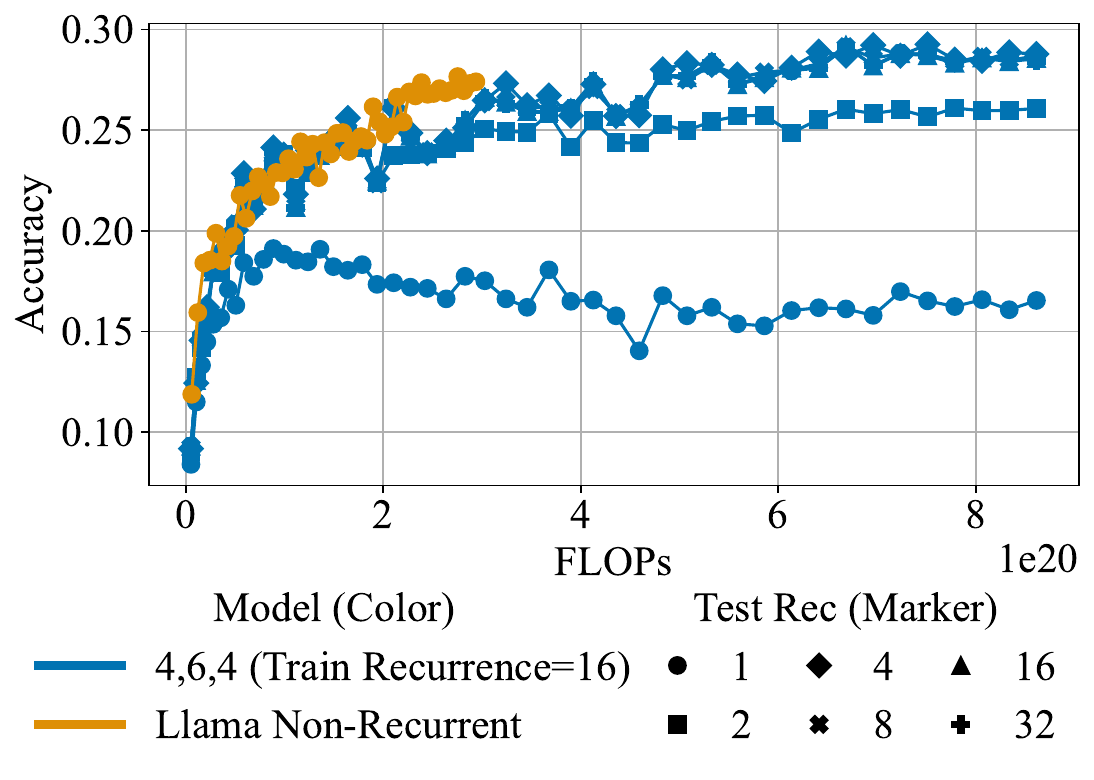}
    \includegraphics[width=0.48\linewidth]{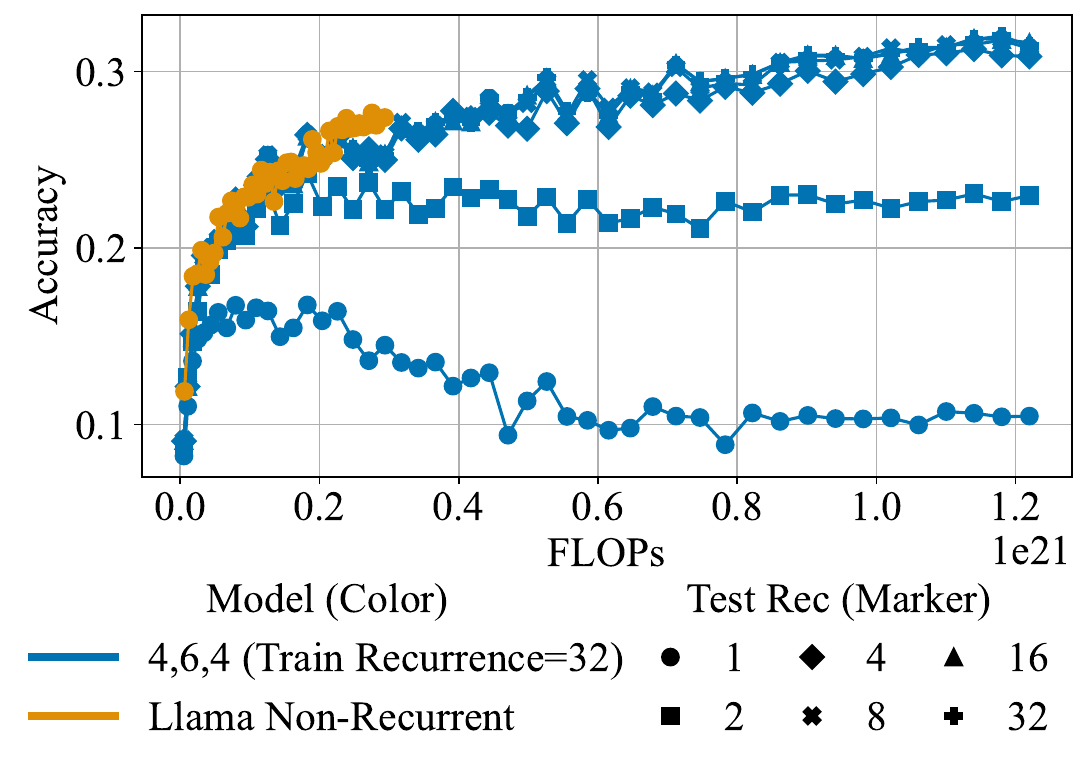}
    \caption{\textbf{Recurrence efficiently improves reasoning.} \textbf{Left}: MATH accuracy over training step for train recurrence equal to $16$ model. \textbf{Right}: MATH accuracy over training step for train recurrence equal to \(32\) model.}
    \label{app-fig:llama-MATH-16-32}
\end{figure}

In \Cref{app-fig:llama-MATH-all},  we show evaluation results for \llama on MATH.
In \Cref{app-fig:llama-MATH-effective-params}, we plot Right of \Cref{app-fig:llama-MATH-all} with an effective parameters x-axis, which can be viewed as proportional to FLOPs required for inference.
In Figures \ref{app-fig:llama-MATH-4-8} and \ref{app-fig:llama-MATH-16-32} we show the MATH accuracy over training step for train recurrences \(4,8,16\) and \(32\).

In \Cref{app-tab:llama-all-evals}, we show a broad range of evaluations for the models in \Cref{app-fig:llama-GSM8K-all}.

\begin{table}
    \centering
    \caption{\textbf{Final step accuracy for models shown in \Cref{app-fig:llama-GSM8K-all} on a broad range of evaluations.} We also include \textit{\llama-3.2-1B Hugging Face} which is our evaluations of the \llama-3.2-1B model downloaded from Hugging Face, i.e. the step 0 accuracy of the non-recurrent \llama model.}
    \begin{tabular}{cccccccccc}
        \toprule
         Test Rec  & Arc-E & Arc-C & HS & WG & MMLU & PIQA & OBQA & GSM8K & MATH \\
        \midrule
        \multicolumn{10}{c}{4,6,4 (Train Recurrence=4)} \\
        \midrule
        1 & 57.7 & 33.1 & 42.6 & 54.5 & 33.3 & 65.6 & 33.6 & 26.4 & 24.4 \\
        2 & 60.9 & 35.6 & 44.2 & 54.1 & 35.3 & 66.3 & 33.0 & 36.2 & 28.5 \\
        4 & 60.4 & 36.3 & 44.6 & 55.8 & 36.1 & 67.2 & 33.2 & 39.4 & 29.0 \\
        8 & 60.3 & 36.1 & 44.7 & 55.2 & 36.1 & 67.5 & 33.0 & 40.1 & 29.0 \\
        16 & 60.2 & 36.2 & 44.7 & 55.2 & 36.1 & 67.5 & 33.0 & 40.1 & 28.9 \\
        32 & 60.2 & 36.2 & 44.7 & 55.2 & 36.1 & 67.5 & 33.0 & 40.0 & 28.9 \\
        \midrule
        \multicolumn{10}{c}{4,6,4 (Train Recurrence=8)} \\
        \midrule
        1 & 55.3 & 34.8 & 42.4 & 53.5 & 33.8 & 66.6 & 32.2 & 21.2 & 21.6 \\
        2 & 59.6 & 36.6 & 44.9 & 55.0 & 36.1 & 67.7 & 33.2 & 36.5 & 28.7 \\
        4 & 61.0 & 37.0 & 45.5 & 54.6 & 36.9 & 67.4 & \underline{34.4} & 43.4 & 29.9 \\
        8 & 61.4 & \underline{37.3} & 45.6 & 53.9 & 36.7 & 67.5 & \underline{34.4} & 44.3 & 29.3 \\
        16 & 61.4 & 37.0 & 45.7 & 54.4 & 36.7 & 67.6 & 34.2 & 44.7 & 29.6 \\
        32 & 61.4 & 36.9 & 45.7 & 54.5 & 36.7 & 67.6 & 34.0 & 44.7 & 29.6 \\
        \midrule
        \multicolumn{10}{c}{4,6,4 (Train Recurrence=16)} \\
        \midrule
        1 & 55.2 & 33.9 & 41.7 & 51.0 & 31.5 & 65.4 & \underline{34.4} & 13.1 & 16.5 \\
        2 & 60.6 & 36.3 & 44.6 & 55.3 & 34.8 & 67.1 & 33.4 & 31.9 & 26.1 \\
        4 & 61.8 & 36.6 & 45.9 & \underline{57.5} & 36.9 & 67.0 & 34.2 & 42.2 & 28.8 \\
        8 & \underline{61.9} & 36.4 & 45.8 & 57.2 & 37.0 & 67.0 & 33.8 & 45.3 & 28.7 \\
        16 & \underline{61.9} & 36.7 & 45.8 & 57.2 & 37.0 & 66.9 & 34.0 & 45.4 & 28.6 \\
        32 & \underline{61.9} & 36.7 & 45.8 & 57.1 & 37.0 & 66.9 & 34.0 & 45.3 & 28.4 \\
        \midrule
        \multicolumn{10}{c}{4,6,4 (Train Recurrence=32)} \\
        \midrule
        1 & 53.1 & 32.4 & 40.2 & 50.2 & 28.2 & 64.9 & 31.2 & 6.4 & 10.5 \\
        2 & 58.2 & 35.2 & 44.4 & 52.7 & 34.4 & 66.5 & 31.0 & 25.5 & 23.0 \\
        4 & 61.2 & 36.4 & \underline{46.4} & 56.4 & 37.9 & 67.2 & 31.4 & 44.6 & 30.9 \\
        8 & 61.5 & 36.9 & 46.2 & 57.1 & 38.5 & 67.3 & 31.2 & 48.1 & 31.3 \\
        16 & 61.4 & 36.6 & 46.2 & 57.0 & 38.5 & 67.5 & 31.6 & \underline{49.1} & \textbf{31.6} \\
        32 & 61.4 & 36.8 & 46.2 & 56.8 & 38.4 & 67.5 & 31.6 & \textbf{49.4} & \underline{31.5} \\
        \midrule
        \multicolumn{10}{c}{Llama Non-Recurrent} \\
        \midrule
         & \textbf{62.6} & \textbf{38.2} & 45.8 & 57.1 & \textbf{38.7} & \underline{68.4} & 33.4 & 37.1 & 27.4 \\
        \midrule
        \multicolumn{10}{c}{Llama-3.2-1B Hugging Face} \\
        \midrule
         & 61.7 & 36.9 & \textbf{64.2} & \textbf{60.9} & \underline{38.6} & \textbf{74.9} & \textbf{37.2} & 4.9 & 4.3 \\
         \bottomrule
    \end{tabular}
    \label{app-tab:llama-all-evals}
\end{table}

\FloatBarrier
\subsection{Data Mixtures}
In \Cref{app-tab:data-mix-full}, we extend \Cref{tab:data-mix}, including more test recurrences. We also include the \texttt{Huginn-0125} evaluations conducted and published by \citet{geiping2025scaling} for comparison.

\begin{table}[ht!]
    \centering
    \caption{\textbf{High quality data and curricula improve recurrent model performance across benchmarks.} We see that the depth-recurrent models increases in accuracy over recurrence and achieves better accuracy when using two phrase training. For the non-recurrent baseline we see single phase training slightly outperforms two phrase training. This table extends \Cref{tab:data-mix}. \\ ** We note our context restricted and without chat template evaluations would more than likely decrease performance of Huginn-0125, hence we do not reevaluate the model under our conditions and instead state the best accuracies released by \citet{geiping2025scaling}. We note that this model has over \(4\times\) as many parameters as our \((4,8,4)\) models.}
    \begin{tabular}{cccccccccc}
        \toprule
         Test Rec & Arc-E & Arc-C & HS & WG & MMLU & PIQA & OBQA  & GSM8K & MATH\\
          \midrule
         \multicolumn{10}{c}{Random} \\
         \midrule
          & 25 & 25 & 25 & 50 & 25 & 50 & 25 & 0 & 0\\
        \midrule
        \multicolumn{10}{c}{4,8,4 (Train Recurrence=4) - Single Phase} \\
        \midrule
        1 & 50.0 & 31.6 & 50.8 & 58.0 & 35.7 & 69.3 & 38.8 & 25.6 & 8.8 \\
        2 & 52.3 & 31.9 & 55.8 & 60.5 & 39.2 & 70.9 & 38.8 & 44.1 & 13.8 \\
        4 & 53.3 & 32.8 & 57.7 & 60.9 & 39.6 & 71.3 & 39.0 & 51.2 & \textbf{14.5} \\
        8 & 52.5 & 32.4 & 58.1 & 60.8 & 39.5 & 71.2 & 38.6 & 51.8 & 14.1 \\
        16 & 52.7 & 32.8 & 58.2 & \underline{61.0} & 39.4 & 71.2 & 38.6 & \underline{51.9} & \underline{14.4} \\
        32 & 52.7 & 32.7 & 58.2 & \textbf{61.1} & 39.4 & 71.4 & 38.6 & \textbf{52.0} & \textbf{14.5} \\
        \midrule
        \multicolumn{10}{c}{4,8,4 (Train Recurrence=4) - Two Phase} \\
        \midrule
        1 & 52.7 & 31.6 & 51.5 & 56.7 & 36.2 & 71.0 & 39.4 & 26.5 & 9.7 \\
        2 & 59.3 & 34.8 & 57.3 & 58.6 & 41.3 & 71.3 & \textbf{41.0} & 44.6 & 12.3 \\
        4 & 63.8 & 36.9 & 60.0 & 58.7 & 44.3 & 73.5 & \underline{40.6} & 51.7 & 13.6 \\
        8 & \underline{65.2} & 37.4 & 60.3 & 59.9 & 44.7 & \underline{73.7} & 40.0 & \textbf{52.0} & 14.3 \\
        16 & \underline{65.2} & \underline{37.7} & \underline{60.4} & 60.2 & \underline{44.8} & 73.6 & 40.0 & 51.4 & 14.3 \\
        32 & \underline{65.2} & \underline{37.7} & \underline{60.4} & 60.5 & \underline{44.8} & 73.6 & 40.0 & 51.2 & 14.2 \\
        \midrule
        \multicolumn{10}{c}{TinyLlama-1.1b-3T
        Static Depth - Single Phase} \\
        \midrule
        & 61.2 & 35.2 & 58.9 & 60.5 & \textbf{45.1} & 71.4 & 39.2 & 46.2 & \underline{14.4} \\
        \midrule
        \multicolumn{10}{c}{TinyLlama-1.1b-3T
        Static Depth - Two Phase} \\
        \midrule
         & 62.5 & 36.5 & 60.3 & 59.6 & 44.4 & 72.9 & 39.4 & 45.2 & 12.8 \\
         \midrule
         \multicolumn{10}{c}{TinyLlama-1.1b-3T \citep{zhang2024tinyllama}} \\
        \midrule
         & 55.7 & 31.0 & 59.1 & 58.9 & 25.4 & 73.0 & 35.0 & 1.6 & 2.3 \\
         \midrule
         \multicolumn{10}{c}{Huginn-0125** -- \(3.5\)b parameters \citep{geiping2025scaling}} \\
        \midrule
         1 & 34.9 & 24.1 & 29.3 & 49.4 & 23.6 & 55.3 & 26.8 & 0.0 & 0.8 \\
         32 & \textbf{69.9} & \textbf{38.2} & \textbf{65.2} & 59.4 & 31.4 & \textbf{76.2} & 38.8 & 42.08 & 12.58 \\
         \bottomrule
    \end{tabular}
    \label{app-tab:data-mix-full}
\end{table}

\FloatBarrier
\section{Hyperparameters}~\label{app-sec:hyperparams}
We use a learning rate of \(5e^{-5}\) for AdamW and \(0.001\) for Muon with weight decay of \(1e^{-4}\).
We clip all gradients at \(1\).
We use a microbatch size of \(8\), global batch size of \(1024\) using \(8\) nodes of \(4\) AMD MI300A GPUs \citep{amd_amd_2023} by default.
For the experiments shown in \Cref{subsec:inits} and \Cref{app-subsec:model-surgery} we use a global batch size of \(4096\) on \(64\) nodes.
For experiments shown in \Cref{subsec:scheduling} and \Cref{app-subsec:schedule-rec} we use a global batch size of \(512\) on \(1\) node.
When using AdamW*, we use the same values as \citet{geiping2025scaling} for all hyper parameters which are not learning rate or weight decay. 

\section{Parameter Counts}~\label{app-sec:param-counts}
In \Cref{app-tab:non_recur_param_counts}, we give exact parameter counts for non recurrent models. 
In \Cref{app-tab:recur_param_counts}, we give exact parameter counts for recurrent models.
In \Cref{app-tab:recur_layers}, we detail the layers we take from the pretrained models to form our depth-recurrent models.

\begin{table}[h]
    \centering
    \caption{Exact parameter counts for non-recurrent models.}
    \begin{tabular}{ccc}
        \toprule
         Model Name&  Embeddings & Body\\
         \midrule
         \tinyllama-1.1B-intermediate-step-1431k-3T & \(131,072,000\) & \(968,976,384\)\\
         \llama-3.2-1B (untied) & \(525,336,576\) & \(973,146,112\)\\
         \olmo-2-0425-1B & \(411,041,792\) & \(1,073,874,944\)\\
         \bottomrule
    \end{tabular}
    \label{app-tab:non_recur_param_counts}
\end{table}

\begin{table}[h]
    \centering
    \caption{Exact parameter counts for depth-recurrent models.}
    \resizebox{\textwidth}{!}{
    \begin{tabular}{ccclll}
        \toprule
         Model Name &  Embeddings & Body & Prelude & Rec Block & Coda\\
         \midrule
         \tinyllama \((2,4,2)\) & \(131,072,000\) & \(486,572,032\) & \(121,643,008\)  & \(243,286,016\) & \(121,643,008\) \\
         \tinyllama \(4,8,4)\) & \(131,072,000\)& \(704,708,608\) & \(176,177,152\)  & \(352,354,304\) & \(176,177,152\) \\
         \tinyllama \((6,10,6)\) & \(131,072,000\) & \(968,974,336\) & \(264,265,728\)  & \(440,442,880\) & \(264,265,728\) \\
         \llama \((4,6,4)\)& \(525,336,576\) & \(851,501,056\)& \(243,286,016\)  & \(364,929,024\) & \(243,286,016\) \\
         \olmo \((4,6,4)\)& \(411,041,792\) & \(939,638,784\)& \(268,468,224\)  & \(402,702,336\) & \(268,468,224\) \\
         \bottomrule
    \end{tabular}
    }
    \label{app-tab:recur_param_counts}
\end{table}

\begin{table}[h]
    \centering
    \caption{Layers taken from original non-recurrent models to form depth-recurrent models.}
    \begin{tabular}{ccll}
        \toprule
         Model Name & Body & Prelude & Rec Block \\
         \midrule
         \tinyllama \((2,4,2)\) & \([0, 1]\)& \([16, 17, 18, 19]\)& \([20, 21]\)\\
         \tinyllama \(4,8,4)\) & \([0, 1, 2, 3]\)& \([10, 11, 12, 13, 14, 15, 16, 17]\)& \([18, 19, 20, 21]\)\\
         \tinyllama \((6,10,6)\) & \([0, 1, 2, 3,4,5]\)& \([6,7,8,9,10, 11, 12, 13, 14, 15]\)& \([16, 17, 18, 19, 20, 21]\)\\
         \llama \((4,6,4)\)& \([0, 1, 2, 3]\)& \([6,7,8,9,10,11]\)& \([12,13,14,15]\)\\
         \olmo \((4,6,4)\)& \([0, 1, 2, 3]\)& \([6,7,8,9,10,11]\)& \([12,13,14,15]\)\\
         \bottomrule
    \end{tabular}
    \label{app-tab:recur_layers}
\end{table}




\end{document}